\RequirePackage{fix-cm}
\documentclass{svjour3}
\smartqed  
\usepackage{cite}
\usepackage{subcaption}
\usepackage{graphicx}
\graphicspath{ {figures/} }
\usepackage{epstopdf}
\usepackage[justification=centering]{caption}
\usepackage{algorithm}
\usepackage{algpseudocode}
\usepackage{mathtools} 
\usepackage{amsfonts}
\usepackage{amssymb}
\usepackage{float}
\usepackage{tabulary}
\usepackage{longtable}
\usepackage{multirow}
\usepackage{enumitem}

\makeatletter
    \def\figcaption{%
        \refstepcounter{figure}%
        \@dblarg{\@caption{figure}}}
\makeatother

\journalname{}
\begin{document}

\title{Task scheduling system for UAV operations in indoor environment
}

\titlerunning{TSS for UAV operations in indoor environment}        

\author{Yohanes Khosiawan \and
    Young Soo Park \and
    Ilkyeong Moon \and
    Janardhanan Mukund Nilakantan \and
    Izabela Nielsen$^*$
}



\institute{
Y. Khosiawan \and J. M. Nilakantan \and I. Nielsen\\
email: \{yok, mnj, izabela\}@m-tech.aau.dk\\
Department of Mechanical and Manufacturing Engineering \at
              Aalborg University, Aalborg 9220, Denmark
           \and
Y. S. Park \and I. Moon\\
e-mail: \{simulacrum, ikmoon\}@snu.ac.kr\\
           Department of Industrial Engineering \at
           Seoul National University, Seoul 151-444, Republic of Korea
}

\date{Received: date / Accepted: date}

\maketitle

\begin{abstract}
Application of UAV in indoor environment is emerging nowadays due to the advancements in technology. UAV brings more space-flexibility in an occupied or hardly-accessible indoor environment, e.g., shop floor of manufacturing industry, greenhouse, nuclear powerplant. UAV helps in creating an autonomous manufacturing system by executing tasks with less human intervention in time-efficient manner. Consequently, a scheduler is one essential component to be focused on; yet the number of reported studies on UAV scheduling has been minimal.
This work proposes a methodology with a heuristic (based on Earliest Available Time algorithm) which assigns tasks to UAVs with an objective of minimizing the makespan. In addition, a quick response towards uncertain events and a quick creation of new high-quality feasible schedule are needed. Hence, the proposed heuristic is incorporated with Particle Swarm Optimization (PSO) algorithm to find a quick near optimal schedule. This proposed methodology is implemented into a scheduler and tested on a few scales of datasets generated based on a real flight demonstration. Performance evaluation of scheduler is discussed in detail and the best solution obtained from a selected set of parameters is reported.

\keywords{Indoor UAV system \and Heuristic \and Particle swarm optimization \and Scheduling \and Autonomous system}
\end{abstract}

\section{Introduction}
\label{sec:sec_intro}
In the recent years, usages of unmanned aerial vehicles (UAVs) have been increasingly prominent for various applications such as surveillance, logistics and rescue missions. UAVs are very useful for monitoring activities which are tedious and dangerous for human intervention \cite{piciarelli2013outdoor}. Most of the UAVs commercially available so far have the capability of operating in an outdoor environment \cite{von2015deploying,ahner2006assignment}. Previously, UAV applications used to be limited for only military purposes, but nowadays the situation has changed \cite{ISI:000365471600001}. UAVs are emerging as a viable, low-cost technology for use in various indoor applications \cite{shima2005uav, semiz2015task}. With advancement in technologies, the scope of UAV application in indoor environment becomes a rising interest among different industries. UAVs can be useful in indoor environments for manufacturing/service (e.g., hospital, green house and manufacturing industry) to execute multiple tasks, which has not been reported so far. UAVs can be equipped with a high resolution camera to monitor the indoor environment and UAVs can support material handling by transporting different parts/materials between locations in an indoor environment. Despite various challenges and growing interests in UAV application in indoor environment, research related to this area is at an early stage.

There are many components involved when UAV system is implemented in an indoor environment, e.g., robust wireless communication, three-dimensional trajectory data, precise UAV control, and a schedule (which reflects the required commands for UAV control). A schedule creation mainly aims at assigning tasks to UAVs which efficiently utilize available UAVs. Since there is a huge demand and minimal reported works on UAV applications in indoor environment, there is a need of research and development of UAV scheduling system, presented in this work. This is the essential gap between the problem faced in this study and state-of-the-art of UAV applications in indoor environment. The main contributions of this paper are mentioned as follows.
\begin{enumerate}[topsep=0pt]
\item Designed a system architecture for UAV applications in indoor environment,
\item Developed a methodology which includes:
\begin{itemize}
\item
heuristic based on Earliest Available Time algorithm for task scheduling with an objective of minimizing makespan,
\item incorporated the proposed heuristic with particle swarm optimization (PSO) algorithm to obtain a feasible solution in a quick computation time,
\end{itemize}
\item Tested and evaluated performance of the proposed methodology using benchmark data generated based on real flight demonstration at lab.
\end{enumerate}
The remainder of the paper is structured as follows. Section \ref{sec:sec_literature review} presents the literature survey, Section \ref{sec:sec_problem_def} explains the problem and detailed framework of proposed scheduling system. Section \ref{sec:sec_pso_for_uav_scheduling} describes the key elements involved in the implementation of PSO and the proposed methodology. Section \ref{sec:sec_numexp} and \ref{sec:sec_resultsdiscussion} discusses numerical experiments and results of the implemented methodology. Section \ref{sec:sec_conclusion} concludes the findings of this research.

\section{Literature Review}
\label{sec:sec_literature review}
The essential key to a successful UAV operations is a robust system of command and control \cite{nigam2012control}. A UAV control acts as a pilot which navigate UAV's movement to have a seamless flights during the operations. The required navigation control is derived from a command which is provided by a command center, referred as scheduler in this study.
This section gives a detailed summary of related literatures which focused on UAV scheduling. Some researchers focused on developing scheduling system for UAVs without considering travel time or distance restriction \cite{zeng2010modeling}. Authors in \cite{zeng2010modeling} developed a single-objective non-linear integer programming model for solving a UAV scheduling problem which aims at allocating and maximizing the utilization of the available UAVs in an efficient manner. They tested the proposed model using a small sized problem for an outdoor environment.

Shima and Schumacher \cite{shima2005assignment} developed scheduling methods for UAVs without fuel limitations. A mathematical model which instructs a cooperative engagement with multiple UAVs was developed. It was addressed that the simultaneous tasks assignment to multiple UAVs is an NP-hard combinatorial optimization problem. To obtain feasible solution, a genetic algorithm was proposed.
The problem mainly aims at assigning different tasks to different vehicles and consequently assigning the flying path of each vehicles. From their experimental results, it is seen that genetic algorithm is very efficient in providing real-time good quality feasible solutions. Kim et al. \cite{kim2013scheduling} proposed a mixed integer linear program (MILP) model to formalize the problem of scheduling system of UAVs. In their model, trajectories or jobs are split into different pieces and are referred as split job. This method is useful because one UAV will not be capable of covering the entire task within a single flight travel due to the fuel or battery constraint.

In another work, Kim and Morrison \cite{kim2014concerted} proposed a mixed integer linear program (MILP) for capacitated UAV scheduling. In the mentioned problem assumption, UAV should complete the tasks within its fuel (battery) capacity and should return to the based before the fuel is emptied. The proposed MILP seeks to minimize the total system cost which comprises of travel and resource cost and tries to ensure that at least one UAV is present all the times. The formulated MILP determines the types and numbers of UAVs, as well as locations and numbers of stations. A modified receding horizon task assignment heuristic was developed and compared with branch and bound algorithm to solve the same problem.

Weinstein and Schumacher \cite{weinstein2007uav} developed a UAV scheduling problem based on the inputs of vehicle routing problem which considers time windows constraint. The vehicle routing problem is solved through MILP (using CPLEX and self-implemented branch \& bound algorithm) with a target to find a global optimal schedule. Kim et al. \cite{kim2007real} proposed a scheduling model for $n$ tasks and $m$ UAVs each having a capacity limit of $q$ in a hostile environment. The proposed model aims at minimizing the cost due to the operation time and risk exposed. An MILP formulation is proposed first which exactly solves the problem and later they proposed four alternative MILPs which are computationally less intensive. The proposed model was highly complicated with huge number of variables and constraints making them impractical for applications. Improvements to the model was proposed in \cite{alidaee2009note} which minimized the number of variables and constraints.

Few works on establishing a persistent UAV service has been addressed in \cite{valenti2007mission,bethke2008uav,nigam2012control,kim2014concerted}, which concentrates on enabling a long-duration task execution. However, none of these works focused on scheduling multiple tasks executed by multiple UAVs. Furthermore, UAV operations in indoor environment makes the complexity of scheduling problem higher and there is minimal work in this area. Some aforementioned works on persistent UAV service is for surveillance purpose in indoor environment, but with no or minimal obstacles. This research focuses on developing a methodology to assign different tasks to different UAVs with obstacle avoidance in time efficient manner for indoor environment and there is a requirement of finding an exact schedule which allows the UAVs to fly autonomously. Scheduling system should react to uncertain events (e.g., UAV breakdown, fuel constraints) which may happen during UAV operations. Hence, there is a need of generating a fast feasible schedule. Generating schedules using MILPs are not computationally viable and hence using a metaheuristic is an alternative in such scenarios.

UAV scheduling is a complex problem and researchers have addressed CPLEX (employing a proprietary method which incorporates branch \& bound and branch \& cut algorithm \cite{mitchell2002branch}), heuristics, and genetic algorithm to solve it. However, from the literature review it could be postulated that there has been no work reported on using other metaheuristic algorithms to solve UAV scheduling problem. Various works have been reported in the literature where different metaheuristics are used to solve specific types of scheduling problems (e.g., job shop, flow shop, and cyclic scheduling problems) \cite{zobolas2008exact,sha2006hybrid} due to the NP-hard nature of these problems. Concept of job shop scheduling problem (where jobs are assigned to resources at particular times) and multiprocessor scheduling (where tasks are assigned to a number of processors) can be seen as a part of UAV scheduling problem \cite{garey1979guide}. Studies on job shop scheduling problem has been focused on solving different objective functions such as minimizing make span, lateness, energy consumption and maximizing utilization \cite{blazewicz1996job}. Researchers shift their focus towards metaheuristic algorithms as a popular way to address approaches on problems of this nature.

This paper proposes a methodology with heuristic based on Earliest Available Time algorithm to solve UAV scheduling problem and incorporated particle swam optimization (PSO) algorithm, a relatively new approach developed by Kennedy and Eberhart \cite{kennedy2010particle}, with an objective of minimizing the makespan. PSO is one of the prominent evolutionary computation methods employed to solve scheduling problems \cite{liu2007investigation}, but it has not been used for the addressed UAV scheduling problem in this paper. The following section provides details of the problem addressed and the framework of the proposed system.

\section{Problem Definition}
\label{sec:sec_problem_def}
This section presents details of UAV scheduling problem in indoor environment. Compared to outdoor environment, UAV application in indoor environment requires more constraints and precise controls \cite{nigam2012control}. Thus, in this section, framework of the UAV system components in indoor environment is designed in a systematical way. As a whole, Section \ref{sec:USIE}-\ref{sec:Ph_b_sf} present a \textit{reference model}\cite{choi2013modeling}, as a guide for various UAV applications in indoor environment. A \textit{reference model} can be used to employ UAV in various general system by specifying domain (environment), platform (\textit{UAV operation system} \textemdash see Figure \ref{fig:figure_uavsystem_architecture}), and the interface between them \cite{gunter2000reference}. Afterwards, the architecture of UAV scheduling system and phase-based scheduling framework are presented. In Section \ref{sec:USIE}, UAV system in indoor environment is defined in three layers.
In Section \ref{sec:ArchUSS}, UAV scheduling system is presented. Finally, the designed scheduling framework is defined in Section \ref{sec:Ph_b_sf}.

\subsection{UAV system in indoor environment}
\label{sec:USIE}
\begin{figure}[htp]
\centering
  \includegraphics[width=0.9\textwidth,keepaspectratio]{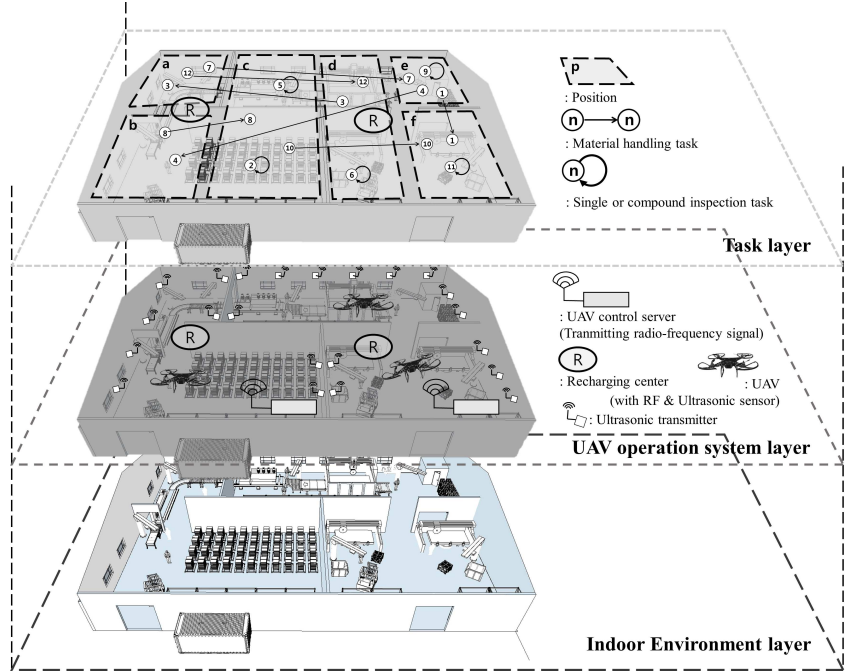}
\caption{Architecture of UAV system}
\label{fig:figure_uavsystem_architecture}
\end{figure}

A three-layers architecture of UAV system is depicted in Figure \ref{fig:figure_uavsystem_architecture}. It is a combination of logical representation of UAV operations and physical representation of UAV environment. The respective three layers are described as follows.
\begin{enumerate}
  \item \textit{Indoor environment layer} contains the infrastructure (e.g., machine, conveyor belt, assembly line) where UAV system is implemented. Infrastructure in the environment and UAV application (executed tasks) performed in it can affect each other's requirement. For instance, a dedicated corridor for UAV material handling task is defined in a highly-occupied shop floor (i.e., occupied by machines and human labors) to suppress the safety risk. Other representation of task and environment might have different characteristics that affect each other.
      Furthermore, for the purpose of collision avoidance during flights among UAVs and obstacles, the environment is segregated into zones, which practically indicate areas which are currently occupied by UAVs, permanent obstacles, and other (environmental) temporary obstacles (e.g., air turbulence due to gas pipe leak). This concept of zones will be incorporated in future work.
  \item \textit{UAV operation system layer} consists of UAVs and other support entities for UAV operation in indoor environment. Ultrasonic transmitters are mounted in indoor environment to establish UAV positioning system. UAV scheduling system operates each UAVs via UAV control server. UAV control server interacts with UAV through radio-frequency signal. Recharge station carries out autonomic UAV recharge, where UAV only needs to land on a recharge pad.
  \item \textit{Task layer} contains actions for UAVs to execute. Detailed information of each task (e.g., type of task, start \& end position, and precedence relationship) needs to be defined. Start and end position signify the origin (pick-up position) and destination (release position) for material handling task, while they will be just (identical) inspection position for single and multiple inspection task. A task may have a precedence relationship, which means it is only executed after specific tasks in predecessor list are completed.

      The proposed indoor UAV system operates multiple UAVs to execute multiple tasks. Tasks are non-preemptive and exclusively assigned to one UAV. In Table \ref{table:table_task_type}, there are three types of task: (1) single inspection, (2) compound inspection, and (3) material handling task. Single inspection consists of a flight to a specific position in three dimension, steering into a predefined direction, and image capture with the built-in camera. Compound inspection consists of multiple single inspections whose point of interests are located around one identical position. Material handling task consists of pick-up action, flight to the release point, and release action. This task is performed using a built-in equipment.
        \begin{table} [htp]
        \caption{Task types}
        \begin{center}
            \begin{tabulary}{\textwidth}{ L L L }
            \hline\noalign{\smallskip}
            Task Type & Action & Description\\
            \noalign{\smallskip}\hline\noalign{\smallskip}
            Single inspection & inspection & capture an image at a designated position\\ \hline
            Compound inspection & inspections (more than one) & capture images of points of interest around a position\\ \hline
            Material handling & pick-up -- flight -- release & transport a material from an origin to destination position\\ \hline
            \end{tabulary}
        \end{center}
        \label{table:table_task_type}
        \end{table}
\end{enumerate}

UAVs considered in this work are identical multicopters with built-in camera and material handling equipment, which can handle every types of tasks. UAVs are capacitated, hence a UAV can execute multiple tasks in one flight route to the limit of the battery capacity. UAVs can hover in air or wait-on-ground in predefined area.
With the assumed proportional weight ratio of payload to UAV, flight speed and battery consumption of the UAV are constant. In the schedule, the dimension of time is continuous. \textit{Task execution timestamp} and other \textit{action execution timestamp} (starting timestamp of task and other actions, such as recharge, hovering, and wait-on-ground), \textit{task execution time} (time required to execute a task) are planned. The integral factors of the problem are defined as follows.
\begin{enumerate}[topsep=0pt]
\item The system is deterministic; there is no uncertain event.
\item Execution of task is non-preemptive, thus not divisible into subtasks.
\item A task is executed by one UAV.
\item Every \textit{task execution time} is shorter than the flight time limit (based on the battery constraint).
\item Within the proportional payload level, UAV has a constant flight speed and battery consumption rate.
\item In every flight, a capacitated UAV has a fixed amount of:
\begin{enumerate}
\item flight time (in this work, it is up to 1200 seconds) and
\item recharge time (in this work, it is 2700 seconds) at a designated recharge station.
\end{enumerate}
\item There is no partial recharge.
\end{enumerate}

\subsection{Architecture of UAV Scheduling System}
\label{sec:ArchUSS}
To be able to execute tasks by UAVs, a UAV scheduling system is needed to assign tasks to UAVs and plan the schedule. Hence, it is an essential part of \textit{UAV operation system}. In UAV scheduling system, \textit{scheduler component} interacts with \textit{task database}, \textit{trajectory database}, and \textit{UAV database}. \textit{Task database} stores detailed information (e.g., processing time, starting \& end position, and precedence relationship) of the tasks to be executed.

\begin{figure} [htp]
\centering
\begin{subfigure}{0.4\textwidth}
\centering
  \includegraphics[scale=0.7]{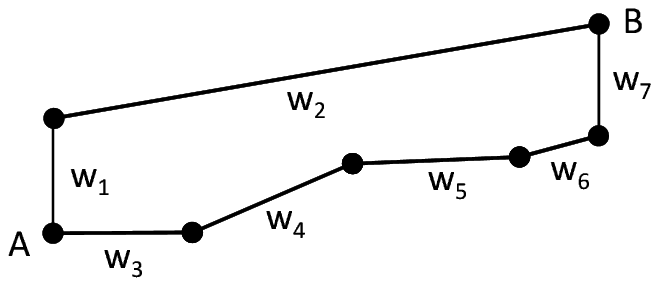}
\caption{} \label{fig:lowmap}
\end{subfigure}
\begin{subfigure}{0.4\textwidth}
\centering
  \includegraphics[scale=0.7]{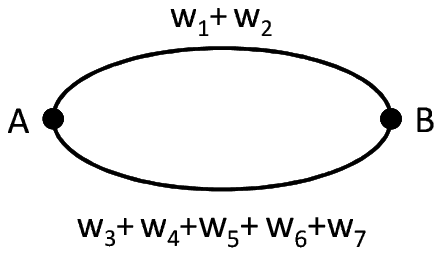}
\caption{} \label{fig:highmap}
\end{subfigure}\\
\caption{Low-level and high-level map representation}
\label{fig:figure_high_lo_map}
\end{figure}

\textit{Trajectory database} provides three-dimensional trajectory map (\textit{waypoints}, \textit{paths}, and \textit{positions} where tasks and recharges are held), including shortest possible routes between \textit{waypoints} or \textit{positions}. Figure \ref{fig:figure_high_lo_map} depicts the concept of low-level map and high-level map. Figure \ref{fig:lowmap} illustrates that low-level map consists of \textit{waypoints}, \textit{paths} between \textit{waypoints}, and \textit{position} A \& B on two designated \textit{waypoints}. Figure \ref{fig:highmap} illustrates that high-level map consists of \textit{positions} and possible routes between \textit{positions}. This route in high-level map has the total weight of respective paths which form the route itself. High-level map is required for reducing the solution space and computation size during the schedule generation, while low-level map is needed for translating the schedule to UAV-compatible instructions before it is sent to UAVs.

\begin{figure} [htp]
\centering
  \includegraphics[width=0.7\textwidth,keepaspectratio]{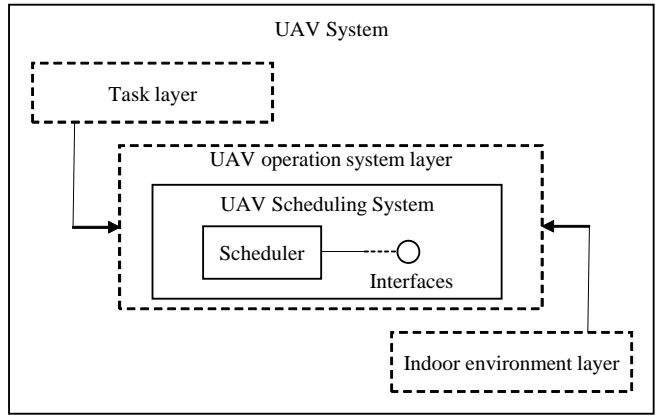}
\caption{Structural distinction of scheduler in UAV system}
\label{fig:figure_uav_system_structure}
\end{figure}

\begin{figure} [htp]
\centering
  \includegraphics[width=0.9\textwidth,keepaspectratio]{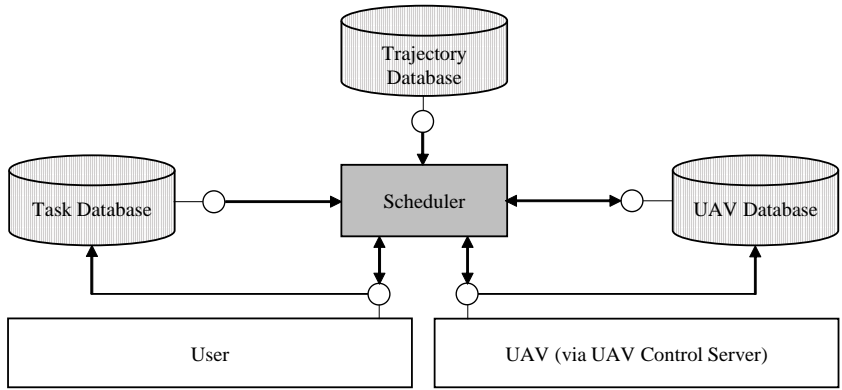}
\caption{Architecture of UAV scheduling system}
\label{fig:figure_uav_sched_arch}
\end{figure}

Figure \ref{fig:figure_uav_sched_arch} depicts interaction of scheduler with UAVs and other components. In addition, Figure \ref{fig:figure_uav_system_structure} depicts the structural distinction of scheduler in the UAV system. \textit{UAV database} consists of current status of UAV. User operates the UAV system with UAV scheduling system. The \textit{scheduler component} creates a schedule and issues it through the UAV control server (see Figure \ref{fig:figure_uav_sched_arch}). Feedbacks (events) from UAV operations are sent to \textit{scheduler component} for monitoring purpose and stored in \textit{UAV database} for historical data. In regard to uncertain event, the main role in UAV scheduling system: \textit{scheduler component} needs to be agile. Thus, there is a need of an algorithm which enables a quick generation of high quality feasible schedule (with a short makespan).

\subsection{Phase-based scheduling framework}
\label{sec:Ph_b_sf}
\textit{Scheduler component} works in phases, which are designed for abstraction of the schedule and map. Abstraction is needed to reduce the solution space and minimize computation size. On the other hand, autonomously-operated UAVs need detailed and precise command in regard to actual physical three-dimensional map. Therefore, \textit{scheduler component} applies two levels of flight schedule and map.
In accordance with the aforementioned trajectory data, there are low-level map and high-level map. Low-level schedule builds on a low-level map, specifies detailed flight routes and timestamps of subtasks (actions). High-level schedule consists of (more highly abstracted) actions such as tasks (e.g., material handling task, inspection task), flight between tasks, hovering, and wait-on-ground.

\begin{figure} [htp]
\centering
  \includegraphics[width=0.6\textwidth,keepaspectratio]{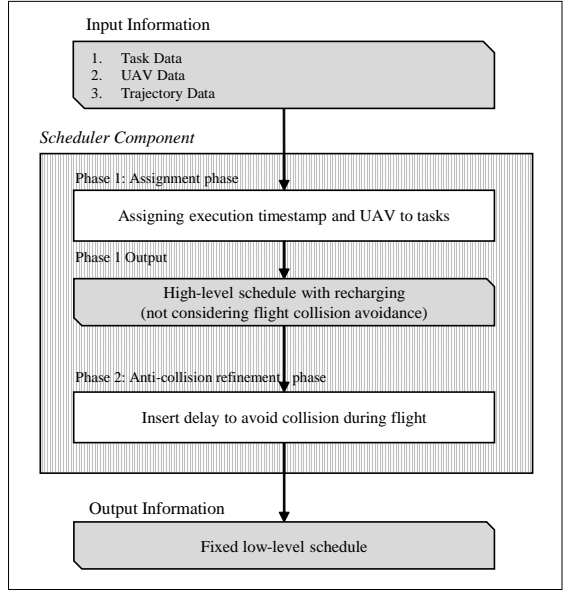}
\caption{Phase-based scheduling framework}
\label{fig:figure_phase_framework}
\end{figure}

Phase-based scheduling framework consists of two phases: assignment and anti-collision refinement. In order to respond to uncertain events, \textit{scheduler component} needs to find a feasible schedule in a short computation time. Hence, separating task assignment and physical-level routing is required for reducing the computation size. Phase-based scheduling framework is presented in Figure \ref{fig:figure_phase_framework}.

The main contribution of this work is scoped as phase 1. In phase 1, scheduler assigns tasks to UAVs. Timestamps for each UAV to start the assigned task, required recharge, hover (wait in air), and wait-on-ground are planned. Output of phase 1 is a high-level schedule of UAVs. Scheduler uses the timespan of task executions and flights to calculate the battery usage and to avoid collision while handling tasks at designated positions. A \textit{position} (high-level \textit{position}) is assigned for only one UAV at a time to avoid collision.
This procedure is organized into a proposed heuristic which is depicted in Algorithm \eqref{algo:algo_heuristic} later.

Output from phase 1 is then processed in the next phase due to the following reasons. First, for producing a UAV-compatible instructions from the obtained (high-level) schedule, a translation to a low-level schedule is needed. Second, the output of phase 1 does not consider the possible collision caused by intersecting flight \textit{path}. In phase 2, the high-level schedule is subdivided into subactions and low-level schedule is derived. To avoid collision during the flight, the delay of the flight schedule or detour planning is done by low-level \textit{path} occupation. The output of the \textit{scheduler component} is a collision-free low-level schedule for each UAV. Phase 2 will be addressed in future work.

In phase 1, there are three types of input data: (1) task data, (2) UAV data, and (3) trajectory data. Task data used in phase 1 consists of task identifier, start \textit{position}, end \textit{position}, processing time, and precedence relationship. Start and end \textit{position} are needed to assign UAV in an efficient flight route. In addition, both are used to assign only one task and UAV to a \textit{position} at a time, to avoid the UAV collision. Processing times are needed while assigning tasks into schedule and precedence relationship between tasks are used to check the current availability of each task. Table \ref{table:table_task} is an example of the task data. UAV data consists of current state, position, and battery status of each UAVs. Trajectory data used in phase 1 is a simple, distance data between positions in a high-level map. The map is considered as a complete graph, thus all the distances are calculated based on the shortest \textit{path} between positions.
\begin{table} [htp]
\caption{Task data}
\begin{center}
    \begin{tabulary}{\textwidth}{ C|C|C|C|C }
    \hline\noalign{\smallskip}
    TaskID & Start Position & End Position & Processing Time (seconds) & Precedence\\
    \noalign{\smallskip}\hline\noalign{\smallskip}
    1 & e & f & 243 & -\\
    2 & c & c & 245 & -\\
    3 & d & a & 719 & -\\
    4 & e & b & 550 & 1\\
    5 & c & c & 235 & 2\\
    6 & d & d & 241 & 2\\
    7 & a & e & 478 & 4\\
    8 & b & c & 304 & 4;5\\
    9 & e & e & 395 & 7\\
    10 & c & f & 344 & 6;8\\
    11 & f & f & 270 & 10\\
    12 & a & d & 514 & 3;6\\ \hline
    \end{tabulary}
\end{center}
\label{table:table_task}
\end{table}

Table \ref{table:table_flight_time} is an example of the trajectory data containing 6 \textit{positions} (\textit{a}$\sim$\textit{f}) and 2 recharge stations (\textit{R1}, \textit{R2}), based on the shortest flight time.
This data example is used to illustrate how the proposed methodology works in the following section.
\begin{table} [htp]
\caption{Positions and distance (in flight time unit) data}
\begin{center}
    \begin{tabular}{ c|c|c|c|c|c|c|c|c }
    \hline\noalign{\smallskip}
    From/To & a & b & c & d & e & f & R1 & R2\\
    \noalign{\smallskip}\hline\noalign{\smallskip}
    a & 0 &  108 & 131 & 222 & 376 & 353 & 40 & 160\\
    b & 108 & 0 & 120 & 241 & 347 & 371 & 60 & 160\\
    c & 131 & 120 & 0 & 127 & 228 & 254 & 60 & 60\\
    d & 222 & 241 & 127 & 0 & 116 & 122 & 160 & 40\\
    e & 376 & 347 & 228 & 116 & 0 & 123 & 260 & 60\\
    f & 353 & 371 & 254 & 122 & 123 & 0 & 260 & 60\\
    R1 & 40 & 60 & 60 & 160 & 260 & 260 & 0 & 120\\
    R2 & 160 & 160 & 60 & 40 & 60 & 60 & 120 & 0\\ \hline
    \end{tabular}
\end{center}
\label{table:table_flight_time}
\end{table}

The problem draws input from multiprocessor scheduling and job shop scheduling, which are well-known NP-hard problems \cite{zobolas2008exact,sha2006hybrid,garey1979guide} (e.g., when a schedule with minimum makespan is pursued), and there is a need of metaheuristic algorithm to obtain a high quality feasible solution. In addition, there are various possible objectives in UAV scheduling problem, similar to the aforementioned scheduling problems. Among different objectives \cite{sha2010multi,ponnambalam2000multi}, minimization of makespan is selected to be optimized in this work. Furthermore, a proposed heuristic is incorporated with particle swarm optimization (PSO) algorithm which will be covered in the following section.

\section{Application of PSO for UAV Scheduling System}
\label{sec:sec_pso_for_uav_scheduling}
Characteristics of the presented problem distinguish its nature as NP-hard. Approaches based on branch \& bound and branch \& cut are tedious in terms of computation time. A promising alternative to those methods are metaheuristic algorithm. Metaheuristics use different concepts derived from artificial intelligence and evolutionary algorithms, which is inspired from mechanisms of natural evolution \cite{sha2010multi}. From the literature it could be found that metaheuristics can also be called as soft computing techniques, evolutionary algorithms and nature inspired algorithms. Metaheuristics methods are designed for solving wide range of hard optimization problem without having to adapt deeply into each problem. These algorithms are fast and easy to implement \cite{sorensen2013metaheuristics}. From the literature review it could be seen that different metaheuristic algorithms have been proposed to solve general scheduling problems \cite{garrido2000heuristic} to obtain a feasible solution.

Particle swarm optimization (PSO) algorithm is a swarm-based stochastic optimization technique developed by Kennedy and Eberhart \cite{kennedy2010particle} based on the characteristics of social behavior of birds in flocks or fish in schools is chosen to solve the problem addressed in this paper. PSO algorithm does not involve usage of genetic operators (mutation and crossover) which are commonly used for evolutionary algorithms. PSO is an optimization method based on the population which is referred as swarm in this paper. Simplicity in application, easy implementation and faster convergence of PSO has made the algorithm widely acceptable among researchers for solving different types of optimization problems \cite{hassan2005comparison}. Different variants of PSO has been developed and employed by researchers to solve scheduling problems. This paper employs the standard PSO \cite{kennedy2010particle} model to solve the UAV scheduling problem. Pseudo code of the PSO is presented in Algorithm \ref{algo:algo_psoheuristic}.
Individuals in the swarm share information among each other which helps to search towards the best position in the search space. Each single solution in the search space is called as a particle. All particles are to be evaluated by an objective function (explained in the following section) which is to be optimized. Each particle in the swarm searches for the best position and it travels in the search space with a certain velocity. Best fitness encountered by each particle (local best) is stored and the information is shared with other particles to obtain the best particle (global best).

\begin{algorithm} [htp]
\caption{Particle Swarm Optimization Algorithm}\label{algo:algo_psoheuristic}
\begin{algorithmic}[1]
\Statex \textbf{Input:} Initial Swarm ($swarm$)
\Statex \textbf{Output:} schedule of tasks on UAVs ($schedule$)
\State Initialize (parameters, swarm, local best and global best)
\While {stop condition not met}
\State $velocity \leftarrow$  updateVel(swarm, velocity, local best, global best);
\State $swarm \leftarrow$ updateSwarm(swarm, velocity);
\State $localBest \leftarrow$ getLocalBest(fitness(swarm), localbest);
\State $globalBest \leftarrow$ getGlobalBest(localBest, globalBest);
\State  generation++;
\EndWhile
\end{algorithmic}
\end{algorithm}

The procedural steps of the PSO algorithm are given below:

\textbf{Step 1}:  Initial population is generated based on heuristic rules. Initial velocities for each particle are randomly generated.

\textbf{Step 2}: Based on the objective function each particle is evaluated.

\textbf{Step 3}: Each particle remembers the best result achieved so far (local best) and exchange information with other particles to obtain the best particle (global best) among the swarm.

\textbf{Step 4}: Velocity of the particle is updated using Equation \eqref{eq:eq_velocity_update} and using Equation \eqref{eq:eq_position_update} the position of the particle is updated.\\

Velocity update equation:
\begin{equation}
\label{eq:eq_velocity_update}
P_i^{t+1}=P_i^{t}+v_i^{t+1}
\end{equation}

Particles moves from their current position to the new position using Equation \eqref{eq:eq_position_update}. Particle positions are updated in every iteration.

Position update equation:
\begin{equation}
\label{eq:eq_position_update}
v_i^{t+1}=v_i^{t}+\underbrace{c_1\text{x}[U_1\text{x}(^{lo}P_i^t)]}_\text{cognitive part}+\underbrace{c_2\text{x}[U_2\text{x}(G_t-P_i^t)]}_\text{social part}
\end{equation}

Where $U_1$ and $U_2$ are known as velocity coefficients (random numbers between 0 and 1), $c_1$ and  $c_2$ are known as learning (or acceleration) coefficients, $v_i^t$ is the initial velocity, $^{lo}P_i^t$ is the Local best, $G^t$ is the global best solution at generation $t$ and $P_i^t$ is the current particle position.
Step 5: Go back to step 2 until the termination criterion is met.
Equation \eqref{eq:eq_velocity_update} and \eqref{eq:eq_position_update} describes the path in which the particles fly in the search space. Equation \eqref{eq:eq_velocity_update} consists of two main parts. First part is known as cognitive part which controls the traveling of the particles based on its own flying experience. Second part is known as social part which helps in collaborating with other particles based on the group flying experience \cite{blondin2009particle}.

\subsection{PSO entities}

To illustrate the explanation of different entities in PSO, a sample dataset presented in Table \ref{table:table_task} and Table \ref{table:table_flight_time} is used.

\subsubsection{Initial population}
Metaheuristic algorithms start with random search space which iteratively evolves to find a near optimumsolution\cite{ponnambalam2000multi}. The purpose for doing this is to start the search from hypothetically good starting points rather than random ones, so that global optimum is more likely achieved in less time. In this paper, six heuristic rules (maximum rank positional weight, minimum inverse positional weight, minimum total number of predecessors tasks, maximum total number of follower tasks, maximum and minimum task time) presented in \cite{ponnambalam2000multi} are used to generate the initial particles. Two more heuristic rules based on the number of predecessor and follower tasks are added to the existing rules and these are named as cumulative number of predecessor and follower tasks. It is reported in the literature that, higher the initial population, the quality of the solution will improve. Hence, in this research forty initial particles are generated. Remaining particles are generated by swapping the tasks without violating the precedence relationship. Table \ref{table:table_heuristic_rules} reports the set of initial particles generated using the heuristic rules and these particles meets the precedence condition. The precedence relationships of tasks are available in Table \ref{table:table_task}. Each particle is structured as a string of tasks which are to be used for the UAV scheduling problem.  All particles are assigned with a random velocity and the number of velocity pairs are generated randomly and the details are explained in the next section.

\begin{table}
\caption{Priority rules for initial particle generation}
\begin{center}
    \begin{tabulary}{\textwidth}{C|C|C|C|C|C|C|C|C|C|C|C|C}
    \hline\noalign{\smallskip}
    Heuristic Rules & \multicolumn{12}{c}{Task Sequence (Generated Particle)}\\
    \noalign{\smallskip}\hline\noalign{\smallskip}
    Maximum Ranked Positional Weight & 1 & 2 & 4 & 6 & 5 & 8 & 3 & 7 & 10 & 9 & 11 & 12 \\ \hline
    Minimum Inverse Positional Weight & 1 & 2 & 3 & 4 & 5 & 6 & 7 & 12 & 9 & 8 & 10 & 11 \\ \hline
    Minimum Total Number Of  Predecessors Tasks	& 1 & 2 & 3 & 4 & 5 & 6 & 7 & 9 & 12 & 8 & 10 & 11 \\ \hline
    Maximum Total Number of Follower Tasks & 1 & 2 & 4 & 5 & 6 & 8 & 3 & 7 & 10 & 9 & 11 & 12 \\ \hline
    Maximum Task Execution Time & 3 & 2 & 1 & 4 & 7 & 9 & 6 & 12 & 5 & 8 & 10 & 11 \\ \hline
    Minimum Task Execution Time & 1 & 2 & 5 & 6 & 4 & 8 & 10 & 11 & 7 & 9 & 3 & 12 \\ \hline
    Minimum Number of Cumulative Predecessor Tasks & 1 & 2 & 3 & 4 & 5 & 6 & 7 & 9 & 8 & 10 & 11 & 12 \\ \hline
    Maximum Number of Cumulative Follower Tasks & 2 & 6 & 1 & 4 & 3 & 5 & 7 & 8 & 10 & 9 & 11 & 12 \\ \hline
    \end{tabulary}
\end{center}
\label{table:table_heuristic_rules}
\end{table}

\subsubsection{Initial Velocity}
Each particle is assigned with velocity pairs and these pairs are randomly generated. In this problem, each pair represents the transpositions in the particle. Based on the pilot experiments it is decided to have different size of velocity pairs for different problems based on the number of tasks. Table \ref{table:table_initial_velocity_size} presents the maximum number of velocity pairs used in this research for different sizes of task. Velocity is updated from the second iteration using Equation \eqref{eq:eq_velocity_update}. The number of pairs is same throughout all iterations. For example, if the task size of the problem is within the range of 0-20, number of velocity pairs is selected as 2.

\begin{table}
\caption{Range of number of initial velocity pairs}
\begin{center}
    \begin{tabular}{ c | c }
    \hline\noalign{\smallskip}
    Task Range & Maximum Number of Velocity Pairs\\
    \noalign{\smallskip}\hline\noalign{\smallskip}
    0-20 & 2\\
    20-50 & 10\\
    50-100 & 30\\ \hline
    \end{tabular}
\end{center}
\label{table:table_initial_velocity_size}
\end{table}

To illustrate how the velocity and position update works is explained with an example. Following data presented in Table \ref{table:table_pso_dataset} is used as the parameters needed to explain the update process.

\begin{table}
\caption{Example data and parameters for PSO updation}
\begin{center}
    \begin{tabular}{ l | c | l }
    \hline\noalign{\smallskip}
    Data or Parameter & Notation & Value\\
    \noalign{\smallskip}\hline\noalign{\smallskip}
    Local Best & $^(lo)P_i^t$ & [1, 2, 4, 6, 5, 8, 3, 7, 10, 9, 11, 12]\\
    Global Best & $G^t$ & [2, 6, 1, 4, 3, 5, 7, 8, 10, 9, 11, 12]\\
    Particle & $P_i^t$ & [1, 2, 4, 6, 5, 8, 7, 3, 10, 9, 12, 11]\\
    Initial velocity & $v_i^t$ & (6,7),(10,11)\\
    Learning coefficient 1 & $c_1$ & 1\\
    Learning coefficient 2 & $c_2$ & 2\\
    Velocity coefficient & $U_1$ & 0.2\\
    Velocity coefficient & $U_2$ & 0.4\\ \hline
    \end{tabular}
\end{center}
\label{table:table_pso_dataset}
\end{table}

These particles represents the tasks arranged in such a manner it satisfies the precedence constraints. Velocity of the current particle is updated based on Equation \eqref{eq:eq_velocity_update} as follows.

\begin{flalign*}
v_i^{t+1}= &(6,7) (10,11) + 0.2 \text{x}&\\
 &[(1,2,4,6,5,8,3,7,10,9,11,12) – (1,2,4,6,5,8,7,3,10,9,12,11)] +&\\
 &0.6 \text{x} [(1,2,3,4,5,6,7,8,9,10,11) - (1,2 ,3,6,5,4,7,8,10,9,11)]&\\
= &(6, 7) (10, 11) + 0.2 \text{x} (6, 7) (10, 11)&\\
 &+ 0.8 \text{x} (0, 1) (1, 3) (2, 3) (4, 7) (5, 7) (10, 11)&\\
= &(6, 7) (10, 11) (0, 1) (1, 3) (2, 3) (4, 7) (5, 7)&
\end{flalign*}

Using Equation \eqref{eq:eq_position_update} current particle is updated to a new particle using the new velocity.
\begin{flalign*}
P_i^{t+1}= &(1, 2, 4, 6, 5, 8, 7, 3, 10, 9, 12, 11) +&\\
&(6, 7) (10, 11) (0, 1) (1, 3) (2, 3) (4, 7) (5, 7)&\\
= &(2, 6, 1, 4, 7, 5, 3, 8, 10, 9, 11, 12)&
\end{flalign*}

The product of coefficient value $c_1$, $U_1$ and $c_2$, $U_2$ works like a probability percentage which decides how many pairs would be copied to form the updated velocity. For example when the probability percentage is 80\% ($c_2$ \text{x} $U_2$= 2 \text{x} 0.4 = 0.8), 80\% of the pairs would be copied to the new updated velocity. In this example, 6 pairs are formed when transpositions takes place between the global best and the current particle and based on this probability percentage 5 pairs are chosen out of 6 for the updated velocity. However, if any of the pairs are already present from other transpositions or initial velocity, it is discarded.
A repair mechanism is incorporated to convert an infeasible sequence to a feasible one which meets the precedence relationship.

\subsection{Schedule creation and evaluation}
\begin{figure} [htp]
\centering
  \includegraphics[width=0.8\textwidth,keepaspectratio]{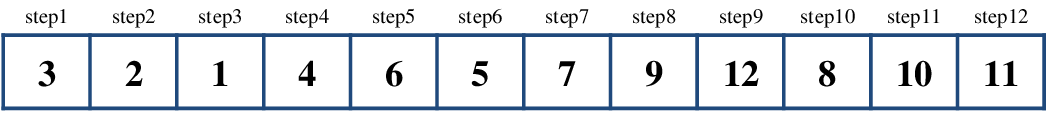}
\caption{Task scheduling steps}
\label{fig:figure_task_scheduling_step}
\end{figure}

In a planning horizon, given tasks are represented as a sequence according to priority rules explained in Table \ref{table:table_heuristic_rules}. Tasks in task sequence are scheduled, each of them is assigned to a UAV to be executed in a particular timespan, one-by-one (per step) according to its order in the sequence. Figure \ref{fig:figure_task_scheduling_step} depicts sequential steps of task scheduling during schedule creation from a sequence of 12 tasks.

The aforementioned heuristic for creating a schedule from a task sequence is depicted in Algorithm \ref{algo:algo_heuristic}. The idea behind this algorithm is to create a schedule which is driven to utilize the available resources in the following manners; which in general lead to a time-efficient characteristic:
\begin{itemize}[noitemsep, topsep=0pt]
  \item balanced
  \begin{itemize}
  \item task assignment towards earliest available UAV (which indicates its relative idleness compared to other UAVs)
  \end{itemize}
  \item safe
  \begin{itemize}
  \item no multiple UAVs are allowed to occupy a position simultaneously
  \item task execution following precedence
  \end{itemize}
  \item early
  \begin{itemize}
  \item recharge station which eventually deliver a recharged UAV to its next destination early is chosen
  \item short makespan for the whole tasks completion
  \end{itemize}
\end{itemize}
The detailed procedure of the heuristic is explained in Algorithm \ref{algo:algo_heuristic}.

\begin{algorithm} [htp]
\caption{Earliest Available Time Algorithm}\label{algo:algo_heuristic}
\begin{algorithmic}[1]
\Statex \textbf{Input:} sequence of tasks ($sequence$), list of UAVs ($uavs$)
\Statex \textbf{Output:} schedule of tasks on UAVs ($schedule$)
\For {each $task$ in $sequence$}
\State $pos\_at \leftarrow$  max(getReleaseTSTP($task$.startPos), getReleaseTSTP($task$.endPos))
\State $pred\_at \leftarrow$ 0
\For {each $predecessor$ in getPrecedence($task$)}
\State $pred\_at \leftarrow$ max(getEndTSTP($predecessor$), $pred\_at$)
\EndFor
\State $task\_at \leftarrow$ max($pos\_at$, $pred\_at$)
\For {each $uav$ in $uavs$}
\State $uav\_rt \leftarrow$ getReleaseTSTP($uav$)
\State $ft \leftarrow$ getFlightTime(getCurPos($uav$))
\State $taskPrepTime \leftarrow$ max($ft$, $task\_at$-$uav\_rt$)
\State $taskPrepTSTP \leftarrow$ max($uav\_rt$+$ft$, $task\_at$)
\State $taskEndTSTP \leftarrow taskPrepTSTP$ + $taskExecTime$
\State $taskUpBoundTime \leftarrow taskPrepTime$ + $task$.procTime + getTimeToNearestRS($task$.endPos)
\If {getBattery($uav$ $<$ $taskUpBoundTime$)}
\State $rchInfo \leftarrow$ getEarliestRechargeCompletion(getCurPos($uav$), $task$.startPos)
\State $preparedTSTP \leftarrow$ max($rchInfo$.endTSTP + getFlightTime($rchInfo$.pos, $task$.startPos), $task\_at$)
\State $startTSTPCandidates$.put($uav$, $preparedTSTP$)
\State $taskEndTSTP \leftarrow taskPrepTSTP$ + $task$.procTime
\EndIf
\EndFor
\State $earliestUAV \leftarrow$ getUAVwithEarliestStartTSTP($startTSTPCandidates$)
\State $taskStartTSTP \leftarrow startTSTPCandidates$.get($earliestUAV$)
\State putTaskIntoSchedule($schedule$, $earliestUAV$, $task$)
\State setCurPos($earliestUAV$, $task$.endPos)
\State setReleaseTSTP($task$.startPos, $taskEndTSTP$)
\State setReleaseTSTP($task$.endPos, $taskEndTSTP$)
\EndFor
\end{algorithmic}
\end{algorithm}

In the incorporated PSO, a particle indicates a solution (one schedule) which is selected through an iterative search process based on its fitness value.
However, in regard to position update process in PSO algorithm, it is difficult to define a step (of position update) due to the rigid structure of the schedule.
A schedule may contain the following possible action abstractions: flight, hover, wait-on-ground, and task. When a schedule is observed, there is a precise timespan in the schedule which most likely fits for only one particular task. An action's existence may also depend on another (i.e., flight, hover, and wait-on-ground existence is relative to task execution manner). Thus, when some elements are swapped (manipulated to produce different solution), an infeasible schedule is frequently formed.

Considering the aforementioned conditions, the sequence representation (before it is created into schedule) of the tasks is considered. Swapping tasks in the sequence forms a new sequence of tasks; it is tractable and robust. Since each task sequence uniquely corresponds to a schedule, it is valid to use task sequence as the particle representation. Obviously, the representation gap (of sequence and schedule) is filled with the proposed heuristic method, which creates a schedule from a task sequence. In the end, fitness of the schedule is evaluated based on its makespan. In this manner, both position update operation (in PSO iteration) and solution fitness evaluation are done seamlessly.

Corresponding to Algorithm \ref{algo:algo_heuristic}, each task in the sequence (line 1) is put into schedule, based on an earliest available time characteristic of involved objects, described as follows.
\begin{enumerate}
  \item Task availability check
  \begin{enumerate}
  \item Position availability (line 2) \\
  An abstraction of task is broken down into possible flights and other actions (e.g., pick-up \& release payload, inspection/capture image), which involves start position and end position. Time spent at start and end position is not defined for any task in this study.
  Hence, both positions are occupied during the whole execution time of the respective task.
  Consequently, start position and end position are checked for its occupancy status when a task is picked to be put into schedule. The latest released (not occupied) timestamp is used for the position availability timestamp.
  \item Task precedence (line 3-6) \\
  Task is checked for existing preceding task. If preceding tasks exist, the last completion timestamp of them will be set as the earliest available timestamp for the task.
  If there is not any preceding task, then the task is available at time 0.
  \end{enumerate}
  \item UAV availability check
  \begin{enumerate}
  \item UAV ready time (line 9) \\
  Task-occupancy of each UAV is checked. The moment it went to idle after completing the most recent task is recorded as its ready time.
  \item Battery level (line 15-20) \\
  After task execution, UAV must have enough battery level to at least fly towards the nearest recharge station.
  \item Recharge time (line 16-17) \\
  If UAV doesn't have enough battery to go to recharge station after executing a task, then UAV needs to go to the nearest recharge station to get fully recharged before flying to the start position of the task and execute it.
  To ensure that the UAV is fully charged (at a capacity of 1200 seconds flight), an actual recharge time-span is always set to 2700 seconds; it is the time required to do one-loop of full battery recharge.
  Recharge time is summation of round-trip time and actual recharging time-span (which might be longer than 2700 seconds due to delayed recharge station availability time).
  \item Recharge station availability (line 16) \\
  A limited number of recharge slots at the recharge station is considered. When all recharge slots at a recharge station are occupied, then its earliest available time is the earliest timestamp when an occupying UAV leaves the recharge station. The next selection criteria of recharge station is based on the shortest round-trip (end position of previous task $\rightarrow$ recharge station $\rightarrow$ start position of current task) time to a recharge station. It means that the nearest recharge station might not be preferred due to its far distance to the start position of the current task. Hence, recharge station availability timestamp (\textit{rsaTSTP}) calculation (which involves recharge station $r$, and end position of previous task $getCurPos(uav)$) is formulated as follows.

  \begin{flalign*}
  chargeTSTP = &uav\_rt + getFlightTime(getCurPos(uav), getPos(r))&\\
  rToS = &getFlightTime(getPos(r), task.startPos)&\\
  rsaTSTP = &(slotReleaseTSTP \leq chargeTSTP ?&\\
  &chargeTSTP : slotReleaseTSTP) + rToS&\\
  \end{flalign*}

  A timestamp when UAV arrives at the recharge station is recorded (\textit{chargeTSTP}). To calculate \textit{rsaTSTP}, \textit{chargeTSTP} is summed with the flight time (\textit{rToS}) from recharge station to the start position of the task at hand. An exception exists if all slots at recharge station are occupied and at least one slot will be unoccupied at timestamp \textit{slotReleaseTSTP}. If \textit{slotReleaseTSTP} is later than \textit{chargeTSTP}, then \textit{slotReleaseTSTP} is used instead of \textit{chargeTSTP} in the \textit{rsaTSTP} calculation.

  This \textit{rsaTSTP} is then incorporated with recharge time for deciding the preferred recharge station. In this manner, not only the earliest available recharge station for the UAV is considered, but also the one that is near to start position of its next task. Then eventually, it promotes an earlier available time of the next task execution.
  \end{enumerate}
  \item Overall availability check (line 11-20)
  Incorporating the aforementioned task availability and UAV availability check, the start timestamp of the respective task is calculated. UAV with overall earliest available time is picked (line 22-23) and the task is put into the respective UAV's schedule. Completion (end) timestamp of a task potentially performed by each UAV is also calculated (line 19) and used to release the occupied positions.
\end{enumerate}

\begin{figure} [htp]
\centering
\begin{subfigure}{0.4\textwidth}
\centering
  \includegraphics[scale=0.5]{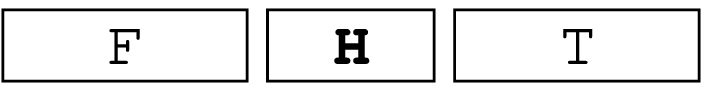}
\caption{} \label{fig:hoverwog1}
\end{subfigure}
\begin{subfigure}{0.4\textwidth}
\centering
  \includegraphics[scale=0.5]{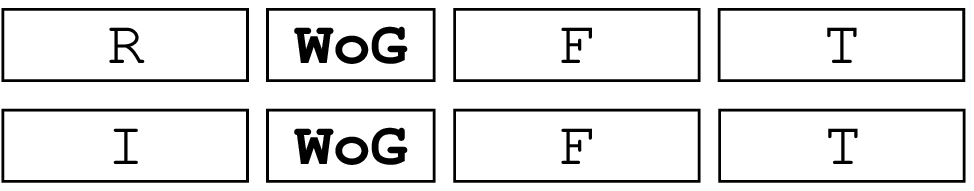}
\caption{} \label{fig:hoverwog2}
\end{subfigure}\\
\begin{subfigure}{0.4\textwidth}
\centering
  \includegraphics[scale=0.5]{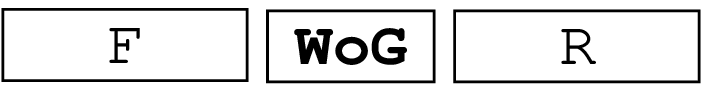}
\caption{} \label{fig:hoverwog3}
\end{subfigure}
\begin{subfigure}{0.4\textwidth}
  \includegraphics[width=\linewidth,keepaspectratio]{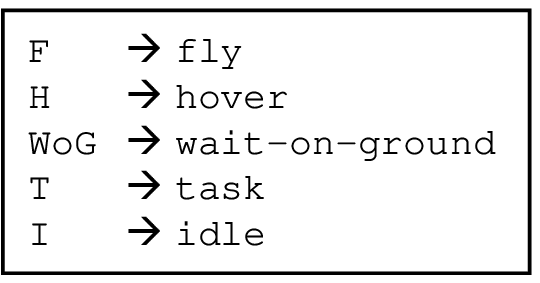}
\label{fig:hoverwoglegend}
\end{subfigure}
\caption{Conditions of hover and WoG insertion into schedule}
\label{fig:figure_hoverwog}
\end{figure}

In Algorithm \ref{algo:algo_heuristic}, for readability, \textit{putTaskIntoSchedule} (line 24) encapsulates processes of:
\begin{enumerate}[topsep=0pt]
  \item Inserting a task into the schedule of the picked (earliest available) UAV
  \item Inserting a recharge action (if required) into the schedule of the picked (earliest available) UAV
  \begin{enumerate}
  \item Updating occupancy status of the respective recharge slot
  \end{enumerate}
  \item Updating the respective UAV's battery level
  \item Inserting \textit{hover} and \textit{wait-on-ground} in between tasks in the schedule when needed.
  There are three conditions when a hover or wait-on-ground is required:
  \begin{enumerate}
  \item \textit{UAV reaches its destination position, and that position is still occupied.}\\
  When this condition occurs, the UAV shall hover around the position, since it is not safe to just land anywhere on the field; only landing at recharge station is allowed.
  Figure \ref{fig:hoverwog1} depicts a sequence of actions where a UAV is most-recently flying towards an occupied position, hovers for a while, and finally perform a task at the freed (available) position.
  \item \textit{UAV is at a recharge station, and the next target position of the UAV is not available.}\\
  When this condition occurs, it is better for the UAV to wait-on-ground instead of start flying right away and hover around afterwards because the position is still occupied.
  Figure \ref{fig:hoverwog2} depicts a sequence of actions where a UAV is at a recharge station (performing recharge or simply being idle in the beginning of planning horizon), wait-on-ground for a while because its destination (position) is occupied, and finally fly towards the start position of the task to-be-performed next.
  \item \textit{Recharge is to be performed at a recharge station and all slots are still occupied.}\\
  When this condition occurs, UAV shall wait-on-ground till one of them is free.
  Figure \ref{fig:hoverwog3} depicts a sequence of actions where a UAV is going to perform a recharge while all slots in the designated recharge stations are occupied. When it arrives at the recharge station, it performs wait-on-ground until one of the slots free, and recharge eventually.
  \end{enumerate}
\end{enumerate}

To illustrate the usage of the heuristic described in Algorithm \ref{algo:algo_heuristic}, step 1-7 of task scheduling process for given task sequence in Figure \ref{fig:figure_task_scheduling_step} are presented. Step 1 and 2 are depicted in Figure \ref{fig:step1} and \ref{fig:step2}, while step 3-7 are presented in Appendix \ref{app:figure_steps_desc}. Steps 8-12 which are principally doing the same procedure as step 1 and 2 are not presented in the paper. In accordance with the explanation of Algorithm \ref{algo:algo_heuristic}, task and UAV availability check are performed every time a task is picked to be put into the schedule. Detailed stepwise procedure for the first two steps are explained below the figures. Task and UAV availability check are referred as point (a) and (b) respectively.

\begin{longtable}{c | l l}
    \multicolumn{3}{c}{
    \parbox{0.9\columnwidth}{
      \centering
      \includegraphics[scale=0.7]{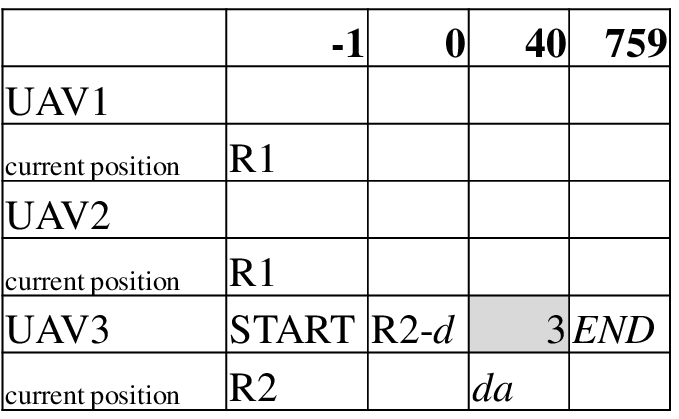}
      \figcaption{Output of step 1 of schedule creation heuristic}
      \label{fig:step1}}
    }\\
    \hline
    Step & \multicolumn{2}{c}{Description}\\
    \hline
    1 & \multicolumn{2}{l}{Task 3}\\
    &\multicolumn{2}{l}{(a) Position availability}\\
    &\multicolumn{2}{l}{Task 3 is started at position d, position d is available from time 0.}\\
    &\multicolumn{2}{l}{$\cdot$ Task precedence}\\
    &\multicolumn{2}{l}{Task 3 has no precedence and position d is currently available from time 0.}\\
    \\
    &\multicolumn{2}{l}{Hence, task 3 is available from time 0.}\\
    &\multicolumn{2}{l}{(b) UAV availability check}\\
    &\multicolumn{2}{l}{$\cdot$ UAV ready time ($rt$)}\\
    &\multicolumn{2}{l}{UAV1, UAV2, and UAV3 are ready (not performing any task) from time 0.}\\
    &\multicolumn{2}{l}{UAV1: 0; UAV2: 0; UAV3: 0}\\
    &\multicolumn{2}{l}{$\cdot$ Battery level \& recharge}\\
    &\multicolumn{2}{l}{Battery consumption for task execution is calculated}\\
    &\multicolumn{2}{p{0.95\dimexpr \columnwidth-3\arrayrulewidth-4\tabcolsep\relax}}{It includes flight towards start position ($s$), task processing time ($pt$), and flight towards nearest recharge station ($rs$).}\\
    &$UAVx$&: $[rt]+[s]+[pt]+[rs]$\\
    &UAV1&: 0+160+719+40 = 200\\
    &UAV2&: 0+160+719+40 = 200\\
    &UAV3&: 0+40+719+40 = 80\\
    &\multicolumn{2}{p{0.95\dimexpr \columnwidth-3\arrayrulewidth-4\tabcolsep\relax}}{Note: if the sum of UAV ready time and flight towards start position is less than task availability time, then they are replaced with it.}\\
    &\multicolumn{2}{p{0.95\dimexpr \columnwidth-3\arrayrulewidth-4\tabcolsep\relax}}{UAV1, UAV2, and UAV3 do not need any recharge because each battery consumption is still \textless 1200.}\\
    &\multicolumn{2}{p{0.95\dimexpr \columnwidth-3\arrayrulewidth-4\tabcolsep\relax}}{If recharge is required, then it will search a recharge station with the shortest round-trip flight.}\\
    \\
    &\multicolumn{2}{l}{Compare the earliest available time for each UAV}\\
    &UAV1&: 0+60 = 60\\
    &UAV2&: 0+60 = 60\\
    &UAV3&: 0+40 = 40\\
    &\multicolumn{2}{l}{$\therefore$ UAV3 is picked for task 3.}\\
    \hline\noalign{\smallskip}
    \multicolumn{3}{c}{
      \parbox{0.9\columnwidth}{
      \centering
      \includegraphics[scale=0.7]{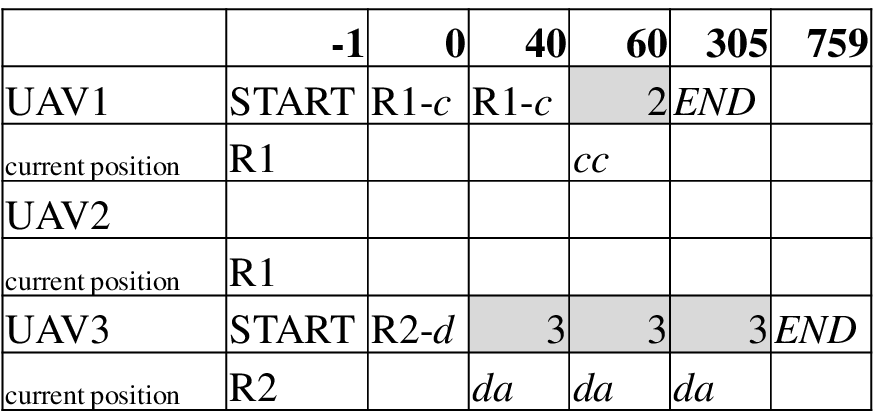}
      \figcaption{Output of step 2 of schedule creation heuristic}
      \label{fig:step2}}
    }\\
    \hline
    Step&\multicolumn{2}{c}{Description}\\
    \hline
    2 & \multicolumn{2}{l}{Task 2}\\
    &\multicolumn{2}{l}{(a) Check earliest available time of task 2}\\
    &\multicolumn{2}{l}{Task 2 is available from time 0.}\\
    \\
    &\multicolumn{2}{l}{(b) Check which UAV needs recharge before executing task 2}\\
    &UAV1&: 0+60+245+60 = 365\\
    &UAV2&: 0+60+245+60 = 365\\
    &UAV3&: 759+131+245+60 = 1195\\
    &\multicolumn{2}{l}{No UAV needs recharge.}\\
    \\
    &\multicolumn{2}{l}{Compare the earliest available time for each UAV}\\
    &UAV1&: 0+60 = 60\\
    &UAV2&: 0+60 = 60\\
    &UAV3&: 759+131 = 890\\
    &\multicolumn{2}{l}{$\therefore$ UAV1 is picked for task 2.}\\
    \hline
\end{longtable}

\begin{figure} [htp]
\centering
  \includegraphics[width=0.8\textwidth,keepaspectratio]{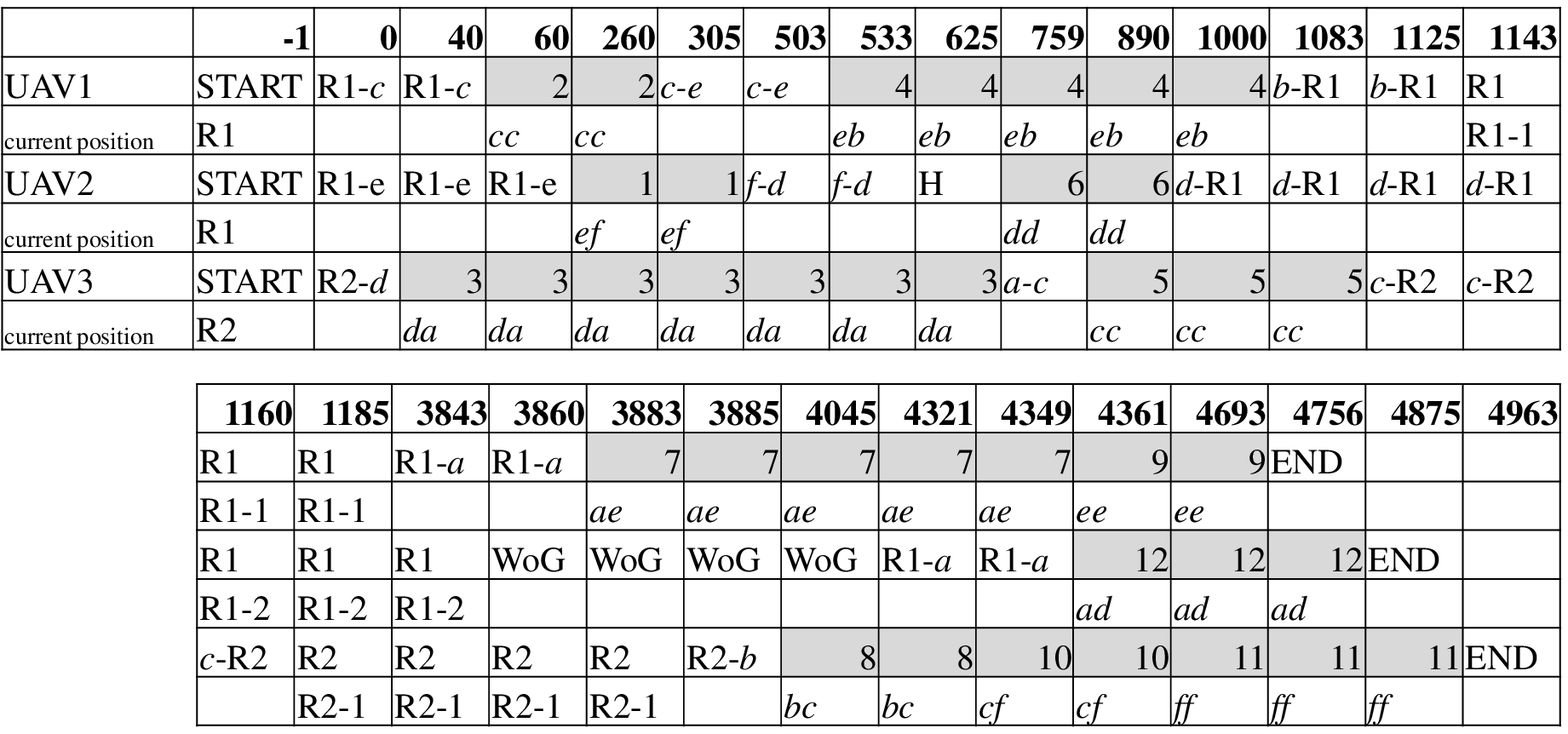}
\caption{Final output of schedule creation}
\label{fig:figure_final_schedule}
\end{figure}

After completing all 12 steps, a schedule of 12 tasks execution is obtained.
The final output of schedule creation from the given sequence is depicted in Figure \ref{fig:figure_final_schedule}.
The schedule has a makespan of 4963 seconds, which is called as fitness value of the particle (in the incorporated PSO).

The sample data (Table \ref{table:table_task} and \ref{table:table_flight_time}) being used in this section brings an example of task dataset which strongly displays the logic of Algorithm \ref{algo:algo_heuristic}. For further analysis and evaluation, a generated data will be used for tractability of producing the desired data volume.
Respectively, the mechanism of generating the data is also presented in the following section.

\section{Numerical Experiments}
\label{sec:sec_numexp}
To examine the behavior and performance of the proposed methodology, numerical experiments are conducted based on 3 task datasets.
Several different treatments are given during those experiments and explained in detail in Section \ref{sec:subsubsec_param_analysis} and \ref{sec:subsubsec_performance_eval}.
The proposed algorithm is coded on Java platform and the numerical experiments are conducted on an Intel Core i7 processor (2.9 GHz) with 32 GB of RAM.

\subsection{Data Generation}
Task dataset used in the experiment is generated based on a test flight conducted in the laboratory.
It is conducted to measure the speed of UAV movement in real-world indoor environment.
Based on the required test case described in \textit{task layer} (see Section \ref{sec:USIE}), several types of task are classified in Table \ref{table:table_task_type_complete}.
Single inspection captures an observation image of a certain area of interest; its processing time is 20--80 seconds.
Compound inspection captures multiple observation images of several areas of interest; its processing time is 100--200 seconds. Unlike single inspection, this task might contain flight action(s) between each image capturing.
Without such abstraction, the level of detail will cause a high number of steps in a solution; which obviously affects the computation time in finding one.
The next type of task is material handling. Material handling consists of pick-up, transport flight, and release. Its processing time is 30 seconds for each pick-up and release (30+30 = 60 seconds), while transport flight varies according to the origin and destination position.

\begin{table} [htp]
\caption{Task types}
\begin{center}
    \begin{tabular}{ l c }
    \hline\noalign{\smallskip}
    Task Type & Processing Time (second)\\
    \noalign{\smallskip}\hline\noalign{\smallskip}
    Single inspection & 20--80\\
    Compound inspection & 100--200\\
    Material handling & 60 $+$ flight\_time\\ \hline
    \end{tabular}
\end{center}
\label{table:table_task_type_complete}
\end{table}

During scheduling process, a task has five attributes attached: task identifier, origin position, destination position, processing time, and predecessor list (see Table \ref{table:table_task_att}). Task identifier is represented as a unique integer value, which means that no multiple tasks share the same task identifier. Origin position and destination position are represented as a unique string each, which acts as position name and position identifier simultaneously. Processing time is represented as an integer value where its execution time shall never exceed the battery capacity of UAV (Equation \eqref{eq:exec_time}). Predecessor list is represented as a set of integers which indicates identifiers of preceding tasks.
\begin{table} [htp]
\caption{Task attributes}
\begin{center}
    \begin{tabular}{ l  l }
    \hline\noalign{\smallskip}
    Attribute & Data Type \\
    \noalign{\smallskip}\hline\noalign{\smallskip}
    task identifier & integer\\
    origin position & string\\
    destination position & string\\
    processing time & integer\\
    predecessor list & integers\\ \hline
    \end{tabular}
\end{center}
\label{table:table_task_att}
\end{table}

\begin{flalign} \label{eq:exec_time}
  execution\_time = & preparation\_flight + processing\_time\\\nonumber
  & + towards\_recharge\_flight\\\nonumber
  execution\_time <= & UAV\_BATTERY\_CAPACITY
\end{flalign}

\begin{figure}
\centering
\begin{subfigure}{0.2\textwidth}
  \includegraphics[width=\linewidth,keepaspectratio]{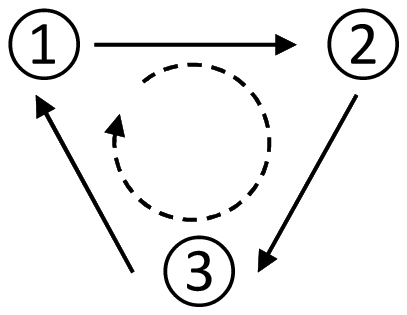}
\caption{Cyclic precedence relationship} \label{fig:pred_cyclic}
\end{subfigure}
\begin{subfigure}{0.4\textwidth}
  \includegraphics[width=\linewidth,keepaspectratio]{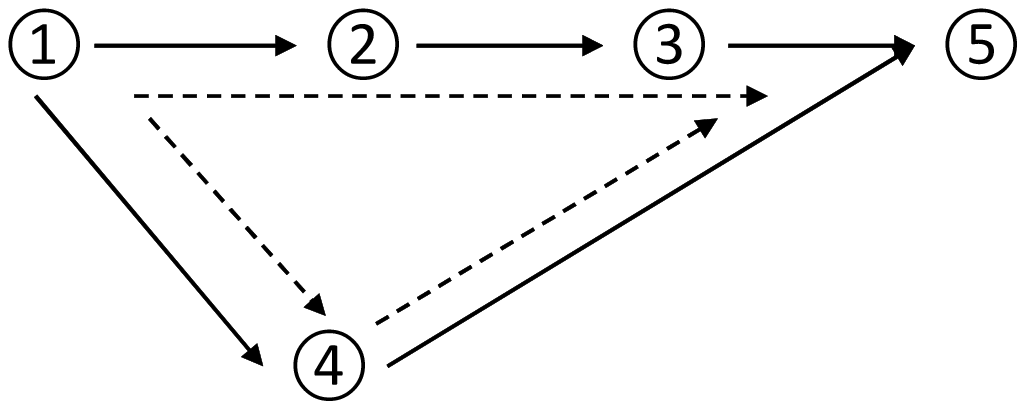}
\caption{Fine precedence relationship} \label{fig:pred_okay}
\end{subfigure}\\
\begin{subfigure}{0.2\textwidth}
  \includegraphics[width=\linewidth,keepaspectratio]{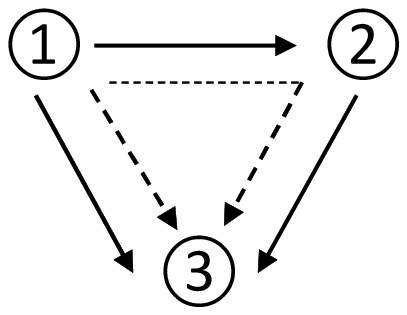}
\caption{Short redundant precedence relationship} \label{fig:pred_redundant_short}
\end{subfigure}
\begin{subfigure}{0.4\textwidth}
  \includegraphics[width=\linewidth,keepaspectratio]{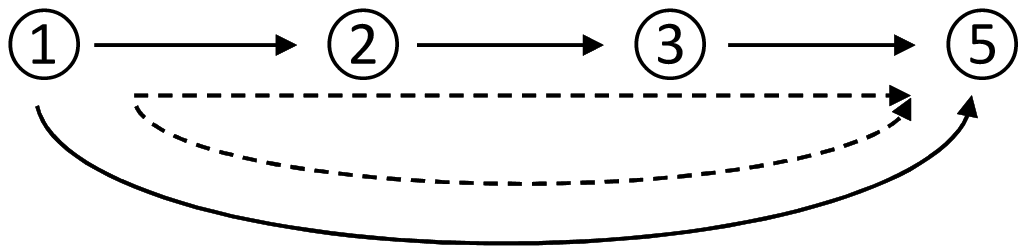}
\caption{Long redundant precedence relationship} \label{fig:pred_redundant_long}
\end{subfigure}
\caption{Possible precedence relationships in a task graph}
\label{fig:figure_precedence}
\end{figure}

Precedence relationships are built by generating predecessor list which may contain up to a certain maximum number of predecessors. The generation is random but yet conforms several characteristics as follows.
\begin{enumerate}[topsep=0pt]
  \item Non-cyclic precedence relationship \\
  A cyclic precedence relationship is produced when a task is directly or indirectly preceded by another one which awaits for that particular task. This condition of precedence relationship is depicted in Figure \ref{fig:pred_cyclic}.
  \item Non-redundant precedence relationship \\
  A redundant precedence relationship is produced when a task is both directly and indirectly preceded by another task simultaneously. This condition of precedence relationship is depicted in Figure \ref{fig:pred_redundant_short} and \ref{fig:pred_redundant_long}. For clarity, Figure \ref{fig:pred_redundant_long} is depicted as follows. Task 5 can only be executed if and only if task 1 and task 3 are completed. At a glance, there is no problem with this precedence relationships since it is not cyclic. However, task 3 can be executed if and only if task 2 is completed, which can be executed if and only if task 1 is completed. It is clear that even without the direct precedence relationship between task 1 and 5, task 5 can only be executed after task 1 is completed. Thus, a redundant relationship is created here.
\end{enumerate}

\subsection{Parameter Analysis \& Performance Evaluation}
In the proposed method, one granularity of data can be seen as a sequence. From a particular sequence, a schedule is created using a heuristic based on the proposed Earliest Available Time algorithm. The schedule is then evaluated through its makespan (the total time required for completing all tasks based on a schedule). The process is done iteratively and executed in a manner according to the PSO algorithm. To control the performance of this PSO algorithm, i.e., tendency of optimality level and convergence speed, a set of parameters need to be configured.
From the conducted experiments, performance from each set of parameters is obtained and evaluated in Section \ref{sec:subsubsec_performance_eval} to decide default values of parameters which most likely bring the best result for any case of task dataset.

In Section \ref{sec:subsubsec_param_analysis}, the following parameters are analyzed:
\begin{enumerate}[topsep=0pt]
  \item Number of initial population \\
  Initial particles in the initial population serves as the initial starting point of the search. The more varying starting points, spread at different locations throughout the search space, there is better chance in reaching the global optimum instead of trapped at a local one.
  \item Number of pairs in initial velocity \\
  To cover sufficient exploration area of the respective solution search space, one must adjust the number of pairs along with the escalation of the number of tasks. By doing so, a particle is capable to explore its surrounding search space and more likely find a local best particle in that area. The other way around, lower number of pairs in initial velocity minimizes the local exploration and force the search to rely more on the variety of particles in the initial swarm alone.
  \item Value of learning coefficients ($c_1$ and $c_2$) and velocity coefficients ($U_1$ and $U_2$) \\
  According to the Equation \eqref{eq:eq_velocity_update} for velocity update, $U_1$ and $U_2$ are fraction numbers randomly generated ranging from 0.0 to 1.0. Furthermore, because of $c_1$ and $c_2$ multiplied respectively with $U_1$ and $U_2$, they will control the search direction. $c_1U_1$ and $c_2U_2$ will decide the number of pairs obtained from distance of current particle to local and global best particle respectively. Since constant $c_1$ is set to be smaller than constant $c_2$, there will be a tendency of getting more pairs produced from the social part which contains the global best sequence. Consequently, all particles in the swarm are alerted and encouraged to move towards the global best particle, while also less encouraged to move towards its own self-obtained local best particle. The movement of all particles towards global particle indicates the action of convergence during the whole search process, while the movement of each particle towards its local best particle allows the swarm to still explore towards various other directions to get potentially better global best particle.
\end{enumerate}

\subsubsection{Parameter Analysis}
\label{sec:subsubsec_param_analysis}
\begin{figure} [htp]
\centering
\begin{subfigure}{0.3\textwidth}
  \includegraphics[width=\linewidth,keepaspectratio]{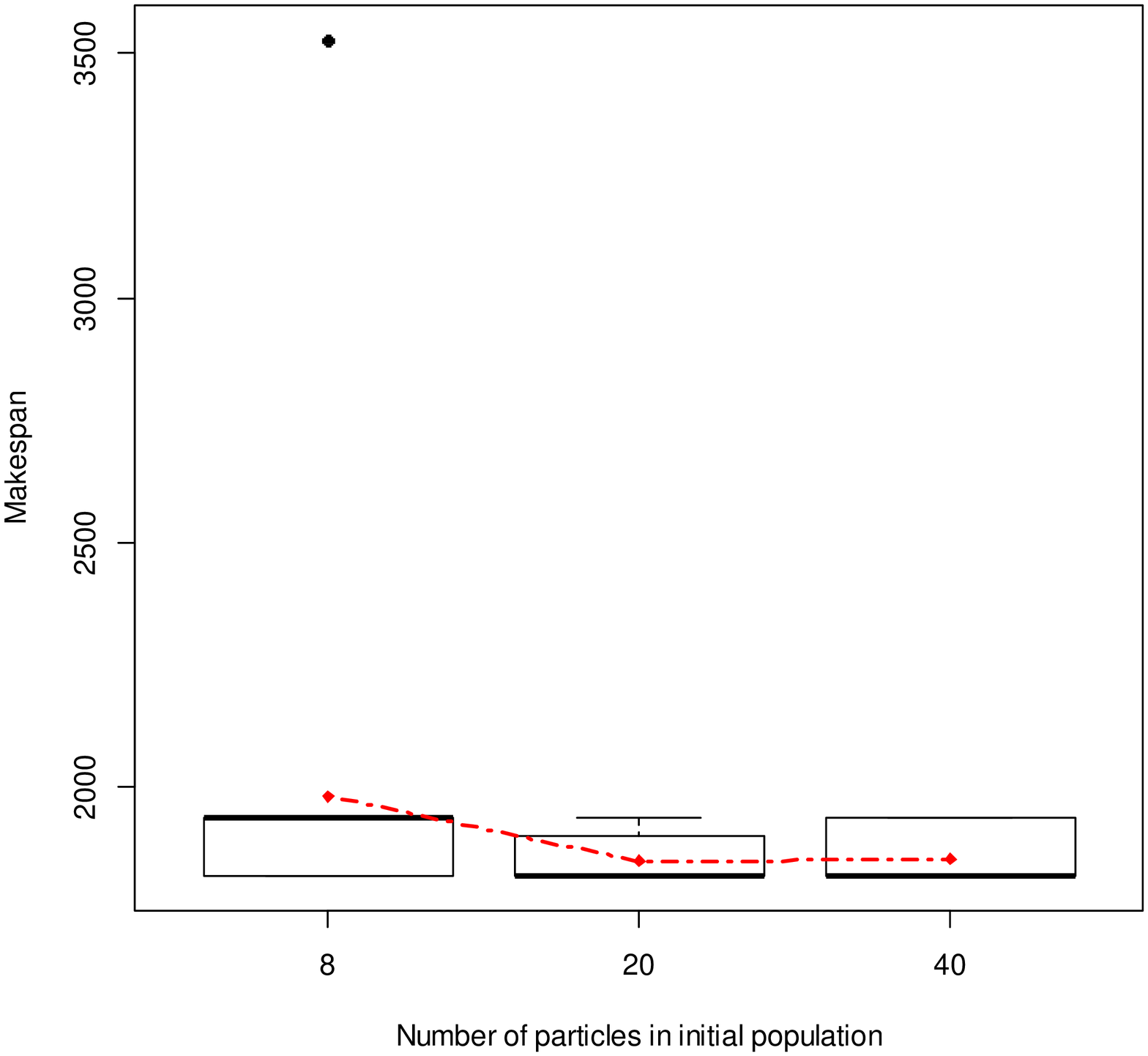}
\caption{$c_1$=1, $c_2$=1, No. of tasks=10} \label{fig:makespan1}
\end{subfigure}
\begin{subfigure}{0.3\textwidth}
  \includegraphics[width=\linewidth,keepaspectratio]{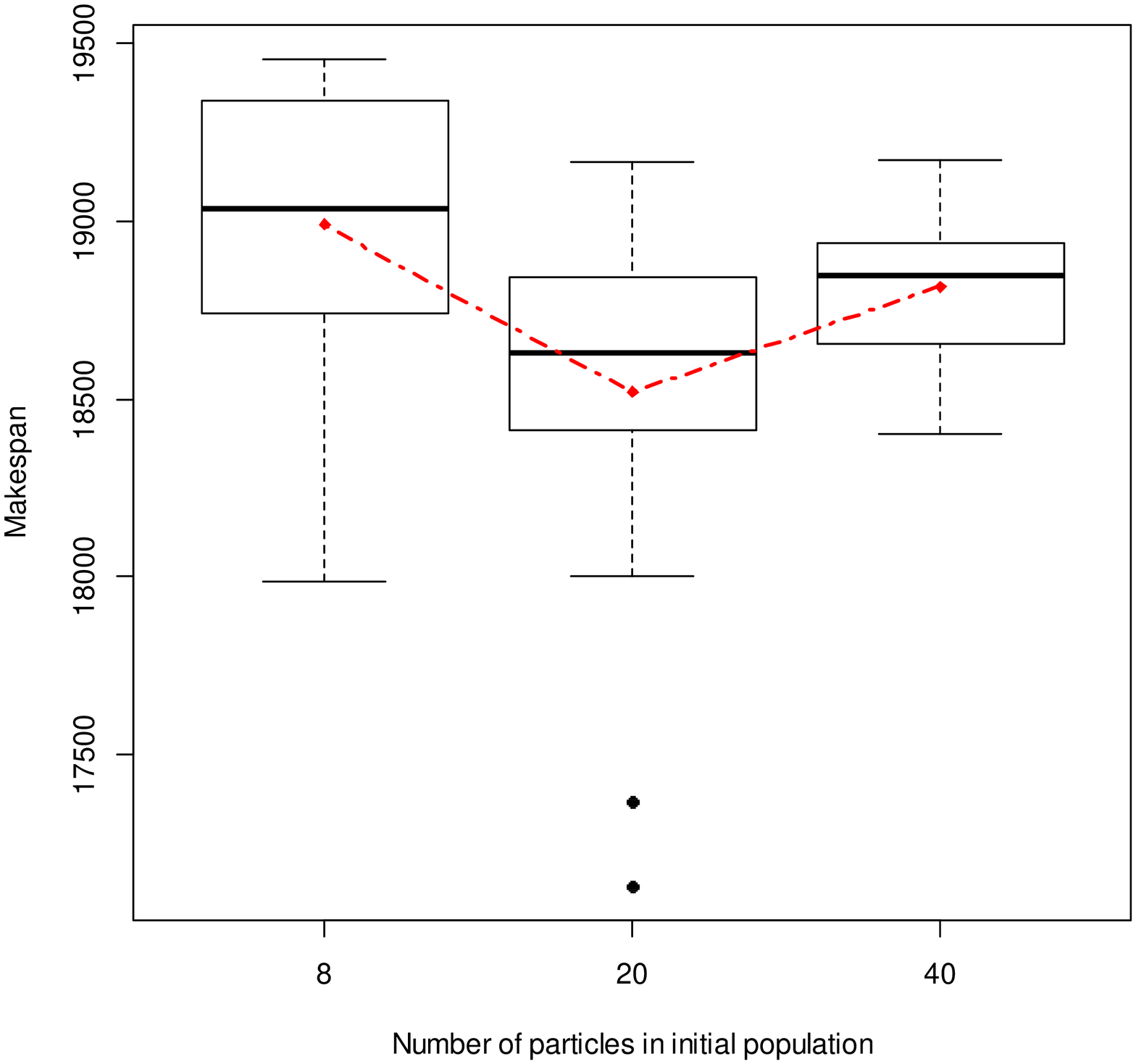}
\caption{$c_1$=1, $c_2$=1, No. of tasks=50} \label{fig:makespan3}
\end{subfigure}
\begin{subfigure}{0.3\textwidth}
  \includegraphics[width=\linewidth,keepaspectratio]{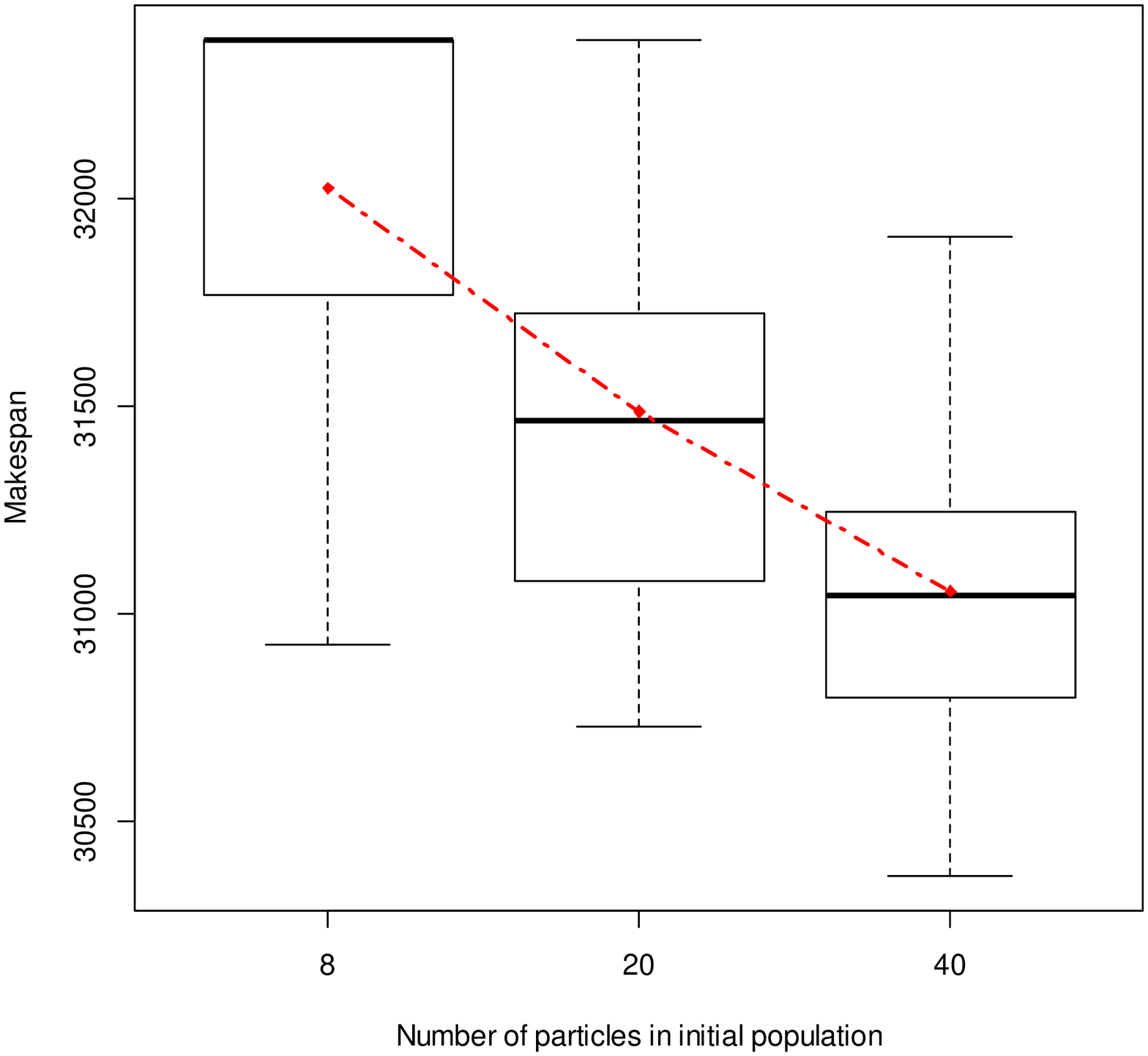}
\caption{$c_1$=1, $c_2$=1, No. of tasks=100} \label{fig:makespan5}
\end{subfigure}
\\
\begin{subfigure}{0.3\textwidth}
  \includegraphics[width=\linewidth,keepaspectratio]{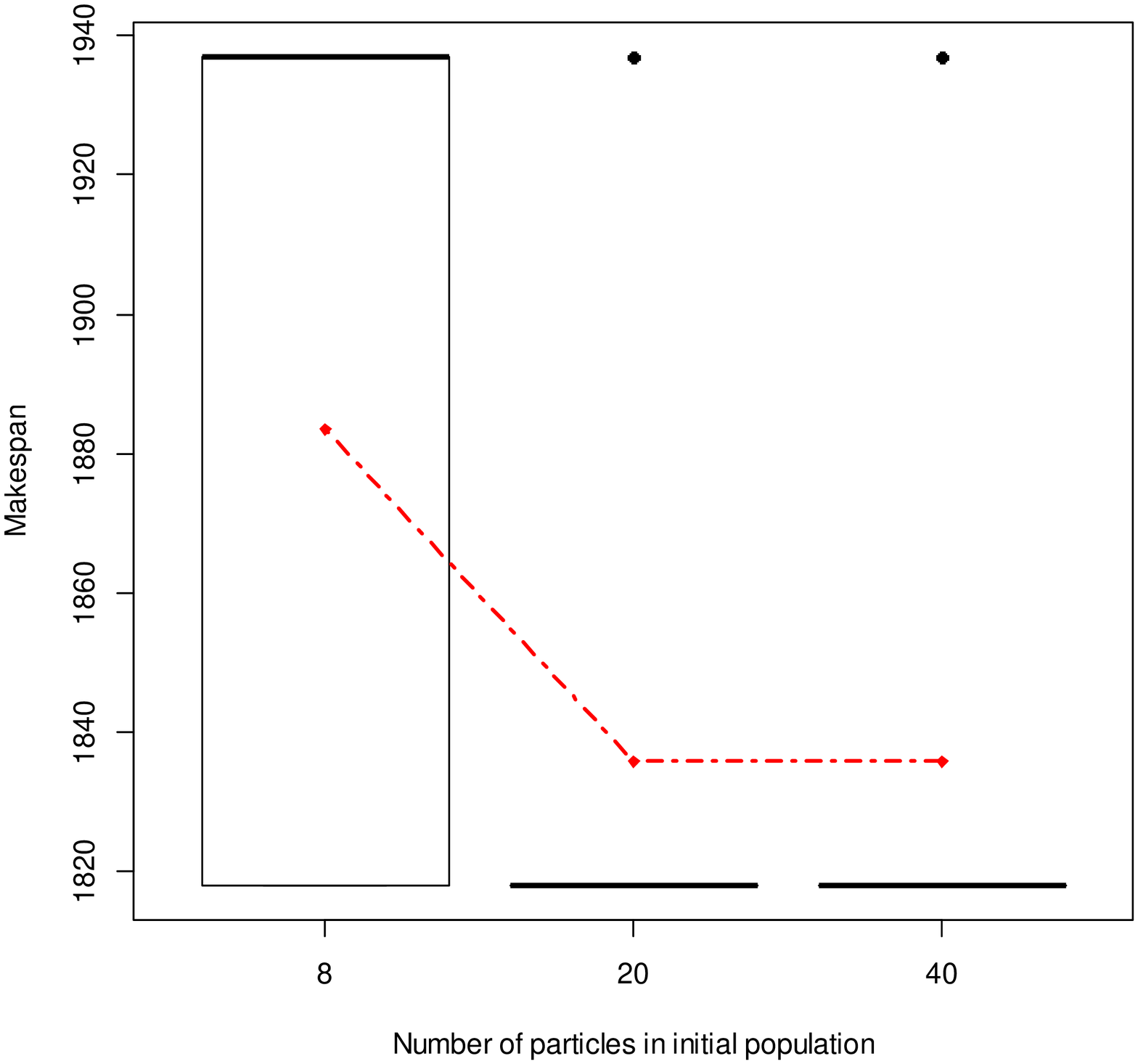}
\caption{$c_1$=1, $c_2$=2, No. of tasks=10} \label{fig:makespan2}
\end{subfigure}
\begin{subfigure}{0.3\textwidth}
  \includegraphics[width=\linewidth,keepaspectratio]{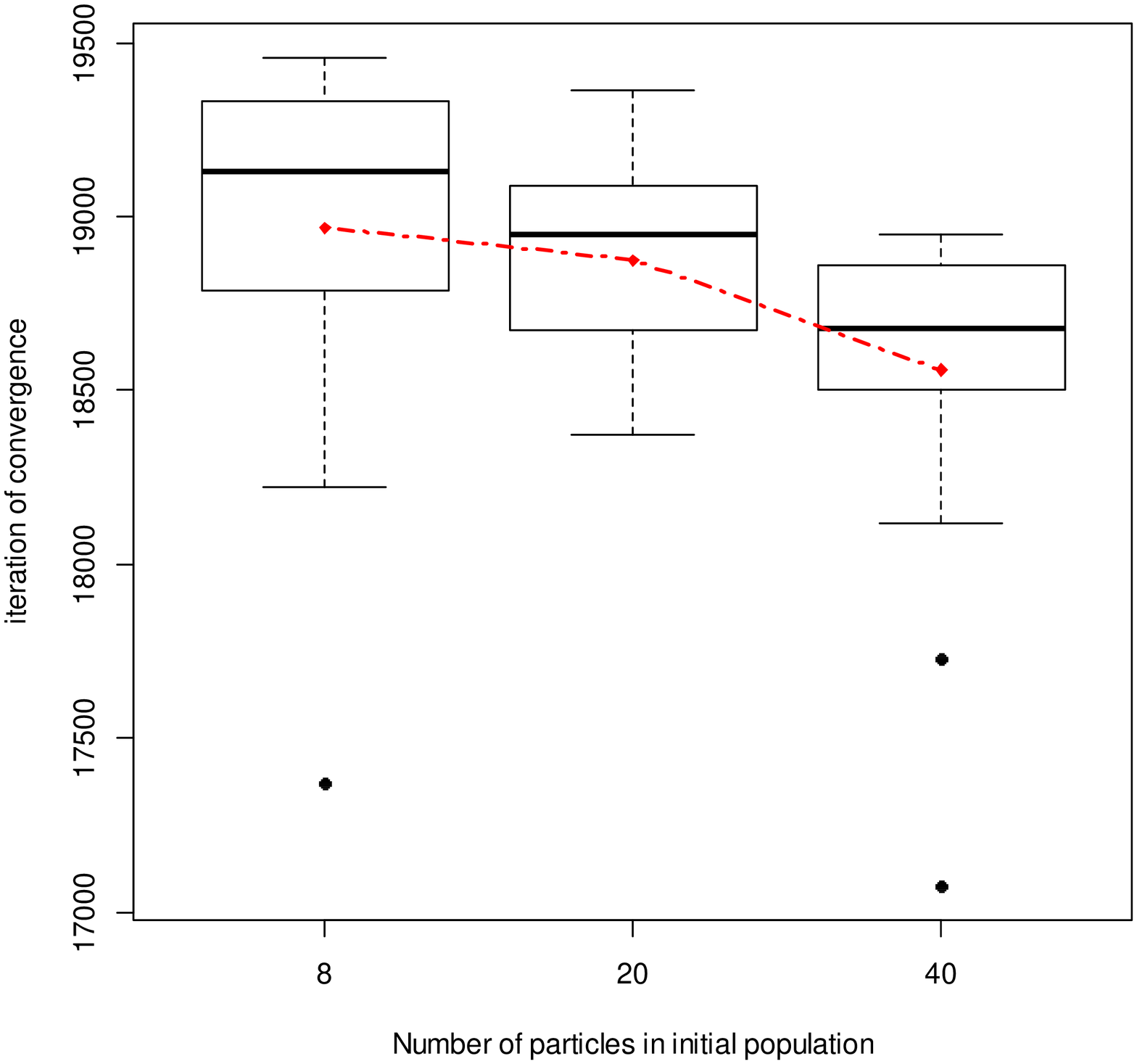}
\caption{$c_1$=1, $c_2$=2, No. of tasks=50} \label{fig:makespan4}
\end{subfigure}
\begin{subfigure}{0.3\textwidth}
  \includegraphics[width=\linewidth,keepaspectratio]{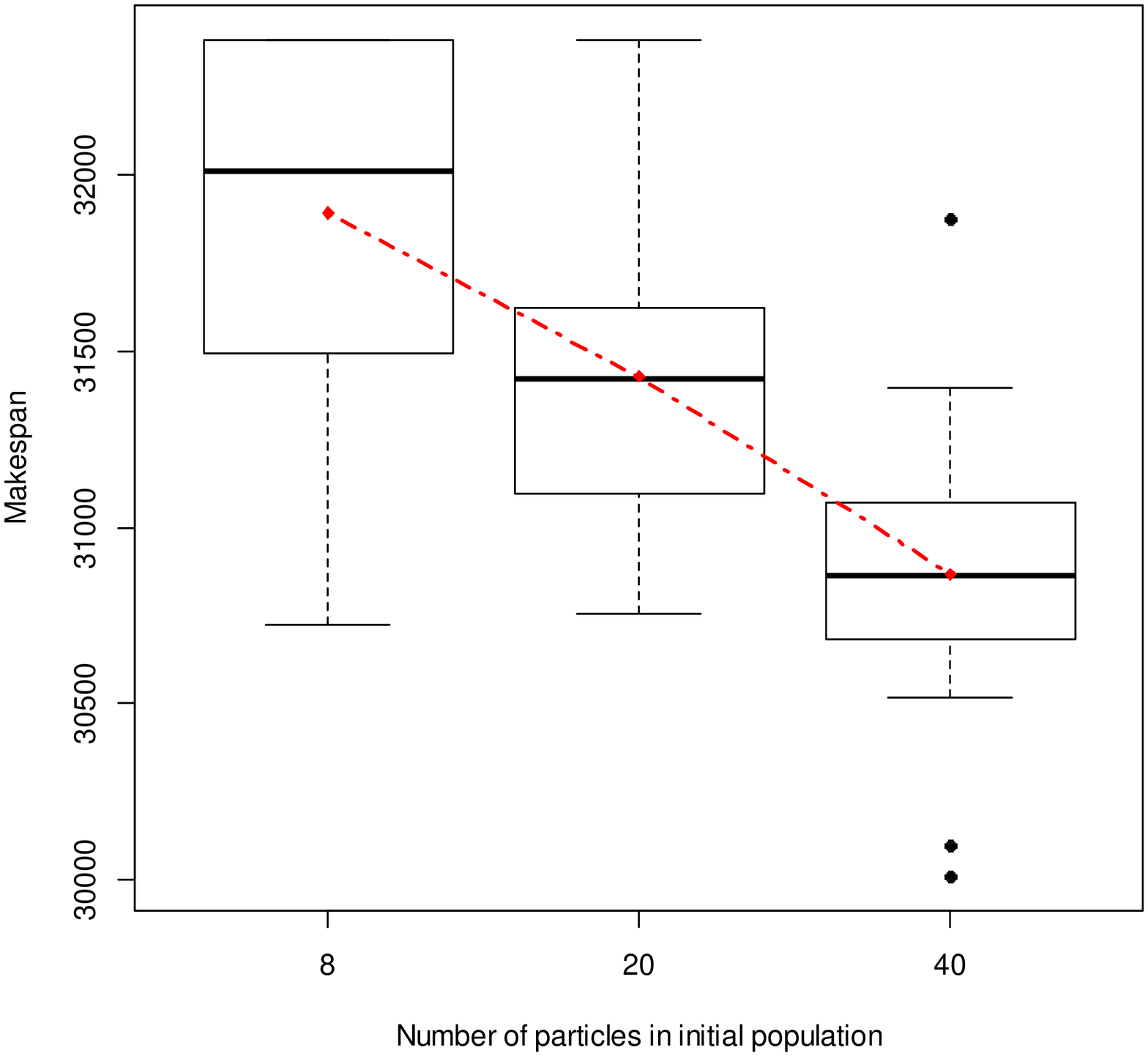}
\caption{$c_1$=1, $c_2$=2, No. of tasks=100} \label{fig:makespan6}
\end{subfigure}
\\
\begin{subfigure}{0.3\textwidth}
  \includegraphics[width=\linewidth,keepaspectratio]{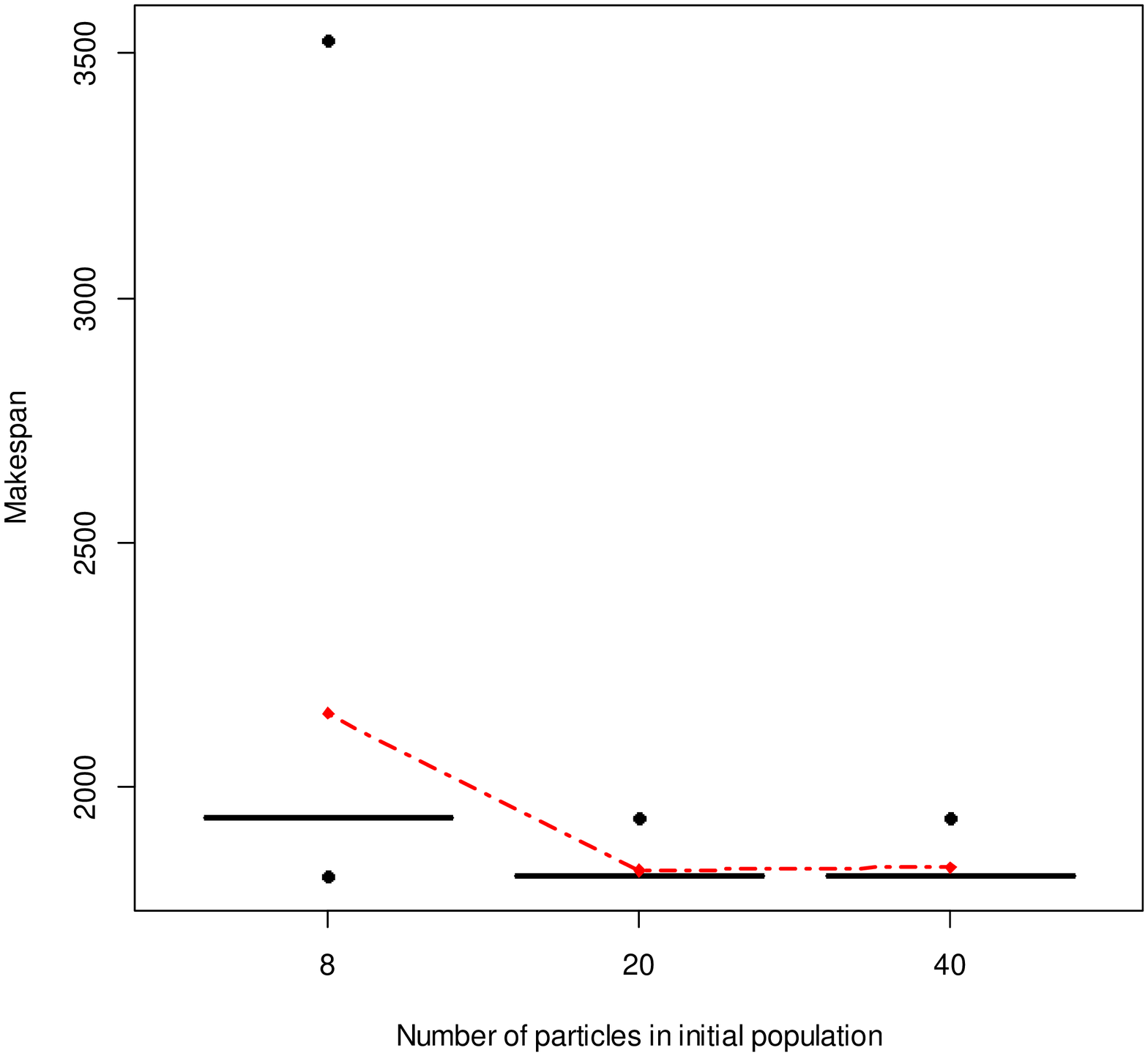}
\caption{$c_1$=2, $c_2$=1, No. of tasks=10} \label{fig:makespan7}
\end{subfigure}
\begin{subfigure}{0.3\textwidth}
  \includegraphics[width=\linewidth,keepaspectratio]{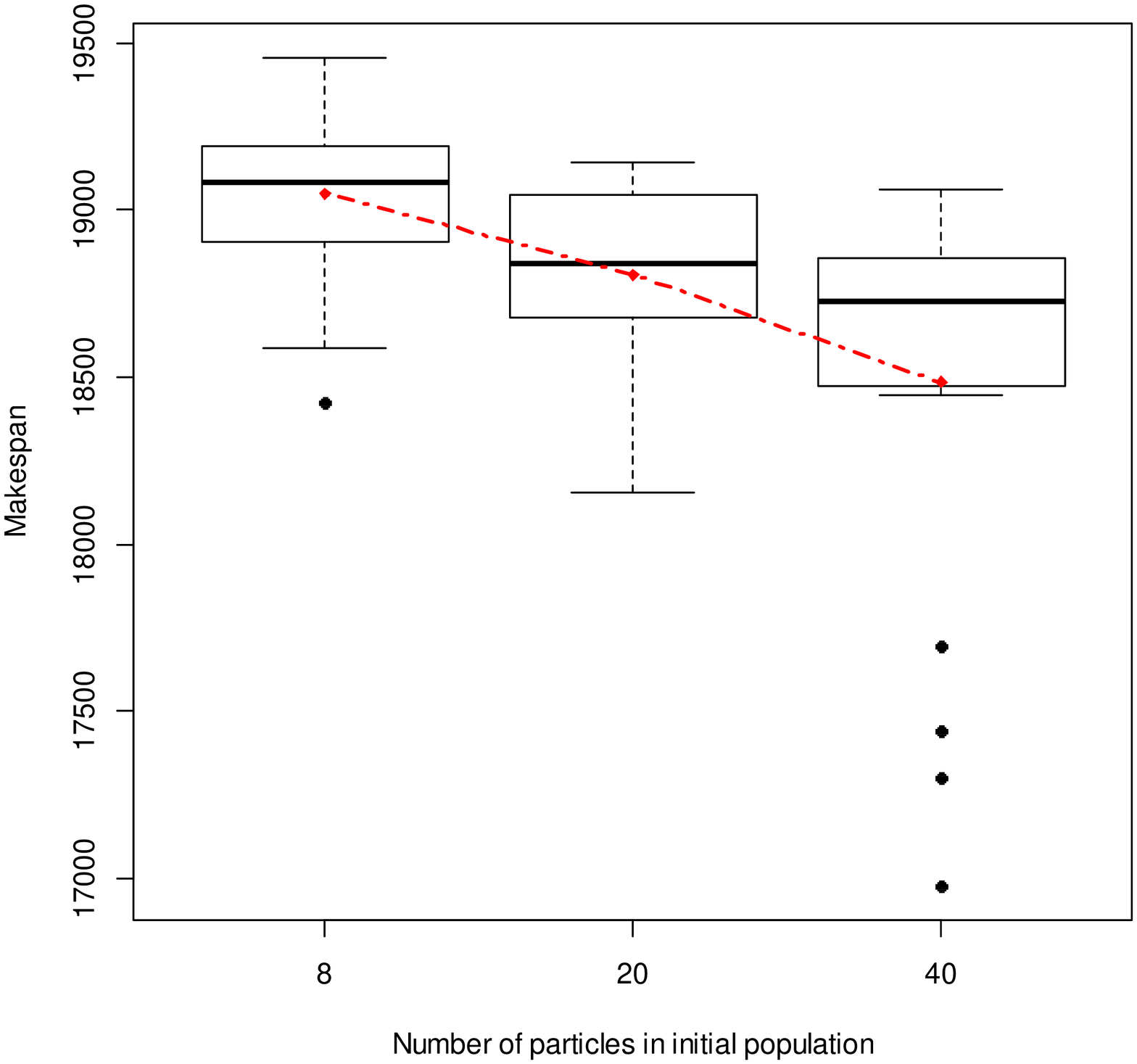}
\caption{$c_1$=2, $c_2$=1, No. of tasks=50} \label{fig:makespan8}
\end{subfigure}
\begin{subfigure}{0.3\textwidth}
  \includegraphics[width=\linewidth,keepaspectratio]{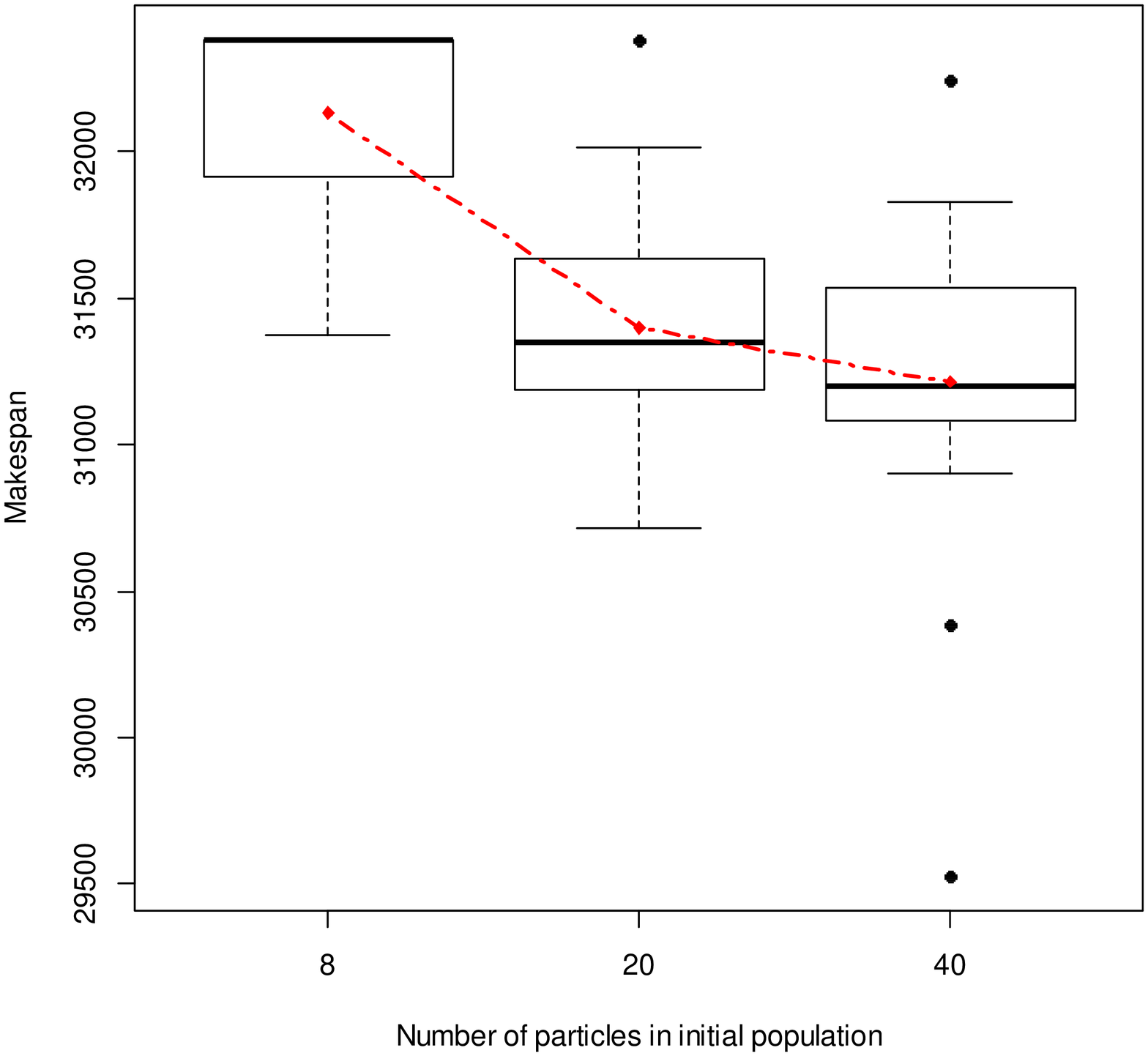}
\caption{$c_1$=2, $c_2$=1, No. of tasks=100} \label{fig:makespan9}
\end{subfigure}
\\
\begin{subfigure}{0.3\textwidth}
  \includegraphics[width=\linewidth,keepaspectratio]{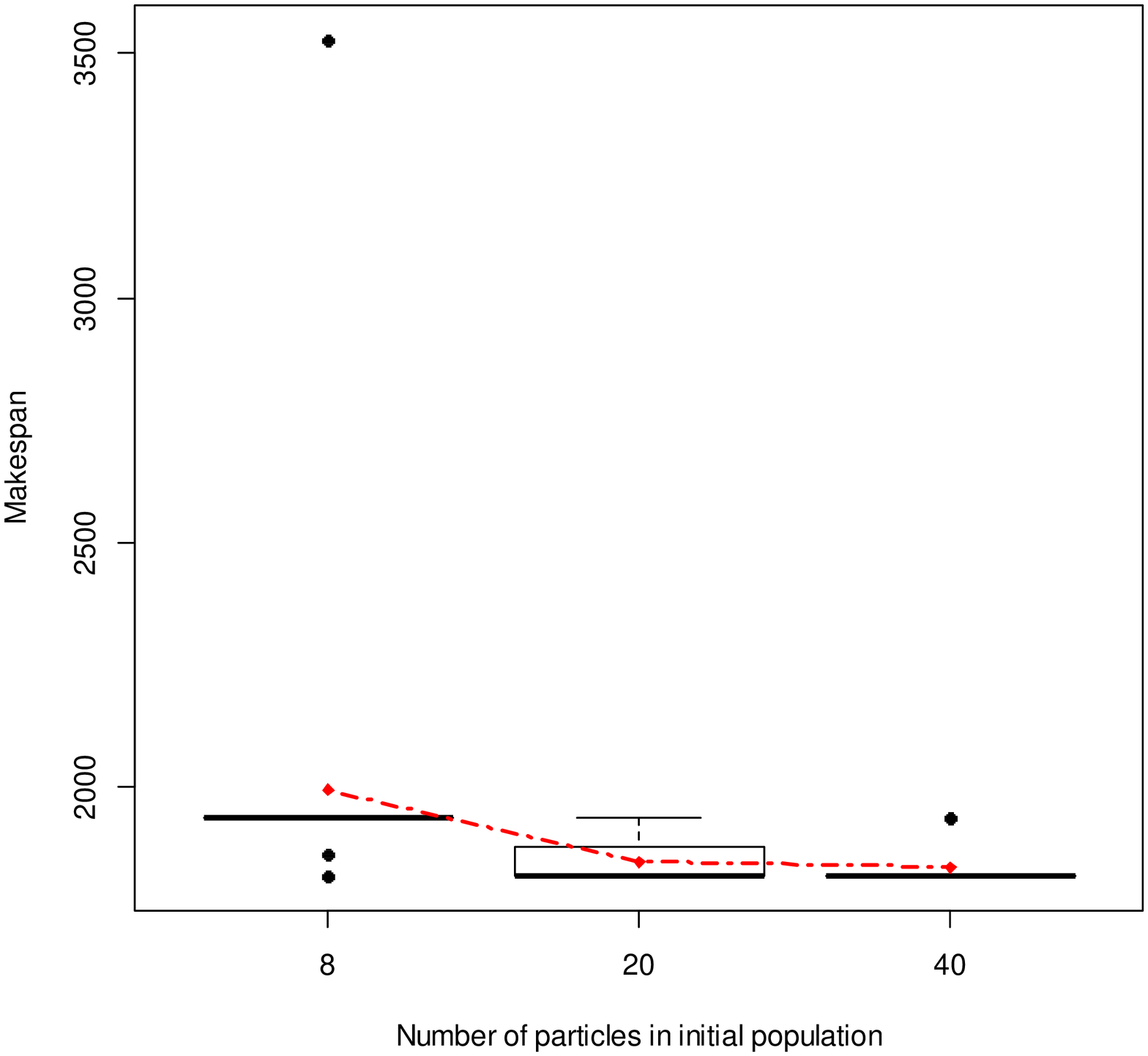}
\caption{$c_1$=2, $c_2$=2, No. of tasks=10} \label{fig:makespan10}
\end{subfigure}
\begin{subfigure}{0.3\textwidth}
  \includegraphics[width=\linewidth,keepaspectratio]{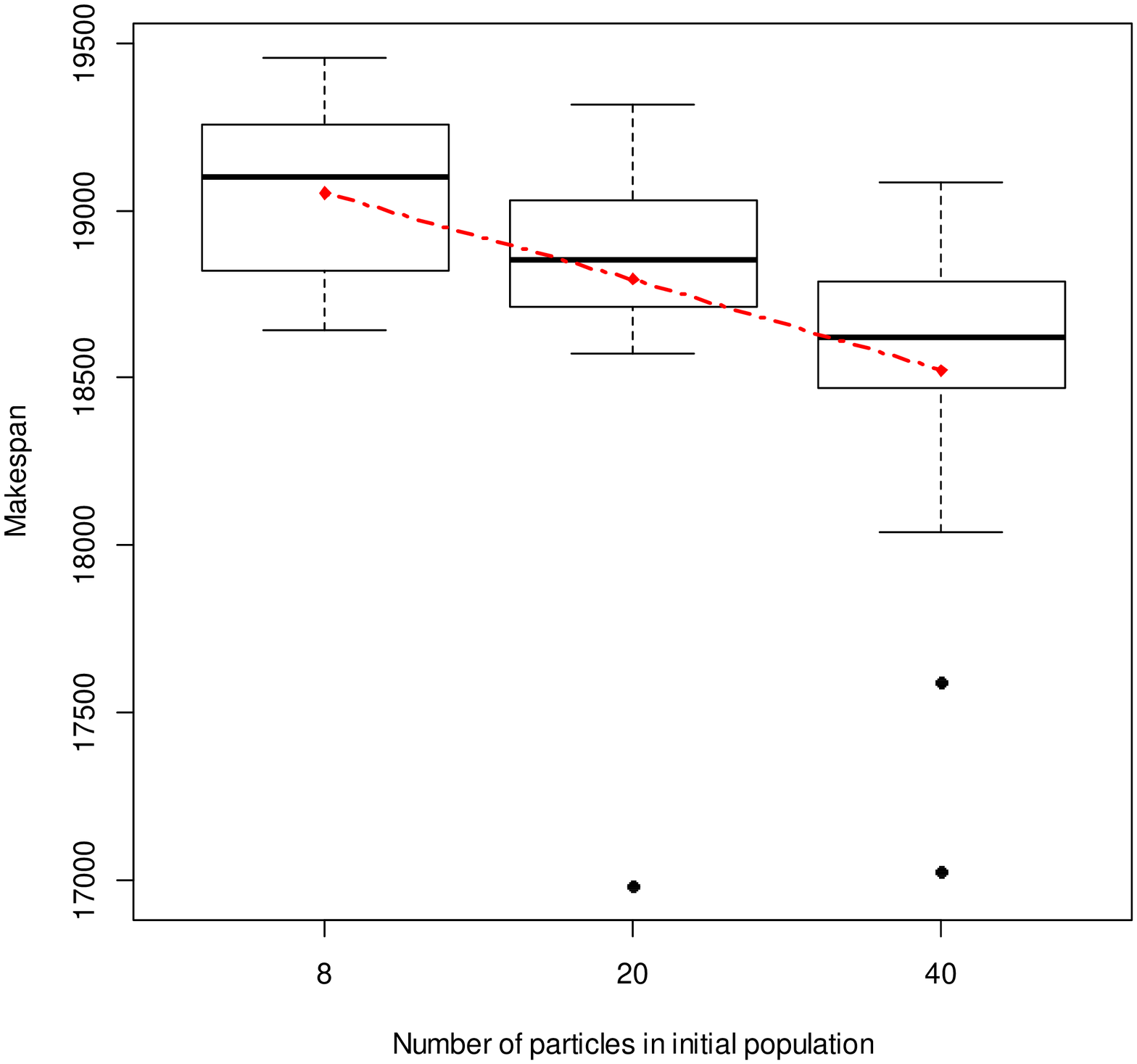}
\caption{$c_1$=2, $c_2$=2, No. of tasks=50} \label{fig:makespan11}
\end{subfigure}
\begin{subfigure}{0.3\textwidth}
  \includegraphics[width=\linewidth,keepaspectratio]{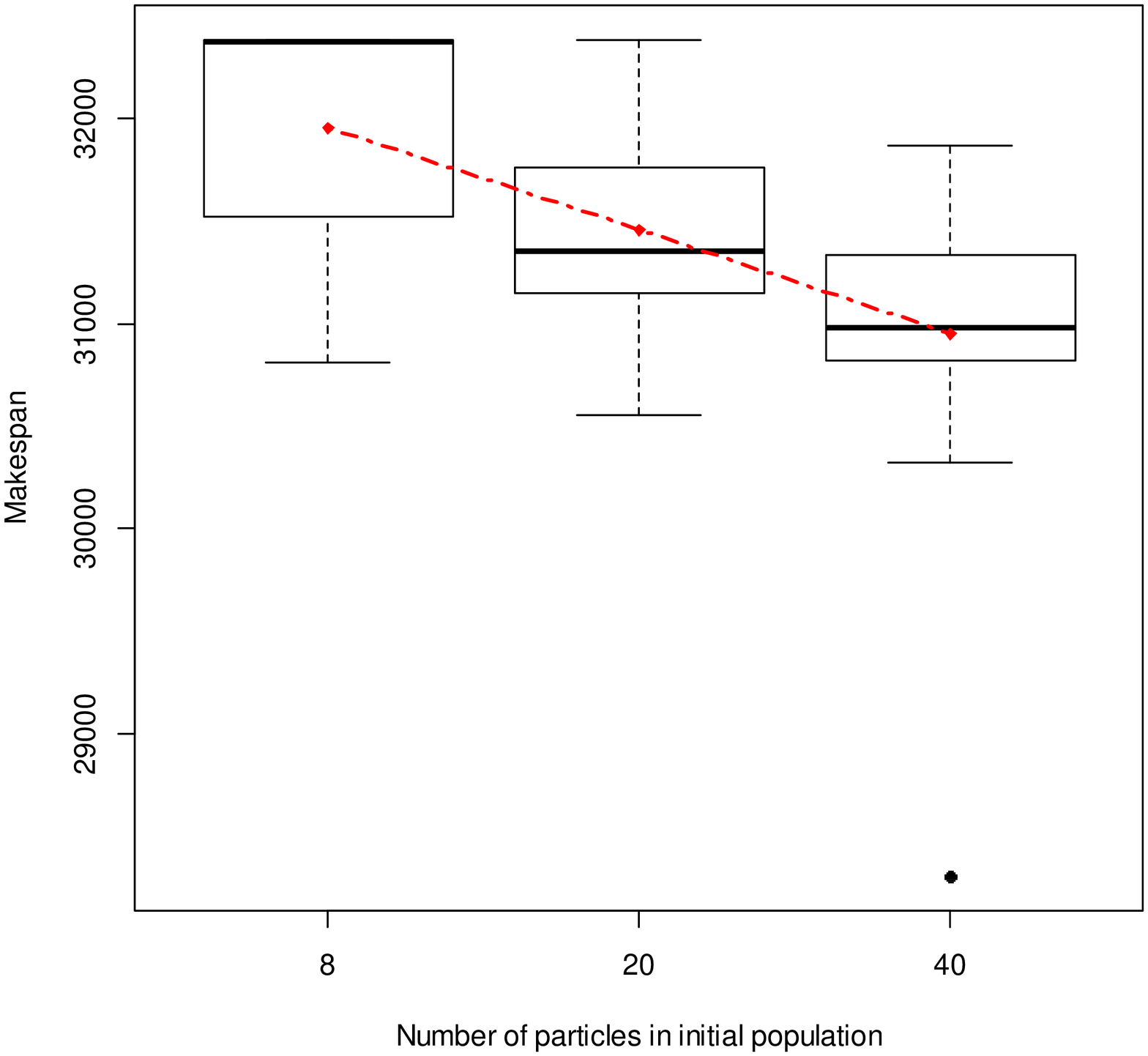}
\caption{$c_1$=2, $c_2$=2, No. of tasks=100} \label{fig:makespan12}
\end{subfigure}
\caption{Learning coefficients ($c_1$ and $c_2$) analysis}
\label{fig:figure_makespan}
\end{figure}

Three different task datasets are used in the experiment, each consists of 12, 50, and 100 tasks, operated by 3 UAVs.
Figure \ref{fig:figure_makespan} depicts makespans of generated schedules based on various combinations of parameters: $c_1$, $c_2$, and number of initial particles on 3 different task datasets. Each graph, for instance Figure \ref{fig:makespan1}, presents a result of 20 experimental runs.
Systematically, the granularity of the experimental run is explained as follows.
\begin{enumerate}[topsep=0pt]
\item There are 4 combinations of $c_1$ and $c_2$ treatment.
\item Each combination of $c_1$ and $c_2$ is applied on 3 task dataset: 10, 50, and 100 tasks.
\item Each task dataset is treated with 3 different number of initial particles: 8, 20, and 40 particles
\item Each treatment of number of initial particles is run for 20 times.
\end{enumerate}
Consequently, there are: $20 \text{x} 3 \text{x} 3 \text{x} 4 = 720$ runs (and 720 makespans respectively) in total.

From the depicted results in Figure \ref{fig:figure_makespan}, several characteristic can be drawn as follows.
\begin{enumerate}[topsep=0pt]
  \item Makespan decreases (gets better) as the number of population increases. \\
  We have three treatments for the number of initial particles: 8, 20, and 40. The first treatment is 8 initial particles because there are 8 rules of particle generation for initial swarm in Table \ref{table:table_heuristic_rules}, where each rule produces one particle. Then the number of particles is then doubled to make a significant difference with the previous treatment; it is roughly rounded up to 20. Afterwards, the number of initial particles is doubled again to 40. From those 3 treatments, one can find that as the number of initial particles increased from 8, 20, to 40, the makespan of the obtained solution decreases. This characteristic tends to hold true for all combinations of parameters $1 \leq c_1 \leq 2$ and $1 \leq c_2 \leq 2$. It is deduced that the best suited number of initial particles is 40.
  \item Makespan lean towards the lowest level as parameter $c_2$ is set to be greater than $c_1$ \\
  Figure \ref{fig:overall_makespan_100tasks} depicts the overall (representing results from 8, 20, and 40 initial particles) makespans from four combinations of $c_1$ and $c_2$ values. Each overall makespan is obtained through applying local polynomial regression fitting \cite{cleveland1992} (with $span=0.75, degree=2$) on the makespan data from 60 runs; which consists of equal running portions of experiment with 8, 20, and 40 initial particles (20 runs each). Makespan line of $c_1=1$, $c_2=2$ is shown to be lower than others most of the time.
  This condition is explained by Equation \eqref{eq:eq_position_update} where $c_1$ is correlated with the determination level of the particle to learn from its local best particle, while $c_2$ is correlated with the determination level of the particle to learn from the global best particle.\\ In this manner, during the iterations of the search, every particle has the determination to learn more from a particle which is recalled as the best one by all particles in the swarm.
  \begin{figure}[htp]
    \centering
    \includegraphics[width=0.7\textwidth,keepaspectratio]{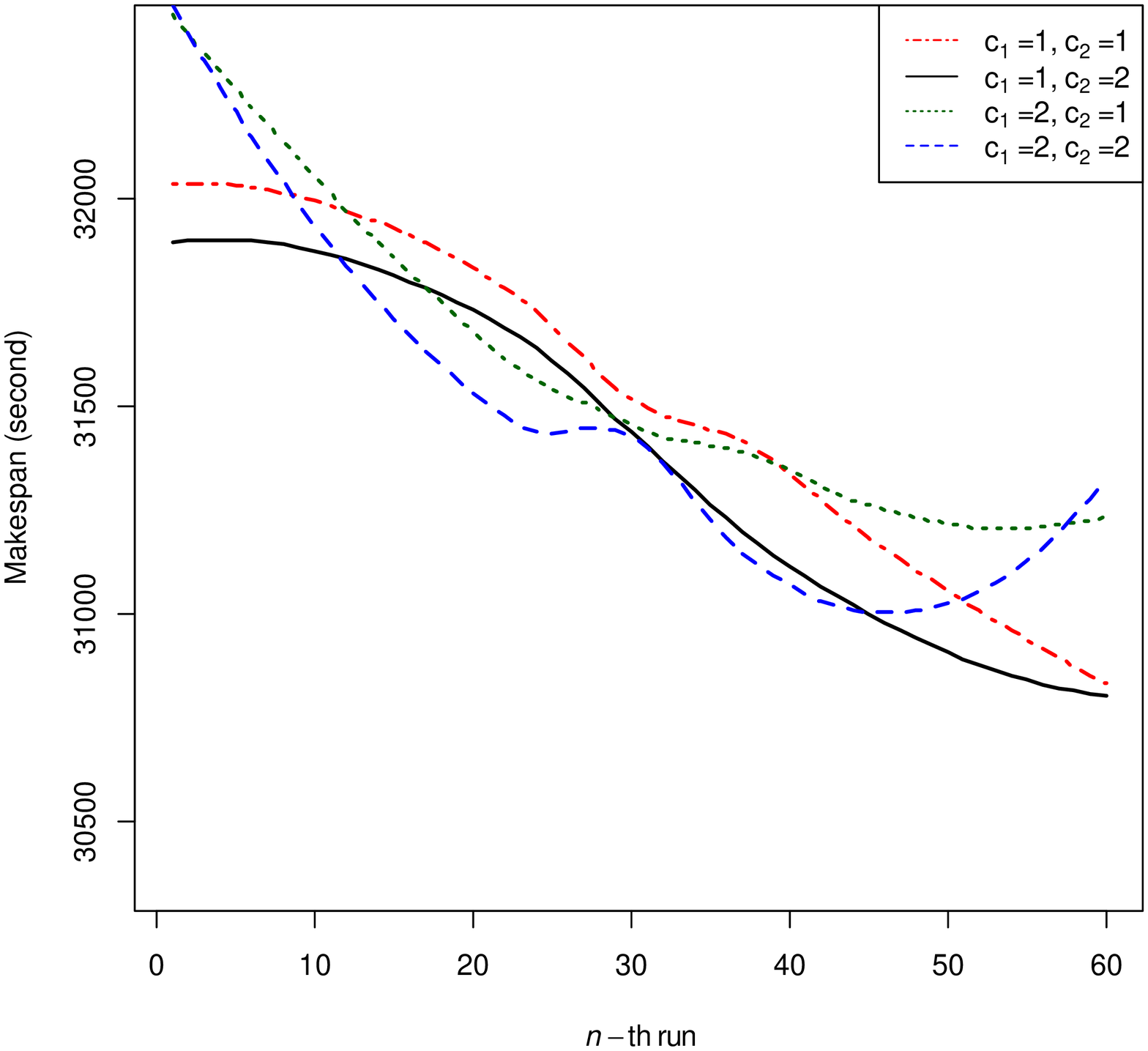}
    \caption{Overall makespans on 100 tasks} \label{fig:overall_makespan_100tasks}
  \end{figure}
  \item Makespan tends to converge at the same convergence point when the number of problem is small\\
  Makespan acts as fitness value of a solution. In Figure \ref{fig:makespan1}, \ref{fig:makespan4}, \ref{fig:makespan7}, \ref{fig:makespan10}, it is shown that among multiple distinct runs, a relatively common makespan is found. That means, those different runs have relatively common convergence point of solution. It is an extreme phenomenon on the benefit of having highly spread initial particles in the beginning of the search. Moreover, since the number of possible of solution is relatively not that high, nearly all local optimum areas are visited, leading to a high chance of finding a global optimum solution which is obviously the same for all runs.
\end{enumerate}

Consequently a trade-off relationship of each parameter is depicted in Table \ref{table:table_param_tradeoff}.
\begin{table}[htp]
\caption{Trade-off relationships among parameters}
\begin{center}
    \begin{tabulary}{\textwidth}{C | L | L}
    \hline\noalign{\smallskip}
    \multirow{2}{*}{Param.} & \multicolumn{2}{c}{Treatment Effect}\\ \cline{2-3}
    & \multicolumn{1}{c}{Increased} & \multicolumn{1}{|c}{Decreased}\\
    \noalign{\smallskip}\hline\noalign{\smallskip}
    $c_1$ & more pairs are derived from local best schedule sequence; it allows the search to step over a local optima and possibly find a global optimum schedule &
    less pairs are derived from local best schedule sequence; it discourages exploration towards various direction in the solution space\\ \hline
    $c_2$ & more pairs are derived from global best schedule sequence; it drives the search quickly towards the overall best schedule so far &
    less pairs are derived from the global best schedule; it slows down the convergence speed and allow each individual particle to have more room for exploration\\ \hline
    $U_1$ & \multicolumn{2}{c}{enhance the behavior of increased/decreased $c_1$}\\ \hline
    $U_2$ & \multicolumn{2}{c}{enhance the behavior of increased/decreased $c_2$}\\ \hline
    number of initial particles & more starting points are spread-out in the solution space, which reduce the probability to fall to a local optima, while promoting a high quality feasible solution in a quick time & less scattered starting points are established and less pulling-each-other-out among particles; it promotes a very quick convergence to the search\\
    \hline
\end{tabulary}
\end{center}
\label{table:table_param_tradeoff}
\end{table}

\subsubsection{Performance Evaluation}
\label{sec:subsubsec_performance_eval}
\begin{figure} [htp]
\centering
\begin{subfigure}{0.3\textwidth}
  \includegraphics[width=\linewidth,keepaspectratio]{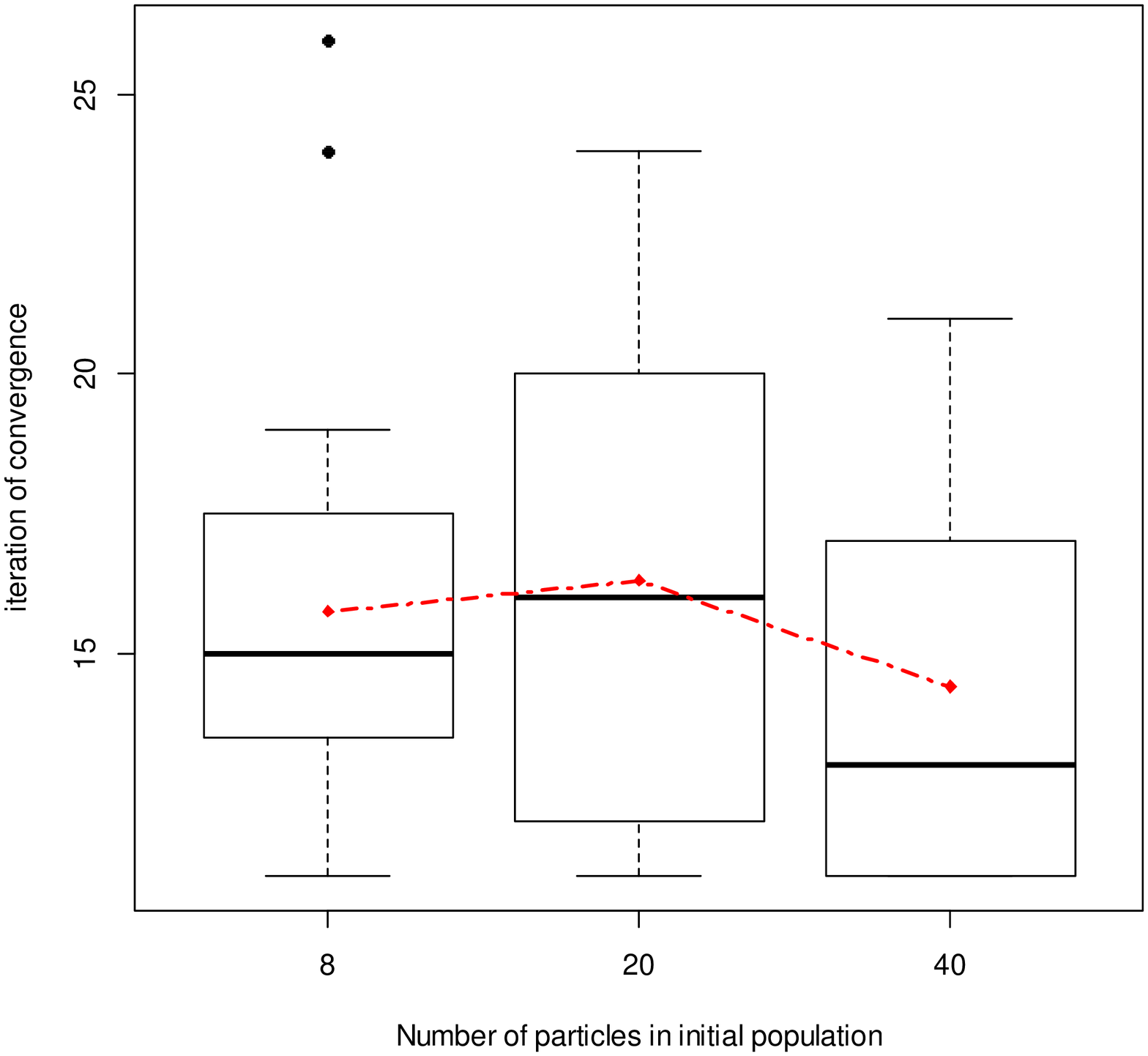}
\caption{$c_1$=1, $c_2$=1, No. of tasks=10} \label{fig:convergence1}
\end{subfigure}
\begin{subfigure}{0.3\textwidth}
  \includegraphics[width=\linewidth,keepaspectratio]{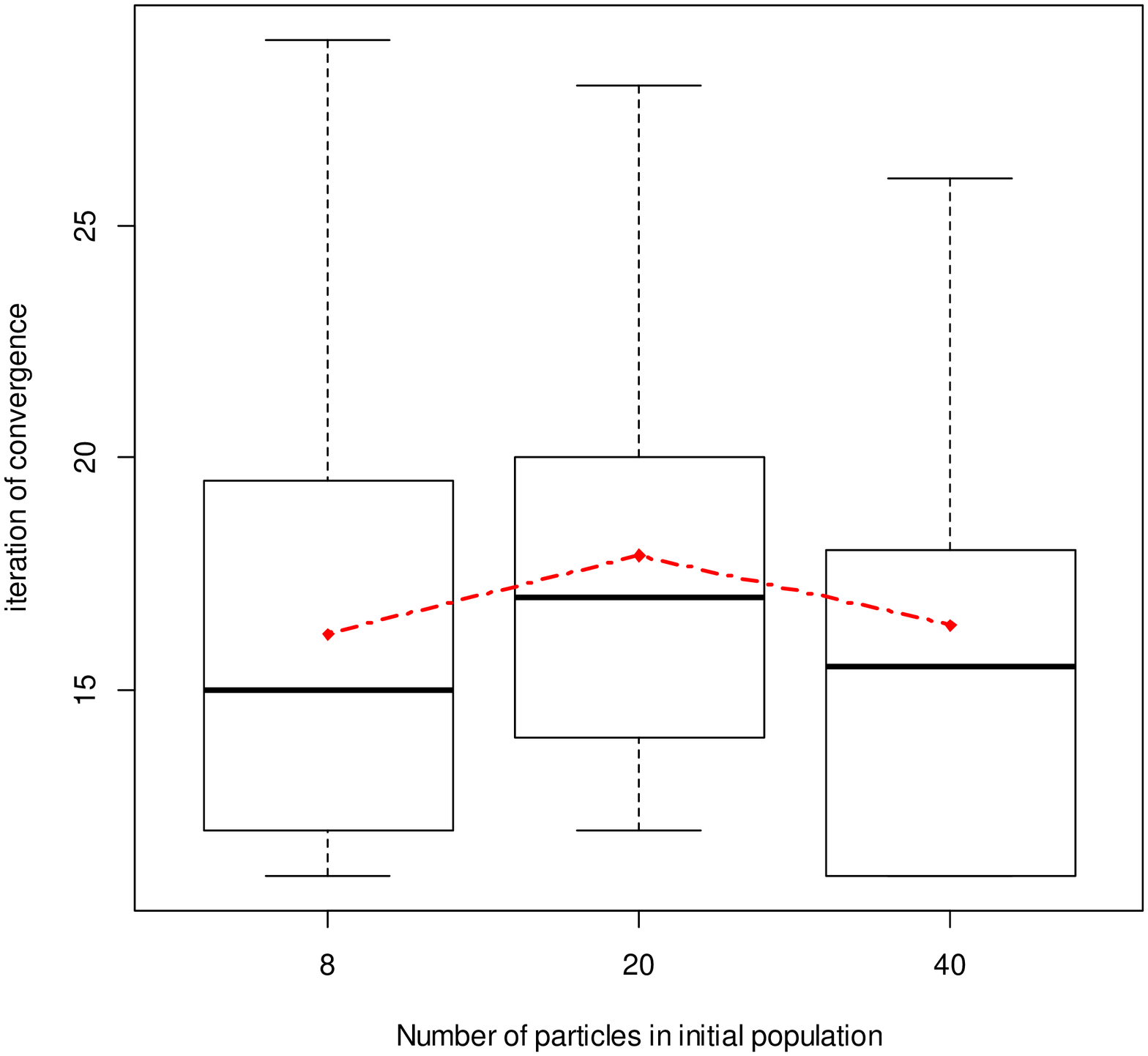}
\caption{$c_1$=1, $c_2$=1, No. of tasks=50} \label{fig:convergence2}
\end{subfigure}
\begin{subfigure}{0.3\textwidth}
  \includegraphics[width=\linewidth,keepaspectratio]{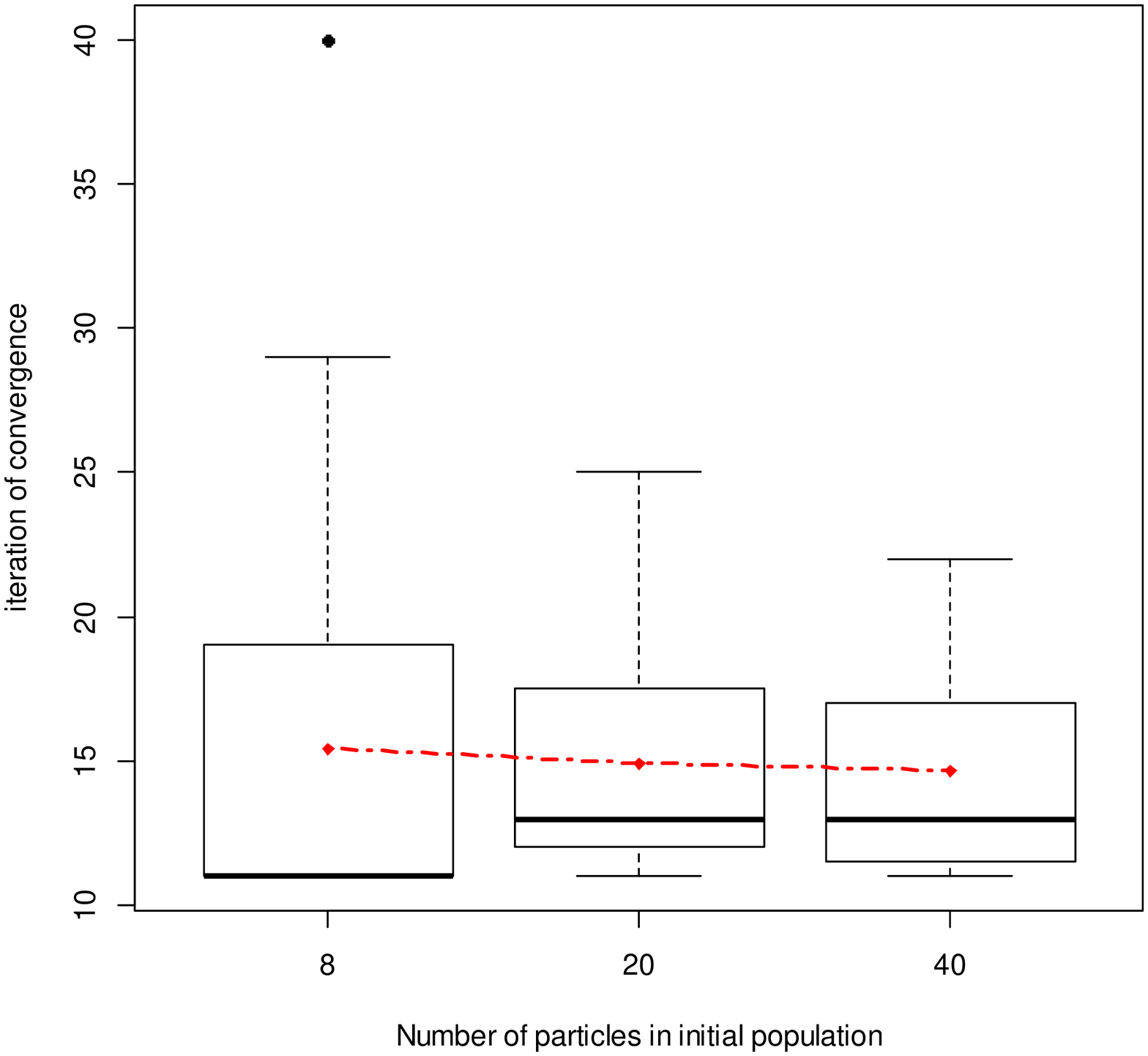}
\caption{$c_1$=1, $c_2$=1, No. of tasks=100} \label{fig:convergence3}
\end{subfigure} \\
\begin{subfigure}{0.3\textwidth}
  \includegraphics[width=\linewidth,keepaspectratio]{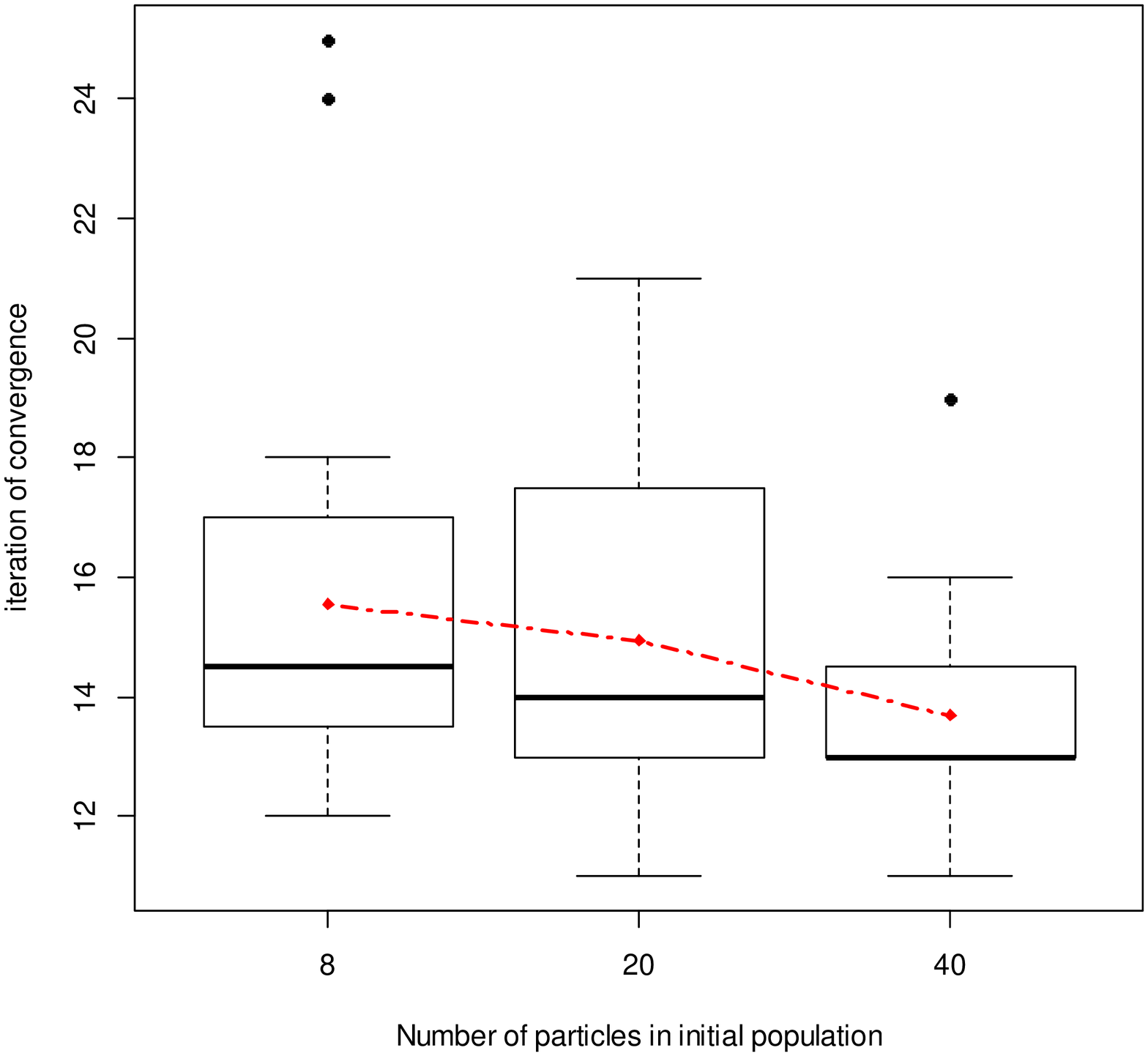}
\caption{$c_1$=1, $c_2$=2, No. of tasks=10} \label{fig:convergence4}
\end{subfigure}
\begin{subfigure}{0.3\textwidth}
  \includegraphics[width=\linewidth,keepaspectratio]{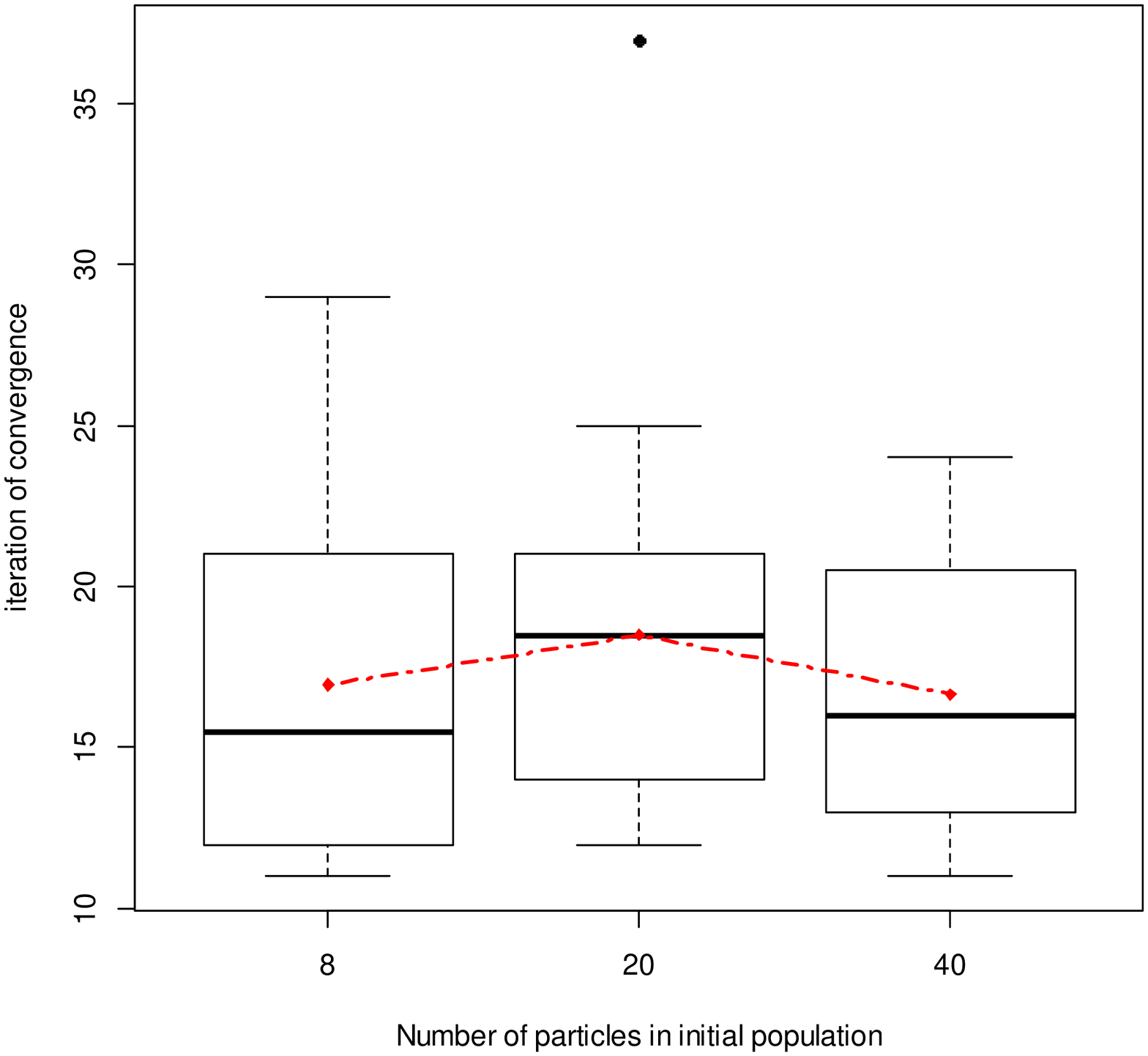}
\caption{$c_1$=1, $c_2$=2, No. of tasks=50} \label{fig:convergence5}
\end{subfigure}
\begin{subfigure}{0.3\textwidth}
  \includegraphics[width=\linewidth,keepaspectratio]{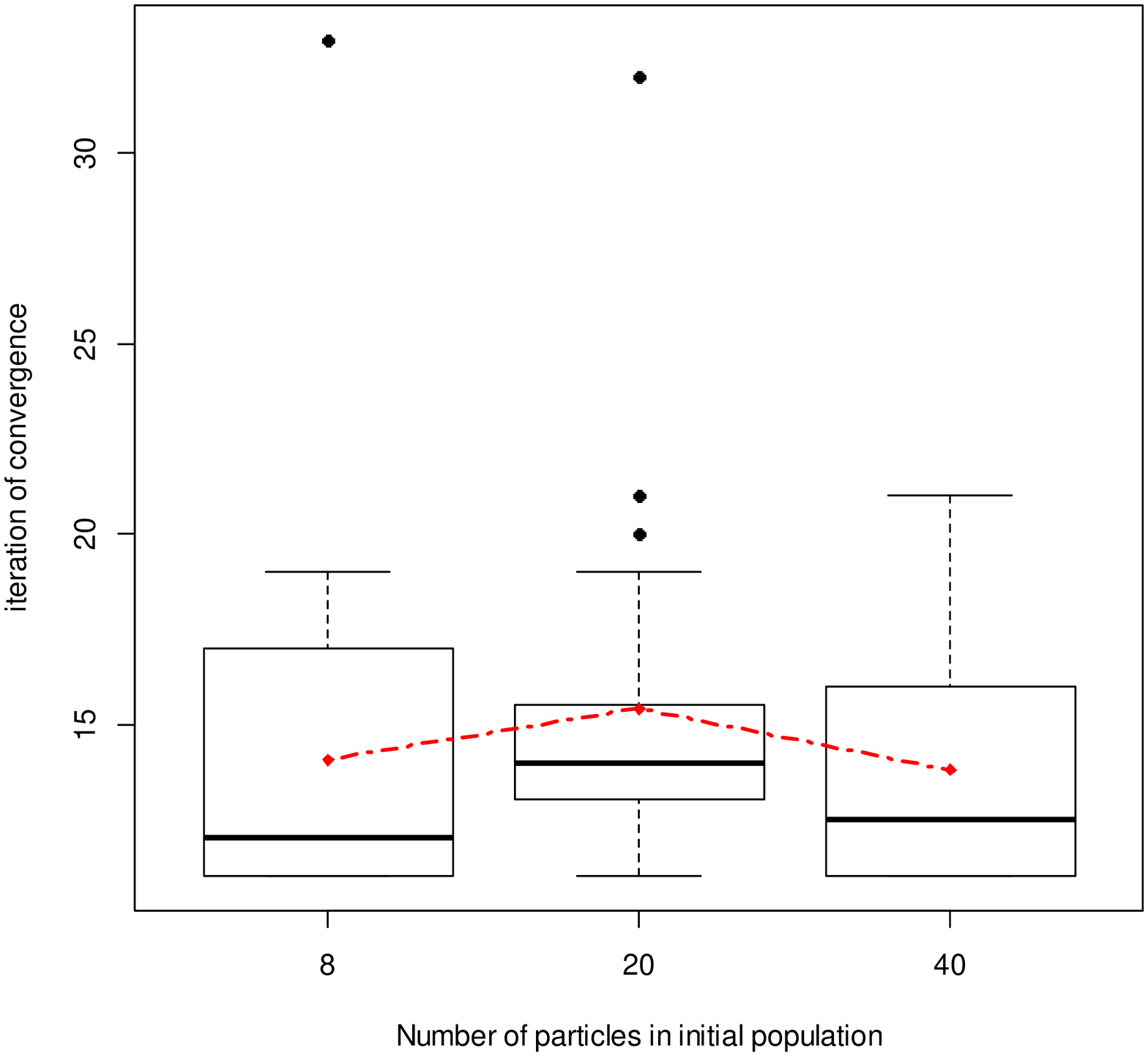}
\caption{$c_1$=1, $c_2$=2, No. of tasks=100} \label{fig:convergence6}
\end{subfigure} \\
\begin{subfigure}{0.3\textwidth}
  \includegraphics[width=\linewidth,keepaspectratio]{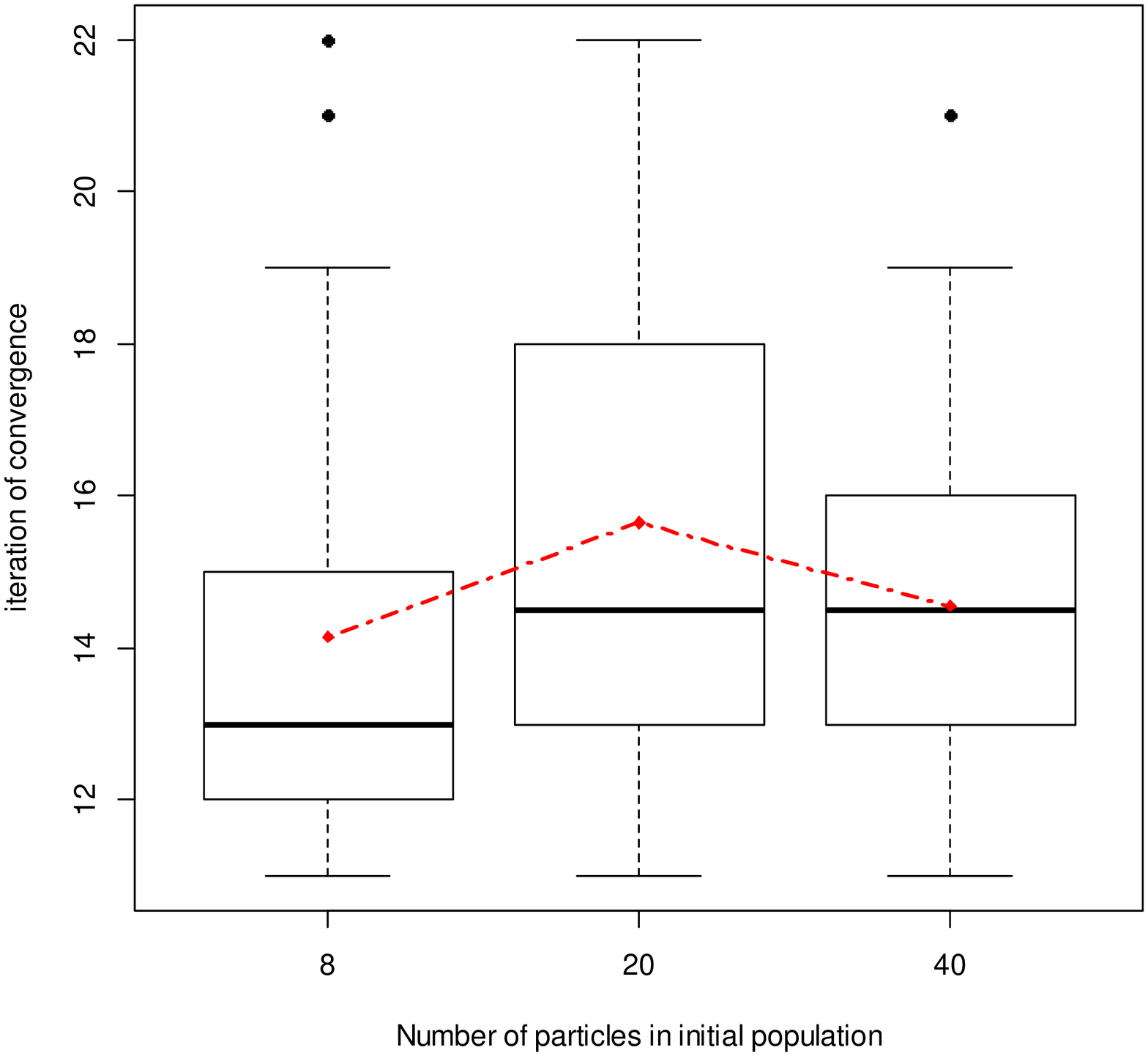}
\caption{$c_1$=2, $c_2$=1, No. of tasks=10} \label{fig:convergence7}
\end{subfigure}
\begin{subfigure}{0.3\textwidth}
  \includegraphics[width=\linewidth,keepaspectratio]{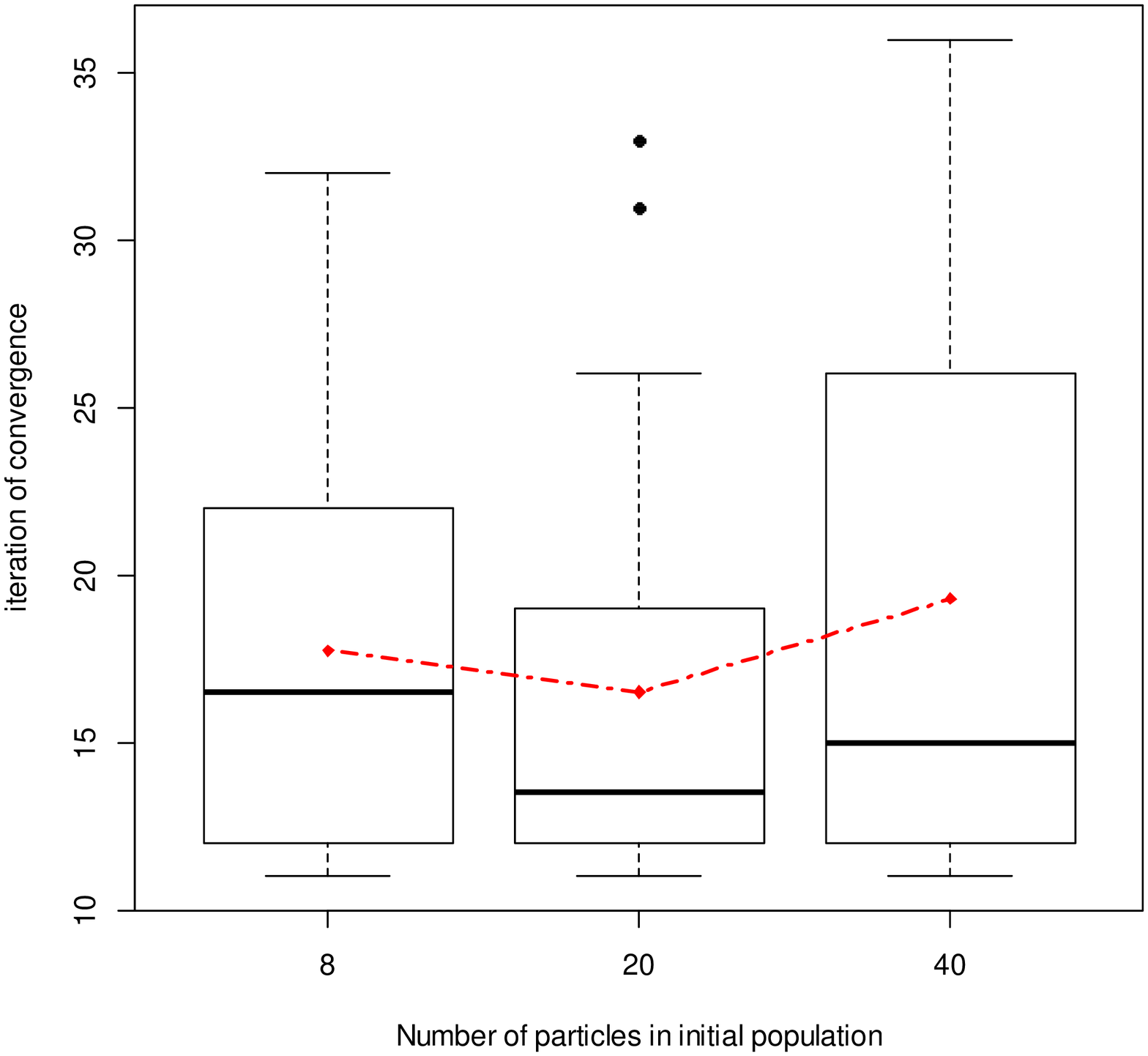}
\caption{$c_1$=2, $c_2$=1, No. of tasks=50} \label{fig:convergence8}
\end{subfigure}
\begin{subfigure}{0.3\textwidth}
  \includegraphics[width=\linewidth,keepaspectratio]{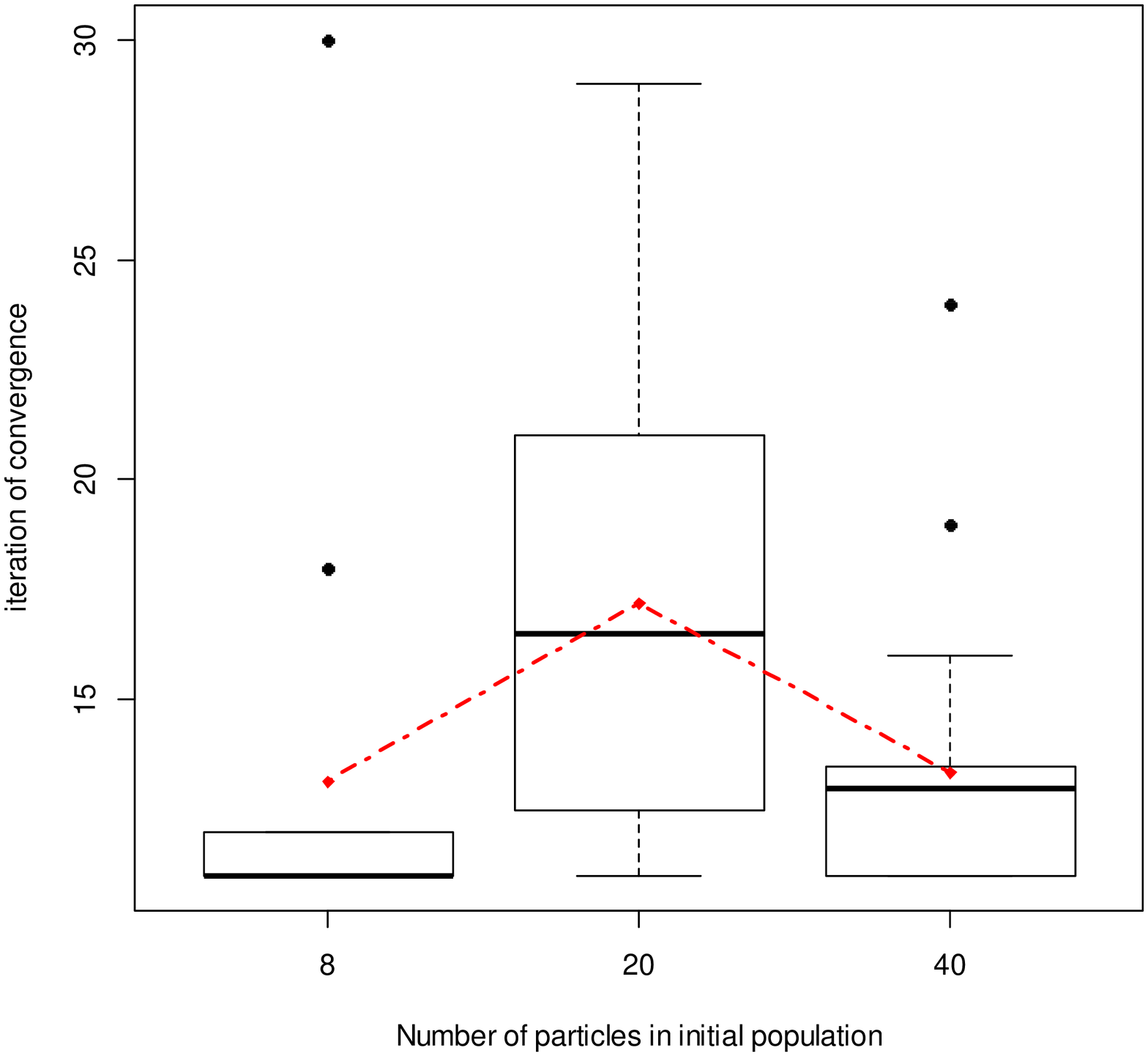}
\caption{$c_1$=2, $c_2$=1, No. of tasks=100} \label{fig:convergence9}
\end{subfigure} \\
\begin{subfigure}{0.3\textwidth}
  \includegraphics[width=\linewidth,keepaspectratio]{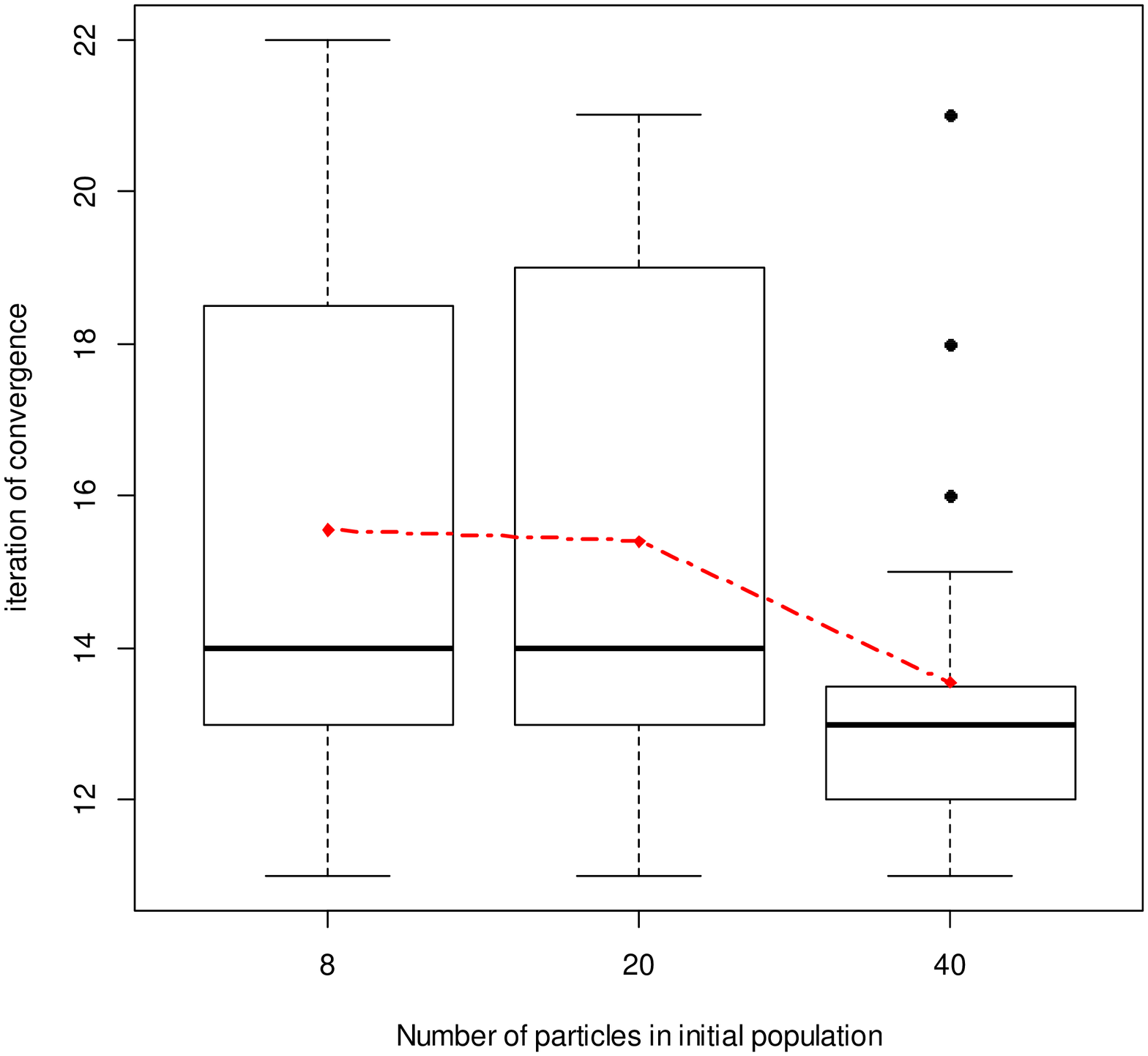}
\caption{$c_1$=2, $c_2$=2, No. of tasks=10} \label{fig:convergence10}
\end{subfigure}
\begin{subfigure}{0.3\textwidth}
  \includegraphics[width=\linewidth,keepaspectratio]{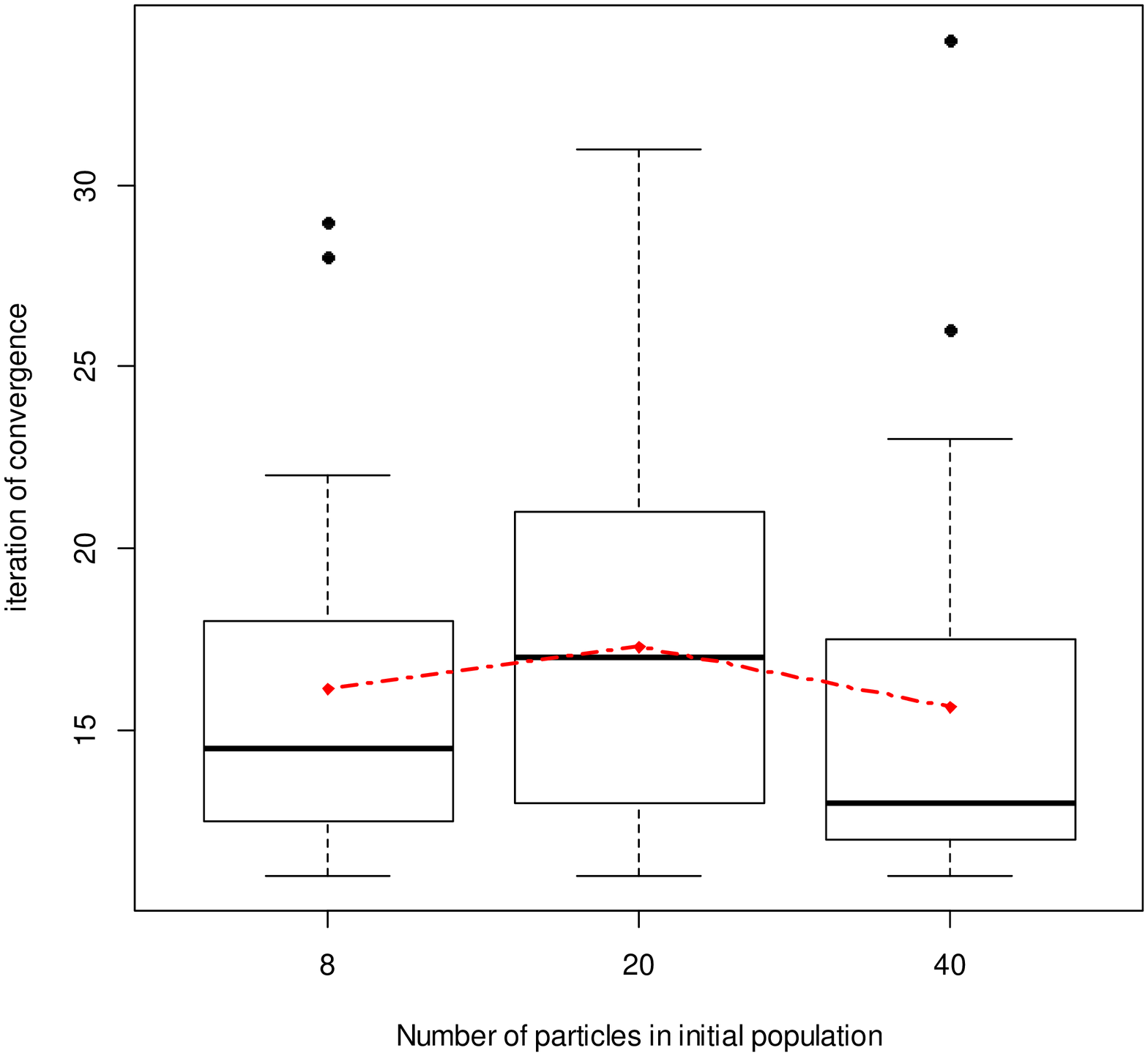}
\caption{$c_1$=2, $c_2$=2, No. of tasks=50} \label{fig:convergence11}
\end{subfigure}
\begin{subfigure}{0.3\textwidth}
  \includegraphics[width=\linewidth,keepaspectratio]{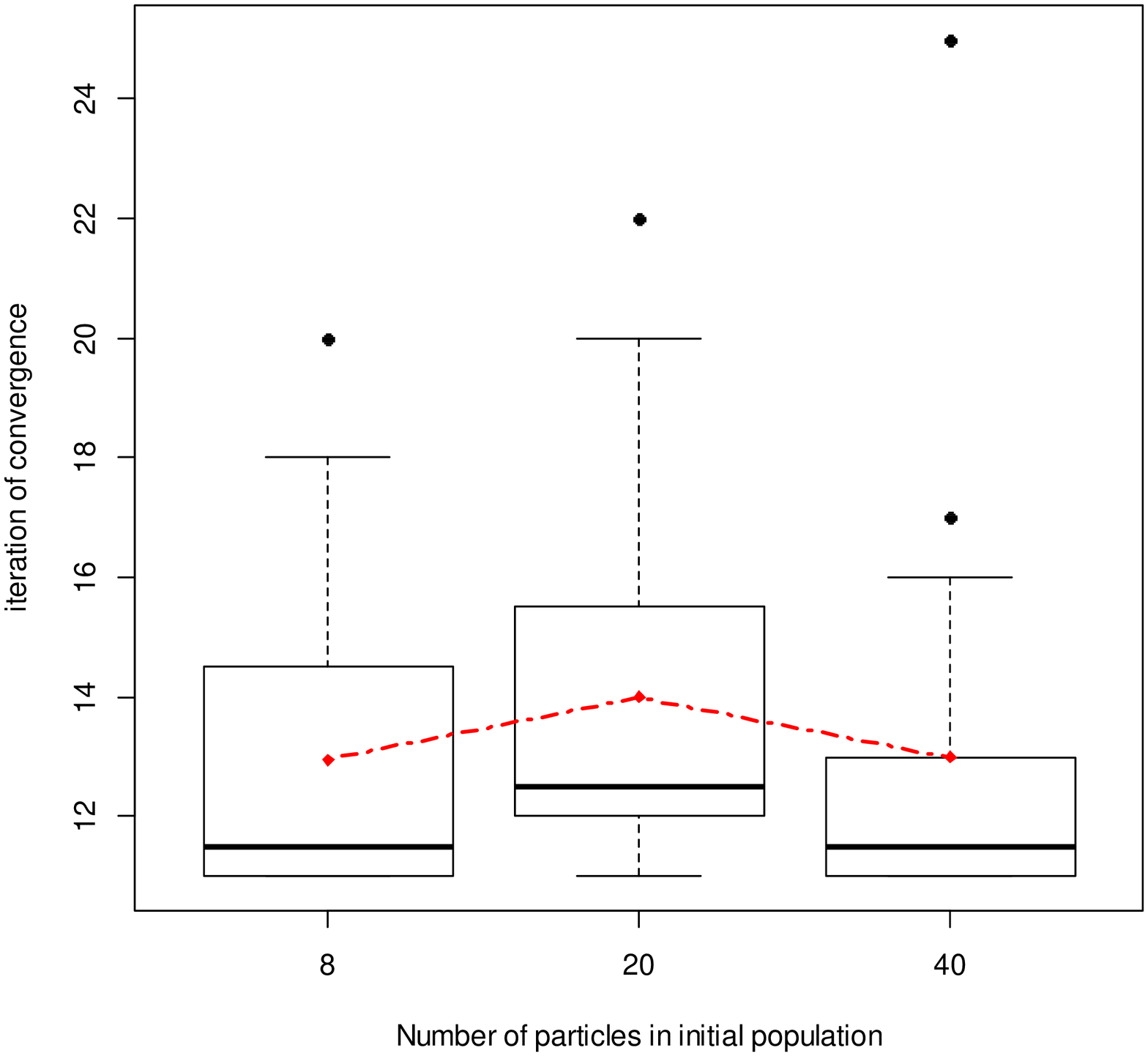}
\caption{$c_1$=2, $c_2$=2, No. of tasks=100} \label{fig:convergence12}
\end{subfigure}
\caption{Convergence speed analysis}
\label{fig:figure_convergence}
\end{figure}

Previously, in Figure \ref{fig:figure_makespan}, the optimality level evaluation for several sets of parameters has been addressed. Afterwards, convergence speed evaluation of each set of parameter is shown respectively in Figure \ref{fig:figure_convergence}. This convergence evaluation is done by checking the value of makespan from the obtained solution. Once there is no improvement after 10 contiguous iterations, the search is concluded to be converged at that particular iteration. For all combinations, most of the test cases converged before iteration 40. For the best case of interest so far, where $c_1=1$, $c_2=2$, and number of initial particles equals to 40, the search managed to converge under iteration 25. The ultimate goal is to cut the excessive number of iterations if the search is proved to most likely converge only after less than the originally given one. Hence, one might think about cutting the number of iteration to 30 because even with this limitation, the search is able to bring a good feasible schedule for the given dataset. However, for a marginal tolerance, the maximum number of iteration may still be set to 40; the highest number of iterations so far from all searches. In this manner, it still gives some space for the search to explore more if it is trapped in some local optimum points in the beginning of the search.

\section{Results and Discussion}
\label{sec:sec_resultsdiscussion}
\begin{figure} [htp]
\centering
\begin{subfigure}{0.45\textwidth}
  \includegraphics[width=\linewidth,keepaspectratio]{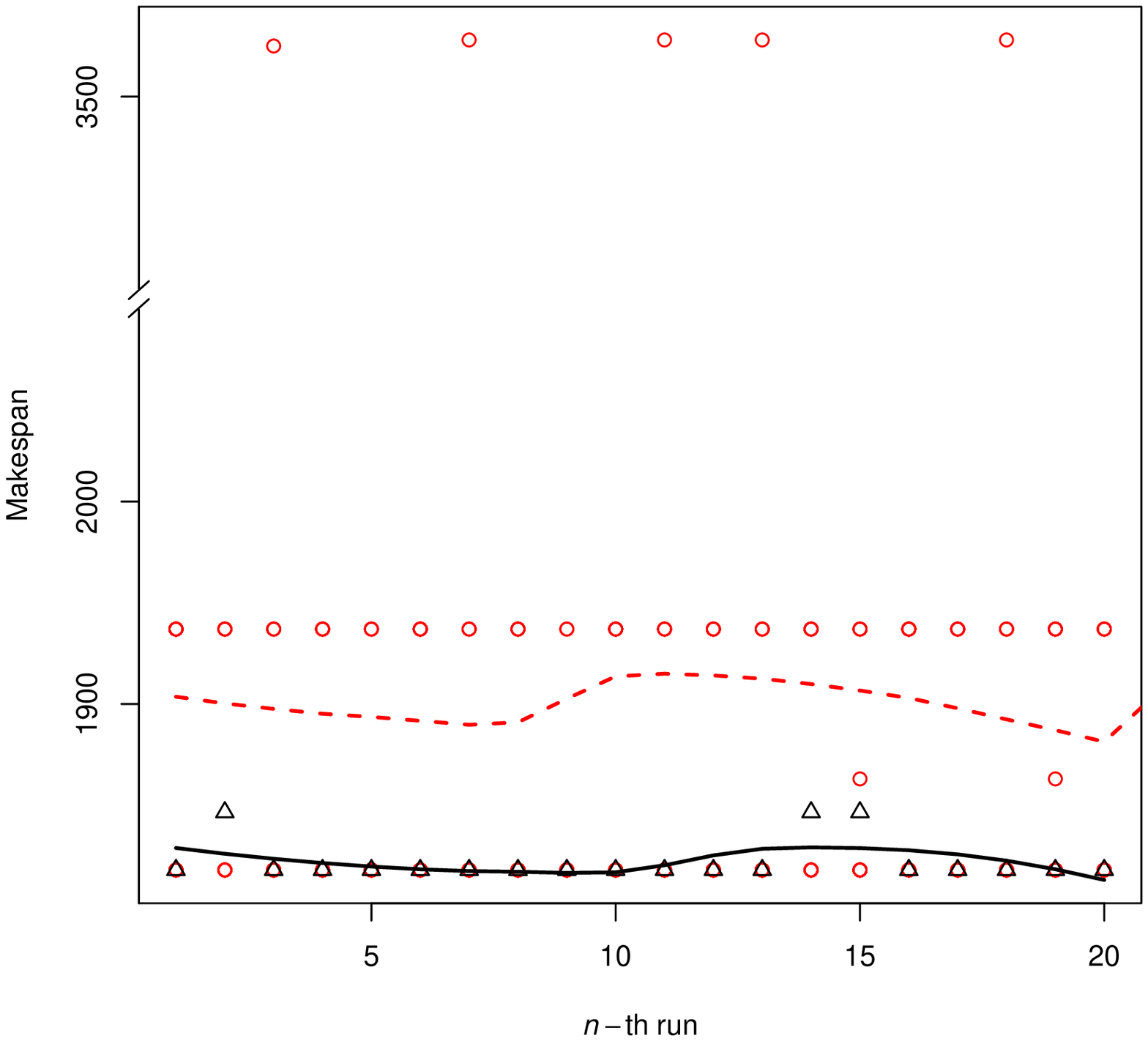}
\caption{No. of tasks=10} \label{fig:res1}
\end{subfigure}
\begin{subfigure}{0.45\textwidth}
  \includegraphics[width=\linewidth,keepaspectratio]{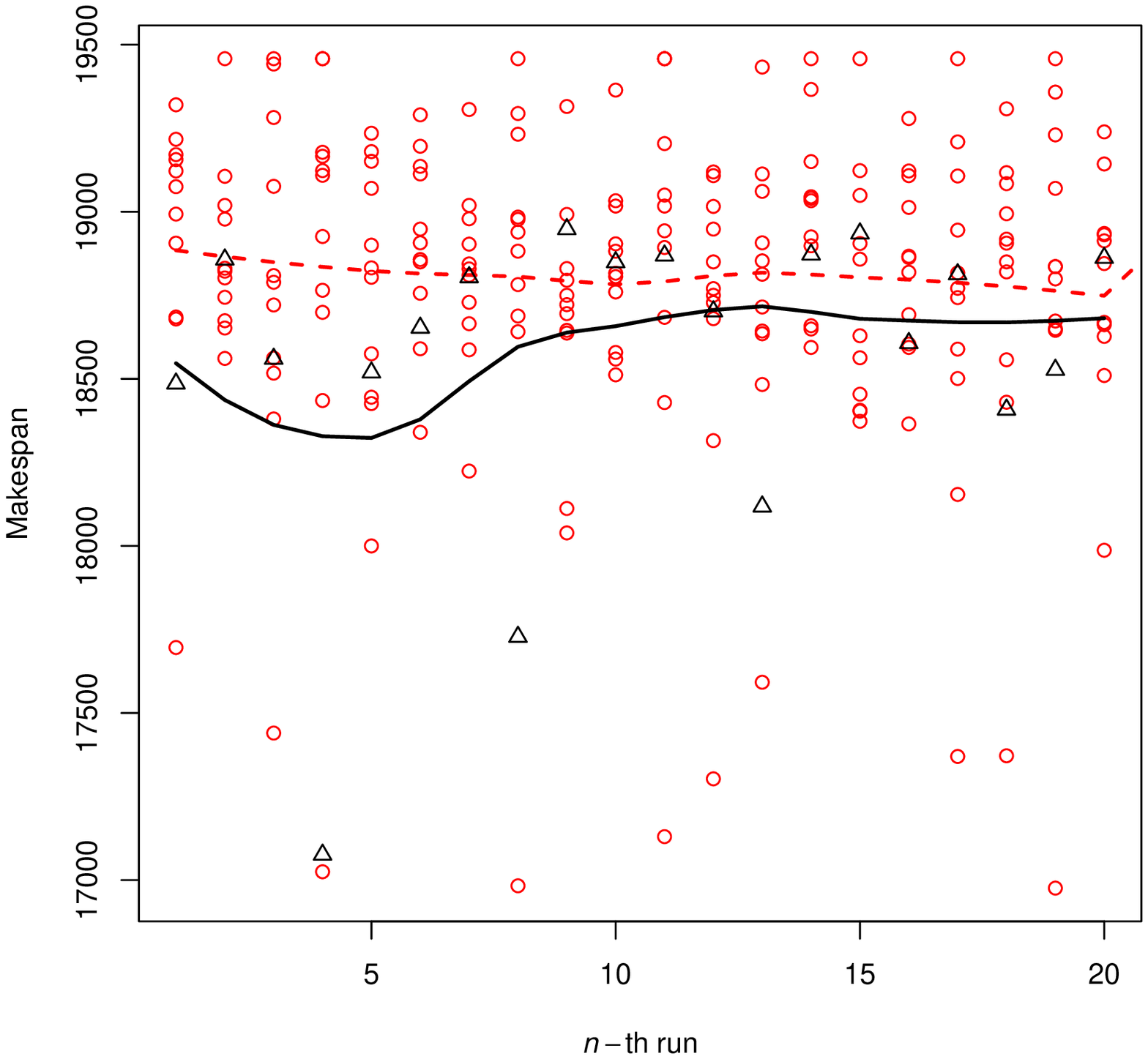}
\caption{No. of tasks=50} \label{fig:res2}
\end{subfigure}\\
\begin{subfigure}{0.45\textwidth}
  \includegraphics[width=\linewidth,keepaspectratio]{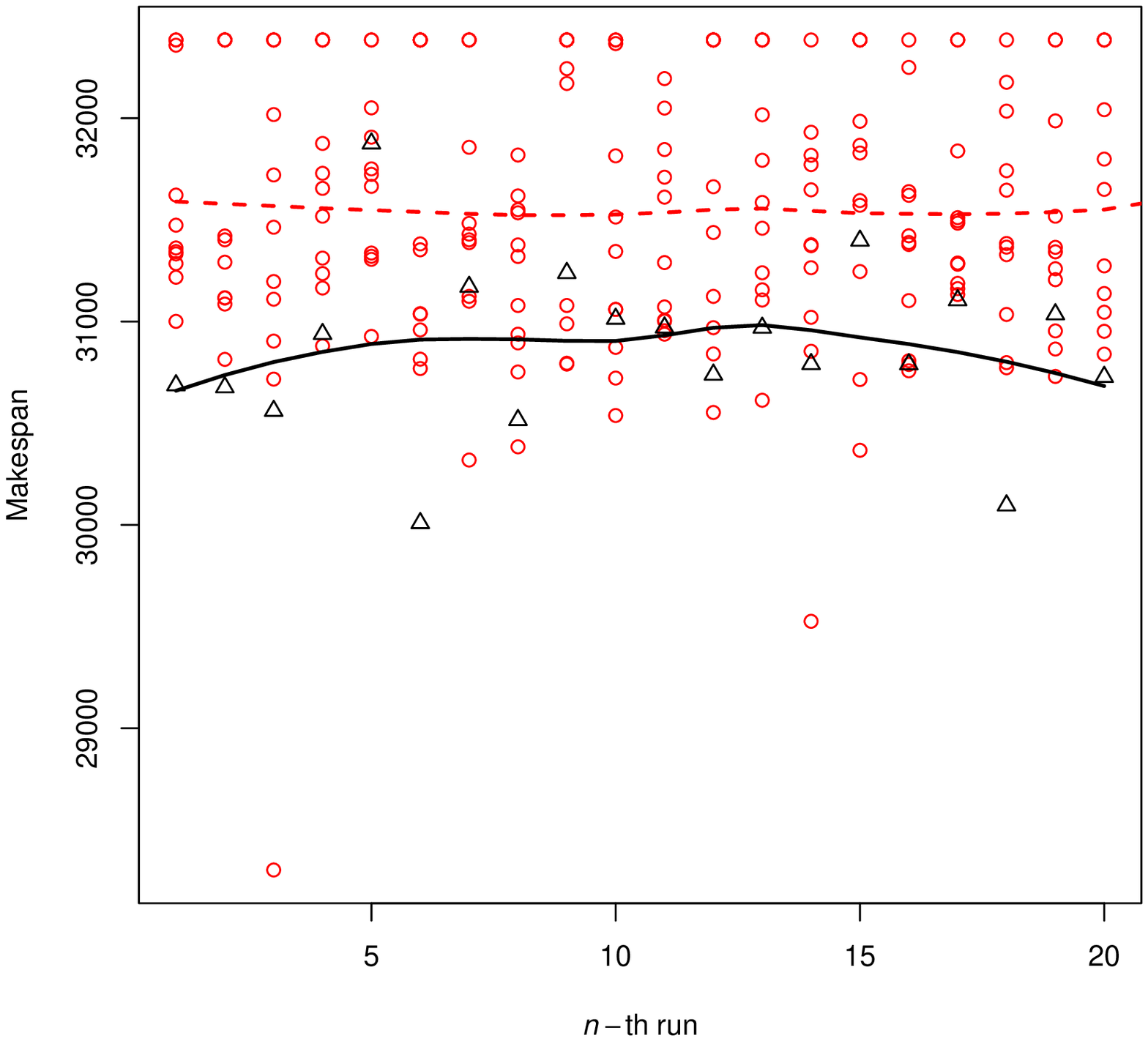}
\caption{No. of tasks=100} \label{fig:res3}
\end{subfigure}
\begin{subfigure}{0.45\textwidth}
\centering
  \includegraphics[scale=0.35]{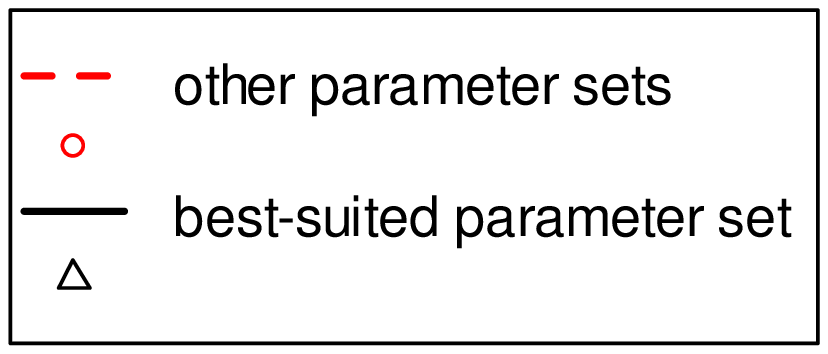}
\end{subfigure}
\caption{Position of makespans from the best suited parameters against others}
\label{fig:figure_result}
\end{figure}

From the numerical experiment, the best suited set of parameters for the search is selected: $c_1=1$, $c_2=2$, number of initial particles = 40, and number of maximum iterations = 40.
The result is shown in Figure \ref{fig:figure_result}; the graphs are distinguished by the number of tasks. Each graph represents points and line (with local polynomial regression fitting) plot of makespans obtained from search which uses the best suited set of parameters (represented by triangle points and full line) against others (represented by circle points and dashed line); with composition of 20:220 respectively, resulting in total of 240 makespans plotted in each graph. The way the data plotted on y-axis is based on the respective makespan values. As for x-axis, it is based on the run index values; as mentioned in \ref{sec:sec_numexp}, each set of parameters is tested for 20 runs.

In reference with data shown in Figure \ref{fig:figure_makespan}, a numerical summary of the experiments with the selected set of parameters on three task datasets are depicted in Table \ref{table:table_selected_result}.
The presented numbers are in accordance with the makespans obtained from the selected parameters, which are displayed in Figure \ref{fig:figure_result}. Both (lines in) Figure \ref{fig:figure_result} and Table \ref{table:table_selected_result} can show a relatively small variance of makespans, which indicates a stable performance of the selected set of parameters.
\begin{table} [htp]
\caption{Numerical summary of experiments with selected set of parameters}
\begin{center}
    \begin{tabulary}{\textwidth}{C | C | C | C | C | C | C | C}
    \hline\noalign{\smallskip}
    \multicolumn{3}{c|}{Parameter} & \multirow{2}{*}{No. of tasks} & \multicolumn{4}{c}{Makespan}\\ \cline{1-3}\cline{5-8}
    $c_1$ & $c_2$ & No. of initial particles && Min & Max & Average & Median\\
    \noalign{\smallskip}\hline\noalign{\smallskip}
    \multirow{3}{*}{1}&\multirow{3}{*}{2}&\multirow{3}{*}{40}&10&1818&1937&1835,85&1818\\
	&&&50&17076&18948&18559,65&18677,5\\
	&&&100&30009&31876&30865,65&30865\\
    \hline
\end{tabulary}
\end{center}
\label{table:table_selected_result}
\end{table}

\begin{figure}[htp]
\centering
  \includegraphics[width=0.6\textwidth,keepaspectratio]{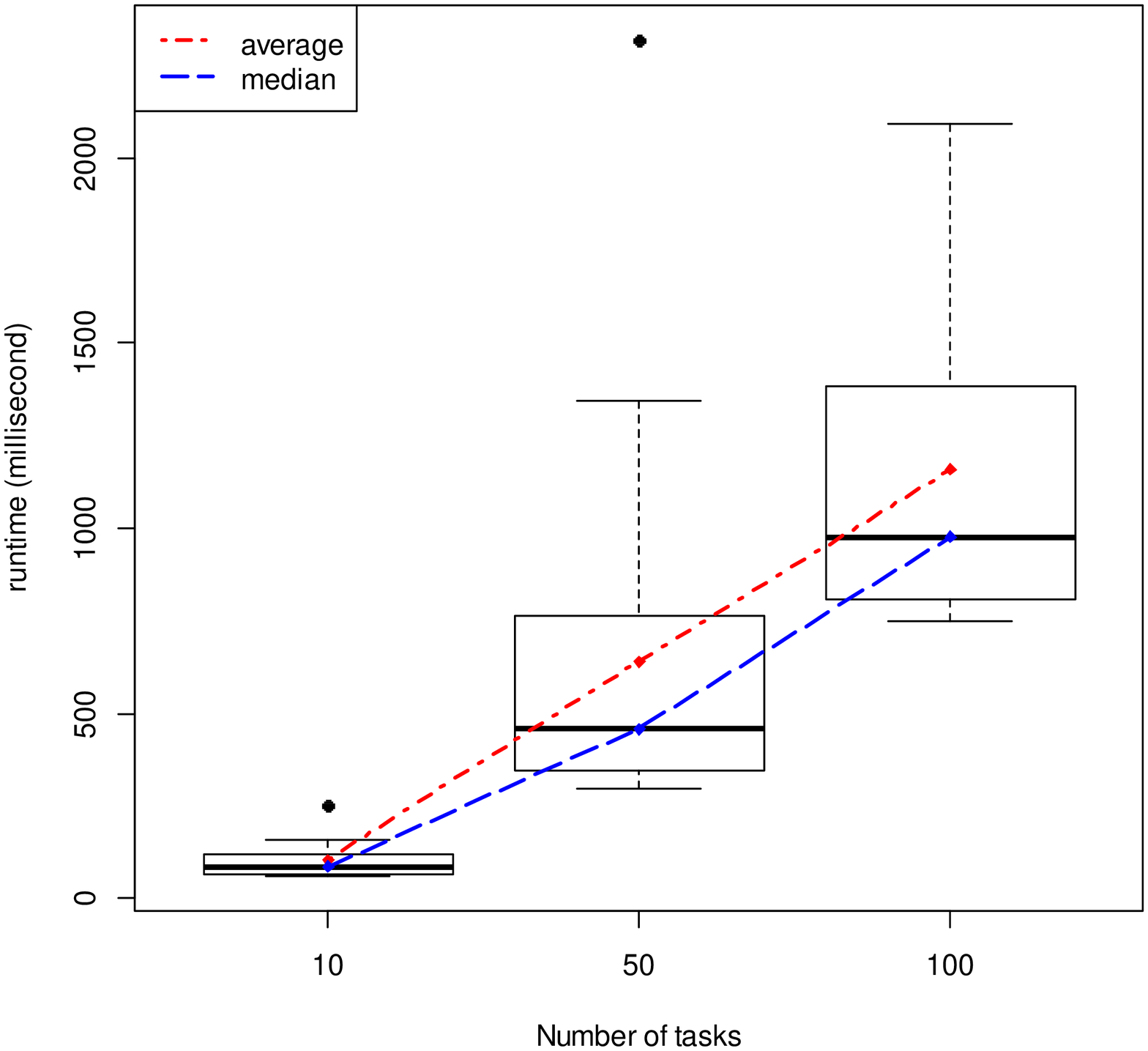}
\caption{Computation time of proposed algorithm for a selected set of parameters}
\label{fig:figure_comp_time}
\end{figure}

Furthermore, a computation time measurement of the selected parameter set is depicted in Figure \ref{fig:figure_comp_time}.
Dotted dashed line in Figure \ref{fig:figure_comp_time} depicts the average computation time while the dashed line depicts median of the makespan throughout the conducted experiments.
It is shown that as the number of tasks scales (10, 50, 100), the computation time tends to scale linearly (it varies around an average of 102.1, 639.25, and 1158.25 milliseconds respectively).
It is possible to see that the presented methodology is efficient because during those short computation times, it can obtain a set of schedules and choose the best one through a number of iterations; conducting a solid optimization. Each boxplot of the task dataset represents 20 computation times. There is a computation time with 2321 milliseconds on 20 tasks schedule generation, which is obviously far from the average. Hence, it can be inferred as an outlier; which is bound to happen in metaheuristic.

\section{Conclusion}
\label{sec:sec_conclusion}
UAV application has been emerging in various domain applications; manufacturing industry environment is one of the most uprising ones. In manufacturing environment, an indoor UAV operations is established. It exposes a number of new benefits and challenges at the same time. Autonomous UAV operations cut back the number of human intervention but it certainly requires a precise UAV flight control and scheduling system. To overcome such challenges of UAV operations in indoor environment, scheduler is one essential component of the solution. This work proposes a methodology which control execution of tasks by UAV over a period of time to achieve a minimum total makespan. A heuristic (based on Earliest Available Time algorithm) for assigning tasks to UAVs which forms a seamless UAV operations schedule (considering required hover and wait-on-ground) is proposed. This heuristic is incorporated with particle swarm optimization (PSO) algorithm which promotes a quick computation time in finding a high quality feasible solution. The implementation of the proposed methodology solves the presented problem, which has not been done so far. Numerical experiments are conducted to analyze the behavior of PSO parameters and select the best-suited set of parameters. The performance of the implemented methodology is also measured through its convergence speed and computation time. The obtained results are presented and discussed in detail. In future work, anti-collision refinement phase will be incorporated with this work. In addition, the performance of current methodology can be compared with other well-known metaheuristics reported in the literature.

\begin{acknowledgements}
This work has partly been supported by Innovation Fund Denmark under project UAWorld; grant agreement number 9-2014-3.
\end{acknowledgements}

\bibliography{citations}{}
\bibliographystyle{plain}

\section*{Appendix}
\begin{appendix}
\section{Illustrative usage of Algorithm \ref{algo:algo_heuristic} (step 3-7) on task sequence in Figure \ref{fig:figure_task_scheduling_step}}
\label{app:figure_steps_desc}
\begin{longtable}{c | l l}
    \multicolumn{3}{c}{
    \parbox{\textwidth}{
      \centering
      \includegraphics[scale=0.7]{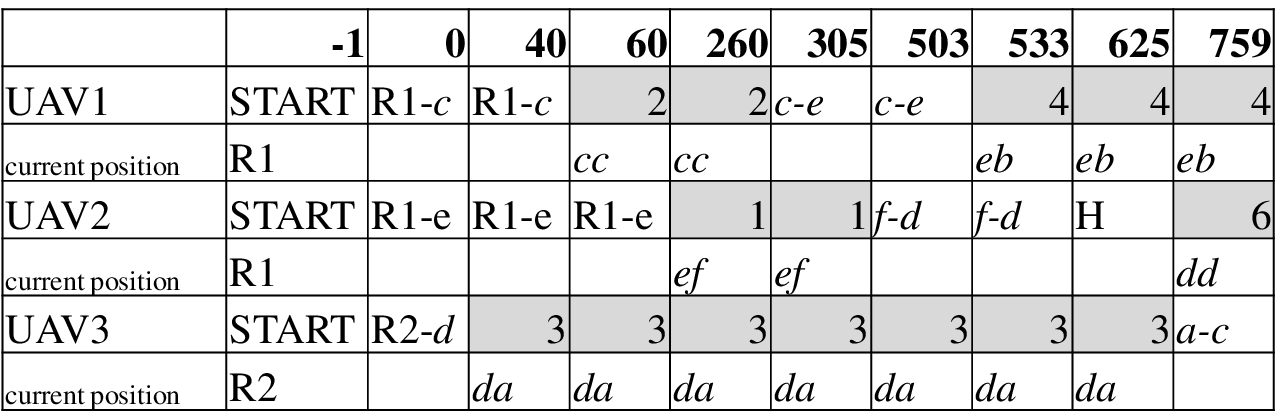}
      \figcaption{Output of step 3 of schedule creation heuristic}
      \label{fig:step3}}
    }\\
    \hline
    Step & \multicolumn{2}{c}{Description}\\
    \hline
    3 & \multicolumn{2}{l}{Task 1}\\
    &\multicolumn{2}{l}{(a) Check earliest available time of task 1}\\
    &\multicolumn{2}{l}{Task 1 is available from time 0.}\\
    \\
    &\multicolumn{2}{l}{(b) Check which UAV needs recharge before executing task 1}\\
    &UAV1&: 305+228+243+60= 836\\
    &UAV2&: 0+260+243+60 = 563\\
    &UAV3&: 759+346+243+60 = 1408\textgreater1200\\
    &\multicolumn{2}{l}{UAV3 needs a recharge.}\\
    \\
    &\multicolumn{2}{l}{Compare the earliest available time of each UAV}\\
    &UAV1&: 305+228 = 533\\
    &UAV2&: 0+260 = 260\\
    &UAV3&: 759+40+2700+260 = 3759\\
    &&: {759+160+2700+60 = 3679}\\
    &\multicolumn{2}{l}{$\therefore$ UAV2 is picked for task 1}\\
    \hline\noalign{\smallskip}
    \multicolumn{3}{c}{
    \parbox{\textwidth}{
      \centering
      \includegraphics[scale=0.7]{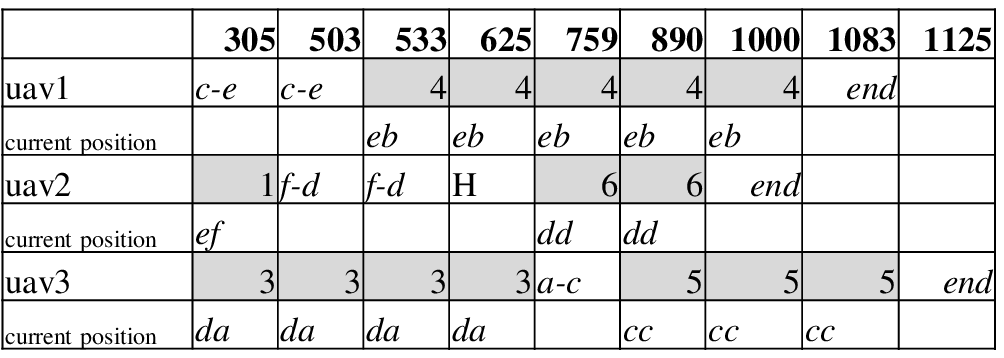}
      \figcaption{Output of step 4-6 of schedule creation heuristic}
      \label{fig:step4-6}}
    }\\
    \hline
    Step & \multicolumn{2}{c}{Description}\\
    \hline
    4 & \multicolumn{2}{l}{Task 4}\\
    &\multicolumn{2}{l}{(a) Check earliest available time of task 4}\\
    &\multicolumn{2}{l}{Task 4 is available from time:}\\
    &&$\cdot$503 based on task precedence\\
    &&$\cdot$503 based on position availability\\
    \\
    &\multicolumn{2}{l}{(b) Check which UAV needs recharge before executing task 4}\\
    &UAV1&: 533 {(305+228)\textgreater503}+550+60 = 1143\\
    &UAV2&: 626 {(503+123)\textgreater503}+550+60 = 1236\textgreater1200\\
    &UAV3&: 759 {(503+123)\textgreater503}+346+550+60 = 1715\textgreater1200\\
    &\multicolumn{2}{l}{UAV2 and UAV3 need a recharge}\\
    \\
    &\multicolumn{2}{l}{Compare the earliest available time of each UAV}\\
    &UAV1&: 305+228 = 533\\
    &UAV2&: 503+260+2700+260 = 3723\\
    &&: 503+60+2700+60 = 3323\\
    &UAV3&: 759+40+2700+260 = 3759\\
    &&: 759+160+2700+60 = 3679\\
    &\multicolumn{2}{l}{$\therefore$ UAV1 is picked for task 4.}\\
    \hline
    5 & \multicolumn{2}{l}{Task 6}\\
    &\multicolumn{2}{l}{(a) Check earliest available time of task 6}\\
    &\multicolumn{2}{l}{Task 6 is available from time:}\\
    &&$\cdot$305 based on task precedence\\
    &&$\cdot$759 based on position availability\\
    \\
    &\multicolumn{2}{l}{(b) Check which UAV needs recharge before executing task 6}\\
    &UAV1&: 1083+241+241+40 = 1605\textgreater1200\\
    &UAV2&: 759 {(503+122)\textless759}+241+40 = 1040\\
    &UAV3&: 759+222+241+40 = 1262\textgreater1200\\
    &\multicolumn{2}{l}{UAV1 and UAV3 need a recharge}\\
    \\
    &\multicolumn{2}{l}{Compare the earliest available time of each UAV}\\
    &UAV1&: 1083+60+2700+160 = 4003\\
    &&: 1083+160+2700+40 = 3983\\
    &UAV2&: 759\\
    &UAV3&: 759+40+2700+160 = 3659\\
    &&: 759+160+2700+40 = 3659\\
    &\multicolumn{2}{l}{$\therefore$ UAV2 is picked for task 6.}\\
    \hline
    6 & \multicolumn{2}{l}{Task 5}\\
    &\multicolumn{2}{l}{(a) Check earliest available time of task 5}\\
    &\multicolumn{2}{l}{Task 5 is available from time:}\\
    &&$\cdot$305 based on task precedence\\
    &&$\cdot$305 based on position availability\\
    \\
    &\multicolumn{2}{l}{(b) Check which UAV needs recharge before executing task 5}\\
    &UAV1&: 1083+120+235+60 = 1498\textgreater1200\\
    &UAV2&: 1000+127+235+60 = 1422\textgreater1200\\
    &UAV3&: 759+131+235+60 = 1185\\
    &\multicolumn{2}{l}{UAV1 and UAV2 need a recharge}\\
    \\
    &\multicolumn{2}{l}{Compare the earliest available time of each UAV}\\
    &UAV1&: 1083+60+2700+60 = 3903\\
    &&: 1083+160+2700+60 = 4003\\
    &UAV2&: 1000+160+2700+60 = 3920\\
    &&: 1000+40+2700+60 = 3800\\
    &UAV3&: 759+131 = 890\\
    &\multicolumn{2}{l}{$\therefore$ UAV3 is picked for task 5.}\\
    \hline\noalign{\smallskip}
    \multicolumn{3}{c}{
    \parbox{\textwidth}{
      \centering
      \includegraphics[scale=0.7]{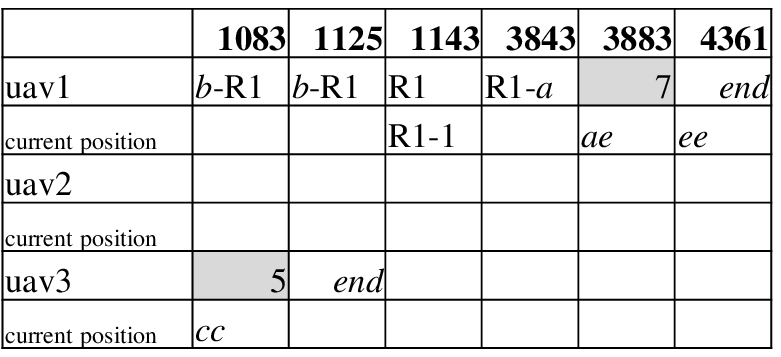}
      \figcaption{Output of step 7 of schedule creation heuristic}
      \label{fig:step7}}
    }\\
    \hline
    Step & \multicolumn{2}{c}{Description}\\
    \hline
    7 & \multicolumn{2}{l}{Task 7}\\
    &\multicolumn{2}{l}{(a) Check earliest available time of task 7}\\
    &\multicolumn{2}{l}{Task 7 is available from time:}\\
    &&$\cdot$1083 based on task precedence\\
    &&$\cdot$1083 based on position availability\\
    \\
    &\multicolumn{2}{l}{(b) Check which UAV needs recharge before executing task 7}\\
    &UAV1&: 1083+108+478+60 = 1729\textgreater1200\\
    &UAV2&: 1000+222+478+60 = 1760\textgreater1200\\
    &UAV3&: 1125+131+478+60 = 1794\textgreater1200\\
    &\multicolumn{2}{l}{UAV1, UAV2, and UAV3 need a recharge}\\
    \\
    &\multicolumn{2}{l}{Compare the earliest available time of each UAV}\\
    &UAV1&: 1083+ 60+2700+40 = 3883\\
    &&:1083+160+2700+160 = 4103\\
    &UAV2&: 1000+160+2700+40 = 3900\\
    &&: 1000+40+2700+160 = 3900\\
    &UAV3&: 1125+60+2700+40 = 3925\\
    &&: 1125+60+2700+160 = 4045\\
    &\multicolumn{2}{l}{$\therefore$ UAV1 is picked for task 7.}\\ \hline
\end{longtable}
\end{appendix}

\end{document}